\documentclass{article}

\usepackage{microtype}
\usepackage{graphicx}
\usepackage{subcaption}
\usepackage{booktabs} 

\usepackage{hyperref}

\usepackage[accepted]{icml2024}

\usepackage{amsmath}
\usepackage{amssymb}
\usepackage{mathtools}
\usepackage{amsthm}

\usepackage[capitalize,noabbrev]{cleveref}

\theoremstyle{plain}

\theoremstyle{definition}

\theoremstyle{remark}

\usepackage[textsize=tiny]{todonotes}

\usepackage{amsmath,amsfonts,bm}

\def\eqref#1{equation~\ref{#1}}

\def\1{\bm{1}}

\DeclareMathAlphabet{\mathsfit}{\encodingdefault}{\sfdefault}{m}{sl}
\SetMathAlphabet{\mathsfit}{bold}{\encodingdefault}{\sfdefault}{bx}{n}

\DeclareMathOperator*{\argmax}{arg\,max}
\DeclareMathOperator*{\argmin}{arg\,min}

\graphicspath{ {./img/} }
\usepackage{bbm}
\usepackage[group-separator={,}]{siunitx}
\usepackage{multirow}
\usepackage{booktabs}
\usepackage{xcolor}
\usepackage{url}
\usepackage{xurl}
\usepackage{xspace}
\usepackage{multicol}
\usepackage{subcaption}
\usepackage[inline]{enumitem}
\usepackage[algo2e, ruled, linesnumbered, commentsnumbered, noend]{algorithm2e}

\usepackage[page,header]{appendix}
\usepackage{titletoc}

\newcommand{\lfbo}{\textsc{LFBO}\xspace}
\newcommand{\lfbobb}{\textsc{LFBO+BB}\xspace}
\newcommand{\ourmethod}{\textsc{MALIBO}\xspace}
\newcommand{\ie}{i.e., }
\newcommand{\eg}{e.g., }

\icmltitlerunning{MALIBO: Meta-learning for Likelihood-free Bayesian Optimization}

\begin{document}

\twocolumn[
\icmltitle{MALIBO: Meta-learning for Likelihood-free Bayesian Optimization}



\icmlsetsymbol{equal}{*}

\begin{icmlauthorlist}
\icmlauthor{Jiarong Pan}{bosch,tue}
\icmlauthor{Stefan Falkner}{bosch}
\icmlauthor{Felix Berkenkamp}{bosch}
\icmlauthor{Joaquin Vanschoren}{tue}
\end{icmlauthorlist}

\icmlaffiliation{tue}{Eindhoven University of Technology, Netherlands}
\icmlaffiliation{bosch}{Bosch Center for Artificial Intelligence, Germany}

\icmlcorrespondingauthor{Jiarong Pan}{fixed-term.jiarong.pan@de.bosch.com}
\icmlkeywords{Machine Learning, ICML}

\vskip 0.3in
]



\printAffiliationsAndNotice{}  

\begin{abstract}
Bayesian optimization (BO) is a popular method to optimize costly black-box functions, and meta-learning has emerged as a way to leverage knowledge from related tasks to optimize new tasks faster. However, existing meta-learning methods for BO rely on surrogate models that are not scalable or are sensitive to varying input scales and noise types across tasks. Moreover, they often overlook the uncertainty associated with task similarity, leading to unreliable task adaptation when a new task differs significantly or has not been sufficiently explored yet. We propose a novel meta-learning BO approach that bypasses the surrogate model and directly learns the utility of queries across tasks. It explicitly models task uncertainty and includes an auxiliary model to enable robust adaptation to new tasks. Extensive experiments show that our method achieves strong performance and outperforms multiple meta-learning BO methods across various benchmarks.
\end{abstract}

\section{Introduction}
\label{sec:introduction}
Bayesian optimization (BO) \citep{shahriari2016loop} is a widely used framework to optimize expensive black-box functions for a wide range of applications, from material design \citep{frazier2015bayesian} to automated machine learning \citep{hutter2019automl}. Traditionally, it uses a probabilistic \textit{surrogate model}, often a Gaussian process (GP), to model the black-box function and provide uncertainty estimates that can be used by an \textit{acquisition function} to propose the next query point.

While BO typically focuses on each new target task individually, recent approaches leverage information from previous runs on related tasks through transfer learning \citep{weiss2016transfer} and meta-learning \citep{vanschoren2018meta} to \textit{warm-start} BO. In this context, each \textit{task} denotes the optimization of a specific black-box function and we assume that related tasks share similarities with the target task. For instance, one can warm-start the tuning of a neural network when the same network was previously tuned on related datasets. Previous runs on related tasks can be used to build informed surrogate models \citep{perrone2018scalable, wistuba2021fewshot, feurer2022practical}, restrict the search space \citep{perrone2019searchspace}, or initialize the optimization with configurations that generally score well \citep{feurer2014meta, volpp2020metalearning}.

However, the use of surrogate models also engenders several issues in many of these approaches:
\begin{enumerate*}[label=(\roman*)]
 	\item GP-based methods scale poorly with the number of observations as well as number of tasks, due to their cubic computational complexity \citep{rasmussen2004gpml}.
    \item In practice, observations across tasks can have different scales, \eg the validation error of an algorithm can be high on one dataset and low on another. Although normalization can be applied to the data from related tasks, normalizing the unseen (target) task data is often challenging, especially when only a few observations are available to estimate its range. Regression-based surrogate models therefore struggle to adequately transfer knowledge from related tasks \citep{bardenet2013collaborative,yogatama2014efficient}.
    \item While GPs typically assume the observation noise to be Gaussian and homoscedastic, real-world observations often have different noise distributions and can be heteroscedastic. This discrepancy can lead to poor meta-learning and optimization performance \citep{salinas2023optimizing}.
\end{enumerate*}
Moreover, when adapting to tasks that have limited observations (\eg early iterations during optimization) or tasks that are significantly different from those seen before, estimating the task similarity becomes challenging due to the scarcity of relevant task information. Hence, it is desirable to explicitly model the uncertainty inherent to such tasks \citep{finn2018probabilistic}. Nevertheless, many existing methods warm-start BO by only modeling relations between tasks deterministically \citep{wistuba2018scalable,volpp2020metalearning}, making the optimization unreliable.

To tackle these limitations, we propose a novel and scalable meta-learning BO approach\footnote{Our code is available in the following repository: \url{https://github.com/boschresearch/meta-learning-likelihood-free-bayesian-optimization}} that is inspired by the idea of likelihood-free acquisition function \citep{tiao2021bore, song2022general}. The proposed method overcomes the limitations of surrogate modeling by directly approximating the acquisition function. It makes less stringent assumptions about the observed values, which establishes effective learning across tasks with varying scales and noises. To account for task uncertainty, we introduce a probabilistic meta-learning model to capture the task uncertainty, as well as a novel adaptation procedure based on gradient boosting to robustly adapt to each new task.

This paper makes the following contributions:
\begin{enumerate*}[label=(\roman*)]
	\item We propose a scalable and robust meta-learning BO approach that directly models the acquisition function of a given task based on knowledge from related tasks, while being able to cope with heterogeneous observation scales and noise types across tasks.
    \item We use a probabilistic model to meta-learn the task distribution, which enables us to account for the uncertainty inherent in each target task.
    \item We add a novel adaptation procedure to ensure robust adaptation to new tasks that are not well captured by meta-learning.
\end{enumerate*}

\section{Related Work}
\label{sec:related_work}

\paragraph{Meta-learning Bayesian optimization}
Various methods have been proposed to improve the data-efficiency of BO through meta-learning and have shown effectiveness in diverse applications 
\citep{andrychowicz2016learning}.

One line of work focuses on the initialization of the optimization (\textit{initial design}) by reducing the search space \citep{perrone2019searchspace, li2022searchspace} or reusing promising configurations from similar tasks, where task similarity can be determined using hand-crafted \textit{meta-features} \citep{feurer2014meta} or learned through neural networks (NNs) \citep{kim2017warmstart}. One can also estimate the utility of a configuration using heuristics \citep{wistuba2015learning} or learning-based techniques \citep{volpp2020metalearning,hsieh2021reinforced,maraval2023nap}. Transfer learning is also employed to modify the surrogate model using multi-task GPs \citep{swersky2013multitask, tighineanu2022transfer, tighineanu2024scalable}, additive GP models \citep{golovin2017vizier}, weighted combinations of independent GPs \citep{wistuba2018scalable, feurer2022practical}, shared feature representation learned across tasks \citep{perrone2018scalable,wistuba2021fewshot,khazi2023deep} or pre-training surrogate models on large amount of diverse data \citep{chen2022optformer,muller2023pfns4bo}.

Several methods \textit{simultaneously} learn the initial design and modify the surrogate model. BOHAMIANN \citep{springenberg2016bayesian} adopts a Bayesian NN as the surrogate model, which is computationally expensive and hard to train. ABLR \citep{perrone2018scalable} and BANNER \citep{berkenkamp2021probabilistic} combine a NN to learn a shared feature representation across tasks and task-specific Bayesian linear regression (BLR) layers for scalable adaptation. While ABLR adapts to new tasks by fine-tuning the whole network, BANNER meta-learns a task-independent mean function and only fine-tunes the BLR layer during optimization. However, both methods are sensitive to changes in scale and noise across tasks. 
To address this, GC3P \citep{salinas2020quantile} transforms the observed values via quantile transformation and fits a NN across all related tasks. Although GC3P warm-starts the optimization by using a NN to predict the mean for a GP on the target task, its scalability is limited by its GP surrogate.

\paragraph{Likelihood-free acquisition functions}
Bayesian optimization does not require an explicit model of the likelihood of the observed values \citep{garnett_bayesoptbook_2022} and can be done by directly approximating the acquisition function. The tree-structured Parzen estimator (TPE) \citep{bergstra2011algorithms} phrases BO as a density ratio estimation problem \citep{sugiyama2012density} and uses the density ratio over `good' and `bad' configurations as an acquisition function. BORE \citep{tiao2021bore} estimates the density ratio through class probability estimation \citep{qin1998density}, which is equivalent to modeling the acquisition function with a binary classifier and can be parallellized \citep{oliveira2022batch}. By transforming the acquisition function into a variational problem, likelihood-free Bayesian optimization (\lfbo) \citep{song2022general} uses the probabilistic predictions of a classifier to directly approximate the acquisition function. In this paper, we leverage the flexibility of likelihood-free acquisition functions and combine it with a meta-learning model to obtain a sample-efficient, scalable, and robust BO method.

\section{Background}
\label{sec:background}
\paragraph{Meta-learning Bayesian optimization}
BO aims to minimize a target black-box function $f: \mathcal{X} \rightarrow \mathbb{R}$ over $\mathbf{x} \in \mathcal{X}$. In the case of meta-learning, $T$ related black-box functions $\{f^t(\cdot) \}^T_{t=1}$ are given in advance, each with the same domain $\mathcal{X}$. The optimization is warm-started with previous evaluations on the related functions, $\mathcal{D}^{\text{meta}} = \{\mathcal{D}^t\}^T_{t=1}$ with $\mathcal{D}^t = \{ (\mathbf{x}^t_i, y^t_i) \}^{N^t}_{i=1}$, where $y^t_i = f^t(\mathbf{x}^t_i) + \epsilon^t$ are evaluations corrupted by noise $\epsilon^t$ and $N^t = |\mathcal{D}^t|$ is the number of observations collected from task $f^t$. Given a new task at step $N+1$ , BO proposes $\mathbf{x}_{N+1}$ and obtains a noisy observation from the target function $y_{N+1} = f(\mathbf{x}_{N+1}) + \epsilon$, with $\epsilon$ drawn i.i.d.\ from some distribution $p_{\epsilon}$. To obtain the proposal $\mathbf{x}_{N+1}$, a probabilistic surrogate model is first fitted on $N$ previous observations on the target function $\mathcal{D}_{N} = \{ (\mathbf{x}_i, y_i) \}^{N}_{i=1}$ and the related functions $\mathcal{D}^{\text{meta}}$. For simplicity, we denote $\mathcal D := \mathcal{D}_N \cup \mathcal{D}^{\text{meta}}$. The resulting model is used to compute an acquisition function, such as, the expected utility of a given query $\mathbf x$,
\begin{equation}
\label{eq:expected_utility}
    \alpha^{\text{U}}(\mathbf{x}; \mathcal D, \tau) = \mathbb{E}_{y \sim p(y\mid\mathbf{x}, \mathcal D)}[U(y; \tau)]\,,
\end{equation}
where $U(y; \tau)$ is a chosen utility function with a threshold $\tau$ that decides the utility of observing $y$ at $\mathbf x$ and controls the exploration-exploitation trade-off. The predictive distribution $p(y\mid\mathbf{x}, \mathcal{D})$ is given by the probabilistic surrogate model and the maximizer $\mathbf{x}_{N+1} = \argmax_{\mathbf{x} \in \mathcal{X}} \alpha(\mathbf{x};\mathcal D, \tau)$ is the proposed candidate. Acquisition functions that take the form of \cref{eq:expected_utility} include Expected Improvement (EI) \citep{mockus1975extremum} and Probability of Improvement (PI) \citep{kushner1964maximum}. Many others exist, such as UCB \citep{srinivas2010ucb}, Entropy Search \citep{hennig2012entropy,hernandez2014predictive,wang2017max} and Knowledge Gradient \citep{frazier2009knowledge}.
\begin{figure}
	\centering
	\includegraphics[width=\linewidth]{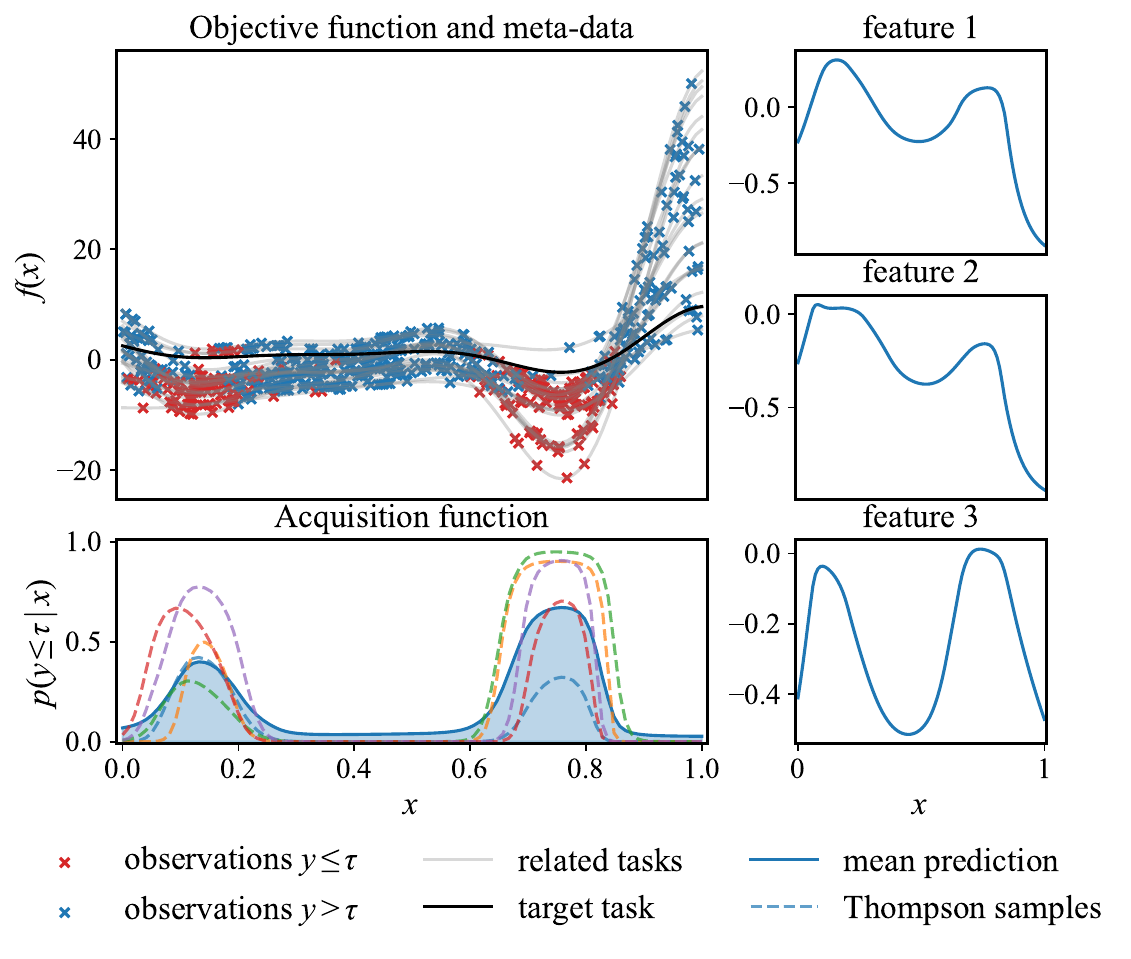}
	\caption{
		Meta-learning the acquisition function. Left: The top panel shows observations from 10 related tasks and the target task. The top performing observations ($\tau = \Phi^{-1}(\gamma), \gamma=1/3$) in each task are shown in \textcolor[HTML]{D62728}{red}, the rest in \textcolor[HTML]{1F77B4}{blue}. The bottom panel shows the maximum-a-posteriori estimate of the acquisition function in solid blue while the Thompson samples are shown as dashed curves. Right: Features learned by our model. \ourmethod successfully identifies the promising areas in the input space, while the Thompson samples show variability in the meta-learned acquisition function.
	}
	\label{fig:meta_forrester}
\end{figure}

\paragraph{Likelihood-free acquisition functions}
\label{ssec:likelihood-free_af}

Likelihood-free acquisition functions model the utility of a query without explicitly modeling the predictive distribution. For example, tree-structured Parzen estimators (TPE) \citep{bergstra2011algorithms} dismiss the surrogate for the outcomes and instead model two densities that split the observations w.r.t.\ a threshold $\tau$, namely
$\ell(\mathbf{x}) = p(\mathbf{x} \mid y \leq \tau, \mathcal{D}_N)$
and
$g(\mathbf{x}) = p(\mathbf{x} \mid y > \tau, \mathcal{D}_N)$ for the promising and non-promising data distributions, respectively.
The threshold $\tau$ relates to the $\gamma$-th quantile of the observed outcomes via $\gamma = \Phi(\tau) := p(y \leq \tau)$. In fact, the resulting density ratio (DR) $\alpha^{\text{DR}}(\mathbf{x}; \mathcal{D}_N, \tau) = \ell(\mathbf{x})/g(\mathbf{x})$ is shown to have the same maximum as PI \citep{song2022general,garnett_bayesoptbook_2022}.

BORE \citep{tiao2021bore} improves several aspects of TPE by directly estimating the density ratio instead of solving the more challenging problem of modeling two independent densities as an intermediate step. It rephrases the density ratio estimation as a binary classification problem where all observations within the same class have the same importance. Specifically, they show $\alpha^{\text{DR}}(\mathbf{x}; \mathcal{D}_N, \tau) \propto C_{\bm\theta}(\mathbf{x}) = p(k=1 \mid \mathbf x, D_N, \tau)$, where $k=\mathbbm{1}(y\leq\tau)$ represents the binary class labels for classification and the classifier $C_{\bm\theta}$ has learnable parameters $\bm\theta$.

Likelihood-free Bayesian optimization (\lfbo) \citep{song2022general} directly learns an acquisition function in the form of \cref{eq:expected_utility} through a classifier. By rephrasing the integral as a variational problem, \lfbo involves solving a weighted classification problem with noisy labels for the class $k=1$, where the weights correspond to utilities. It is shown that the EI acquisition function, where $U(y; \tau) := \max (\tau - y, 0)$, can be estimated by a classifier that optimizes the following objective:
\begin{equation}
\label{eq:lfbo_loss}
\begin{split}
    &\mathcal{L}^{\lfbo} (\bm{\theta}; \mathcal{D}_N, \tau) =\\ 
    &- {\mathbb{E}}_{(\mathbf{x}, y) \sim \mathcal{D}_{N}} \big[\max(\tau - y, 0) \ln C_{\bm{\theta}}(\mathbf{x}) +  \ln (1 - C_{\bm{\theta}}(\mathbf{x})) \big].
\end{split}
\end{equation}
The resulting classifier splits promising and non-promising configurations with probabilistic predictions that can be interpreted as the utility of queries, leading to scale-invariant models without noise assumptions and allowing the application of any classification methods \citep{song2022general}. Further details of the algorithms are provided in \cref{appendix:likelihood_free}.

\section{Methodology}
\label{sec:methodology}
In this section, we introduce our MetA-learning for LIkelihood-free BO (\ourmethod) method, which extends \lfbo with an effective meta-learning approach. An illustration of our method on a one-dimensional problem is shown in \cref{fig:meta_forrester}. Our approach uses a neural network to meta-learn both a task-agnostic model based on features learned across tasks (right panel in \cref{fig:meta_forrester}), and a task-specific component that provides uncertainty estimation to adapt to new tasks. Additionally, we use Thompson sampling (dashed lines in \cref{fig:meta_forrester}) as an exploratory strategy to account for the task uncertainty. Finally, a residual prediction model (see below) is added to adapt to tasks that are not well captured by the meta-learned model.
\begin{figure}
    \centering
    \includegraphics[width=.8\linewidth]{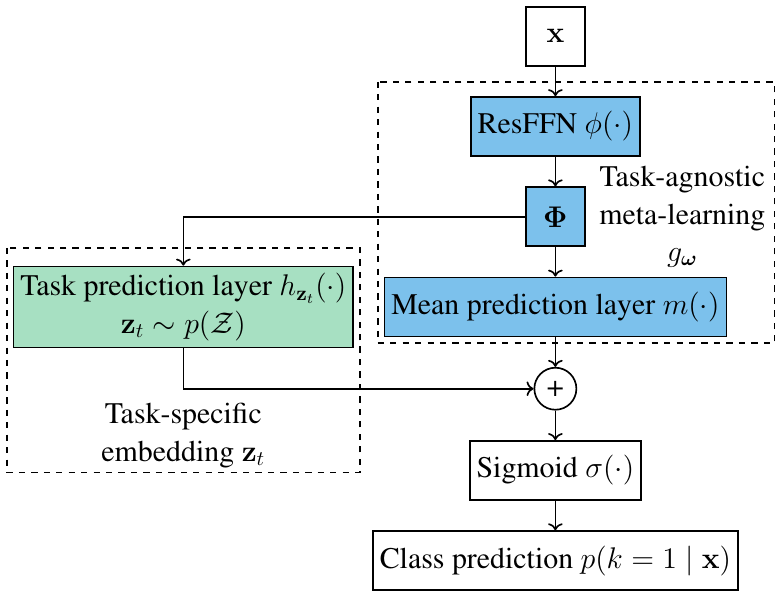}
    \caption{Schematic representation of our meta-learning classifier. A residual feedfoward network (ResFFN) maps the input $\mathbf{x}$ via a shared feature mapping function $\phi$. From this, we construct a task-agnostic mean prediction $m(\bm\Phi)$ and a task embedding $\mathbf{z}_t$, which is distributed according to a prior distribution $p(\mathcal Z)$. The feature mapping function $\phi$ and mean prediction layer $m$ are fixed after meta-training, denoted by the task-agnostic component $g_{\bm \omega}$. Finally, we add and convert them to a class prediction via the sigmoid function.}
    \label{fig:malibo_scheme}
\end{figure}
\paragraph{Network structure}
\ourmethod uses a structured neural network that combines a meta-learned, task-agnostic model with a task-specific layer. We show an overview in \cref{fig:malibo_scheme} and provide details for the choices below. 
Following previous works \citep{perrone2018scalable,berkenkamp2021probabilistic}, our meta-learning model uses a deterministic, task-agnostic model to map the input into features $\bm\Phi=\phi(\mathbf x)$, where $\phi: \mathcal{X} \rightarrow \mathbb{R}^d$ is a learnable feature mapping shared across all tasks and $d$ is the predefined dimensionality of the feature space. 
We use a residual feedforward network (ResFFN) for learning $\phi$, which has been shown to be robust to network hyperparameters and generalizes well to different problems \citep{huang2020residual}. 
To enable our model to provide good initial proposals, we introduce a task-agnostic mean prediction layer $m: \mathbb{R}^d \rightarrow \mathbb{R}$ that learns the promising areas from the related tasks.
We refer to the combined task-agnostic components $m$ and $\phi$ as $g_{\bm \omega}$ (shown in blue), which is parameterized by $\bm \omega$. 
To allow adaptation on each task $t$, we use a task prediction layer $h_{\mathbf{z}_t}: \mathbb R^d \rightarrow \mathbb R$, which is parameterized by layer weights $\mathbf{z}_t \in \mathcal{Z} \subseteq \mathbb{R}^d$. Since each $\mathbf z_t$ embeds in a low dimensional latent space $\mathcal Z$ and is a unique vector for each task, we refer to $\mathbf z_t$ as the task embedding.
We will train our model such that the $\lbrace\mathbf z_t \rbrace_{t=1}^T$ follow a known distribution $p(\mathcal Z)$  and discuss below how to use this as a prior for target task adaptation.
Lastly, in order to obtain classification outputs as in \lfbo, we apply the sigmoid function to produce probabilistic class predictions $p(k=1 \mid \mathbf x)$. The prediction for an observation in task $t$ is then given by $C(\mathbf x_t) = \sigma(m(\phi(\mathbf x)) + h_{\mathbf{z}_t}(\phi(\mathbf x))) = \sigma(m(\bm\Phi) + \mathbf{z}_t^\mathsf{T} \bm\Phi)$.

\paragraph{Meta-learning}
Directly optimizing $\mathcal{L}^{\text{\lfbo}}$ to meta-learn our model would lead to task embeddings that do not conform to any particular prior task distribution $p(\mathcal Z)$, and thus render task adaptation difficult and unreliable \citep{finn2018probabilistic}. Therefore, we regularize the task embeddings $\lbrace\mathbf z_t \rbrace_{t=1}^T$ during training to enable Bayesian inference. In addition, such regularization can also avoid overfitting in the task space $\mathcal{Z}$ and improves the generalization performance of our model. Specifically, we assume the prior of the task embeddings to be a multivariate normal (MVN), $p(\mathcal Z) = \mathcal{N}(\mathbf{0},\mathbf{I})$ and apply a regularization term to bring the empirical distribution of the $\lbrace\mathbf z_t \rbrace_{t=1}^T$ close to the prior distribution. The loss used for training on the meta-data reads:
\begin{multline}
\label{eq:negative_log_likelihood}
    \mathcal{L}^{\text{meta}} (\bm \omega, \{ \mathbf{z}_t \}_{t=1}^T) = \frac{1}{T} \sum_{t=1}^{T} \mathcal{L}^{\lfbo}(\bm \omega, \mathbf{z}_t ; \mathcal{D}^t, \tau)\\ + \lambda \mathcal{R}(\{ \mathbf{z}_t \}_{t=1}^T ; p(\mathcal Z))\,,
\end{multline}
where the first term is the loss function from \lfbo as in \cref{eq:lfbo_loss}, weighting the observations in the meta-data with improvements and the second term $\mathcal{R}$ is the regularization term weighted by $\lambda$. We regularize the empirical distribution of $\lbrace\mathbf z_t \rbrace_{t=1}^T$ to match the Gaussian prior in a tractable way \citep{saseendran2021shape}:
\begin{multline}
\label{eq:regularization}
    \mathcal{R}(\{ \mathbf{z}_t \}_{t=1}^T ; p(\mathcal Z))
    = \lambda_{\text{KS}} \sum^d_{j=1} (F([\mathbf{z}_t]_j) - \Phi([\mathbf{z}_t]_j))^2 \\
    + \lambda_{\text{Cov}} \| \mathbf{I} - \text{Cov}(\lbrace\mathbf z_t \rbrace_{t=1}^T) \|^2_{\mathrm{F}}\,,
\end{multline}
where the first term matches the marginal cumulative distribution functions (CDFs) similar to a Kolmogrov-Smirnov (KS) test, and the second term matches the empirical covariance of the task embeddings to the covariance of the prior. The hyperparameters $\lambda_{\text{KS}}$ and $\lambda_{\text{Cov}}$ encode the trade-off between these two terms. We denote $F$ as the empirical CDF and $\text{Cov}$ as the empirical covariance matrix. For more details we refer to \cref{sec:latent_reg}.

We only consider a uni-modal Gaussian prior in this work, as we will show it already demonstrates strong performance against other baselines. For more complex task distributions, one could extend it with multi-modal Gaussian prior \citep{saseendran2021shape}.

\paragraph{Task adaptation}
\label{ssec:task_adaptation}

After meta-training, the model can adapt to new tasks by estimating an embedding $\mathbf{z}$ based on the learned feature mapping function $\phi$. In principle, one could use a maximum likelihood classifier obtained by directly optimizing \cref{eq:lfbo_loss} w.r.t.\ $\mathbf{z}$. However, such a classifier does not consider the task uncertainty and would suffer from unreliable adaptation \citep{finn2018probabilistic} and over-exploitation \citep{oliveira2022batch}.
Furthermore, when a potential disparity between the distribution of the meta-data and the non-i.i.d.\ data collected during optimization arises, a probabilistic model would be informed via uncertainty estimation and thereby can exploit the meta-learned knowledge less. Therefore, we propose to use a Bayesian approach for task adaptation, which makes our classifier uncertainty-aware and more exploratory.

Consider the task embedding $\mathbf{z}$ for the target task follows a distribution $p(\mathbf{z} \mid \mathcal{D}_N)$ after $N$ observations, then the predictive distribution of our model can be written as
\begin{equation}
\label{eq:predictive_p}
    C(\mathbf x;\bm \omega, \mathcal{D}_N) = \int p(k=1 \mid \bm \omega, \mathbf{z}) p(\mathbf{z} \mid \mathcal{D}_N)\,\mathrm{d}\mathbf{z}\,,
\end{equation}
which accounts for the epistemic uncertainty in the task embedding. Since the parameters $\bm \omega$ of task-agnostic model $g_{\bm{\omega}}$ are fixed after meta-training, we denote our classifier as $C(\mathbf x)$ for simplicity.

As there is no analytical way to evaluate the integration in \cref{eq:predictive_p}, we have to resort to approximation methods, such as Laplace approximation \citep{bishop2006pattern}, variational inference \citep{graves2011variational}, 
and Markov chain Monte Carlo \citep{homan2014nuts}. We consider Laplace approximation for $p(\mathbf{z} \mid \mathcal{D}_N)$ as a fast and scalable method, and show its competitive performance against other, more expensive alternatives in \cref{ssec:inference_method}.

Laplace's method fits a Gaussian distribution around the maximum-a-posteriori (MAP) estimate of the distribution and matches the second order derivative at the optimum. In the first step, we obtain the MAP estimate by maximizing the posterior of our classifier $C$ parameterized by $\mathbf{z}$. 
To be consistent with the regularization used during meta-training, we use a standard, isotropic Gaussian prior for the weights: $p(\mathbf{z}) = \mathcal{N}(\mathbf{z}\mid \mathbf{0}, \mathbf{I})$. Given observations $\mathcal{D}_N$, the negative log posterior $p(\mathbf{z} \mid \mathcal{D}_N)$ is proportional to
\begin{multline}
\label{eq:log_posterior}
	\mathcal{L}^{\ourmethod}(\mathbf{z}) = \frac{1}{2} \mathbf{z}^\mathsf{T} \mathbf{z} - \\ \sum^{N}_{n=1} \left( k_n (\tau - y) \ln \hat{k}_n + \ln(1 - \hat{k}_n) \right)\,,
\end{multline}
where $\hat{k} = \sigma(m(\bm{\Phi}) + \mathbf{z}^\mathsf{T}\bm{\Phi})$ is the class probability prediction and the MAP estimate of the weights given by $\mathbf{z}_{\text{MAP}} = {\arg \min}_{\mathbf{z} \in \mathcal{Z}} \mathcal{L}^{\ourmethod}$. As a second step, we compute the negative Hessian of the log posterior
\begin{equation}
\label{eq:covariance_update}
    \mathbf{\Sigma}_N^{-1} 
    = \mathbf{\Sigma}_0^{-1} + \sum_{n=1}^{N} (k_n (\tau - y) + 1) \hat{k}_n(1 - \hat{k}_n)\bm{\Phi}_{n}\bm{\Phi}_{n}^\mathsf{T}\,,
\end{equation}
which serves as the precision matrix for the approximated posterior $q(\mathbf{z}) = \mathcal{N}(\mathbf{z}\mid\mathbf{z}_{\text{MAP}}, \mathbf{\Sigma}_N)$. Therefore \cref{eq:predictive_p} can be approximated as
\begin{equation}
\label{eq:predictive_q}
    C(\mathbf{x}) \simeq \int p(k=1 \mid \bm \omega, \mathbf{z}) q(\mathbf{z})\,\mathrm{d}\mathbf{z}\,.
\end{equation}
Having developed a meta-learning model, we now focus on how to utilize this model to encourage exploration and ensure reliable task adaptation.

\paragraph{Uncertainty-based exploration}
In the early phase of optimization, every meta-learning model has to reason about the target task properties based only on the limited data available, which can lead to highly biased results and over-exploitation \citep{finn2018probabilistic}. Moreover, \lfbo also suffers from similar issue even without meta-learning \citep{song2022general}. Therefore, we propose to use Thompson sampling based on task uncertainty for constructing a more exploratory acquisition function, and the resulting sampled predictions is generated by
\begin{equation}
\label{eq:predictive}
    \hat{C}(\mathbf{x})
    = \sigma \left( m(\phi(\mathbf x)) + h_{\hat{\mathbf z}}(\phi(\mathbf x)) \right), \quad \hat{\mathbf{z}} \sim q(\mathbf{z})\,.
\end{equation}
Besides stronger exploration in the early phases of optimization, Thompson sampling also enables us to extend \ourmethod to parallel BO by using multiple Thompson samples of the acquisition function in parallel. It is shown that this bypasses the sequential scheme of traditional BO, without introducing the common computational burden of more sophisticated methods \citep{kandasamy2018parallelised}. We
briefly explore this strategy in \cref{appendix:step-through-vis}.

\paragraph{Gradient boosting as a residual prediction model}
Operating in a meta-learned feature space enables fast task adaptation for our Bayesian classifier. However, it relies on the assumption that the meta-data is sufficient and representative for the task distribution, which does not always hold in practice. Moreover, a distribution mismatch between observations $\mathcal{D}_N$ and meta-data $\mathcal{D}^{\text{meta}}$ can arise when $\mathcal{D}_N$ is generated by an optimization process while $\mathcal{D}^{\text{meta}}$ consists of, \eg i.i.d.\ samples.

We employ a residual model independent of the meta-learning model, such that, even given non-informative features, our classifier is able to regress to an optimizer that operates in the input space $\mathcal{X}$. We propose to use gradient boosting (GB) \citep{friedman2001boosting} as a residual prediction model for classification, which consists of an ensemble of weak learners that are sequentially trained to correct the errors from the previous ones.
Specifically, we replace the first weak learner by a strong learner, \ie our meta-learned classifier. With Thompson sampling, our classifier can be written as
\begin{equation}
\label{eq:gradient_boosting}
    C_{\text{GB}}(\mathbf{x}) =\sigma \left( m(\phi(\mathbf x)) + h_{\hat{\mathbf z}}(\phi(\mathbf x)) + \sum^{M}_{i=1} r_i(\mathbf{x}) \right)\,,
\end{equation}
where each $r_i$ represents the $i$-th trained base-learner for the error correction from gradient boosting. In addition to robust task adaptation, this approach offers two advantages: First, gradient boosting does not require an additional weighting scheme for combining different classifiers and automatically determines the weight of the meta-learned model; Second, gradient boosting demonstrates strong performance for \lfbo on various benchmarks \citep{song2022general}, which makes our classifier achieve competitive performance even when meta-learning fails, as shown in \cref{ssec:ablation_gradient_boosting}. 
The resulting residual model is trained solely on the target task data
and thus might overfit in the early iterations with limited data. To avoid this, we apply gradient boosting only after a few iterations of Thompson sampling exploration and train it with early stopping. Note that this does not diminish the usefulness of the residual model, because our goal is to encourage exploration in early iterations as outlined in \cref{ssec:task_adaptation}, and gradually rely more on the knowledge from the target task.
We refer to \cref{appendix:experimental_deatails} for more implementation details.

\begin{algorithm}[tb]
\caption{\ourmethod: Meta-learning for likelihood-free Bayesian optimization}
\label{alg:malibo_algo}

\textbf{\quad Meta-learning}:

\textbf{\qquad Input}: $\mathcal{D}^{\text{meta}} = \{\mathcal{D}^t\}_{t=1}^T$\,, proportion $\gamma \in (0, 1)$

\begin{algorithmic}[1]
\STATE \ $k = \mathbbm{1}(y \leq \tau)$, where $\tau = \Phi^{-1}(\gamma)$\\
\tcc{\footnotesize \textcolor{gray}{generate binary labels}}
\STATE \ $g_{\bm \omega} \leftarrow \argmin_{\bm \omega} \mathcal{L}^{\text{meta}}$\tcp*[r]{\footnotesize \textcolor{gray}{\cref{eq:negative_log_likelihood}}}
\end{algorithmic}

\textbf{\quad Bayesian optimization with meta-learning}:

\textbf{\qquad Input}: Fixed $g_{\bm \omega}$ after meta-leaning

\begin{algorithmic}[1]
\STATE \ $\mathbf x_0 \leftarrow \argmax_{\mathbf x}g_{\bm \omega}(\mathbf x)$\,
\STATE \ $\mathcal{D} \leftarrow \{ (\mathbf{x}_0, f(\bm{x}_0) + \epsilon) \}$\,
\WHILE{\textit{has budget}}
\STATE $\mathbf{z}_{\text{MAP}} \leftarrow \argmin_{\mathbf  z} \mathcal{L}^{\ourmethod}$\tcp*[r]{\footnotesize\textcolor{gray}{\cref{eq:log_posterior}}}
\STATE Update precision matrix $\mathbf{\Sigma}_N^{-1}$\tcp*[r]{\footnotesize{\textcolor{gray}{\cref{eq:covariance_update}}}}
\STATE $\hat{\mathbf{z}} \sim \text{MVN}(\mathbf{z}_{\text{MAP}}, \mathbf{\Sigma}_N)$\;
\STATE $\mathbf{x}_{*} \leftarrow \arg \max_{\mathbf{x}} C_{\text{GB}}(\mathbf{x};{\hat{\mathbf{z}}})$\tcp*{\footnotesize \textcolor{gray}{\cref{eq:gradient_boosting}}}
\STATE $\mathcal{D} \leftarrow \mathcal{D} \cup \{ (\mathbf{x}_{*}, f(\mathbf{x}_{*}) + \epsilon) \}$\,
\ENDWHILE
\end{algorithmic}
\end{algorithm}

\section{Experiments}
\label{sec:experiments}
In this section, we first show the effects of using Thompson sampling and gradient boosting through a preliminary ablation study. Subsequently, we describe the experiments conducted to empirically evaluate our method. For the choice of problems, we focus on automated machine learning (AutoML), \ie hyperparameter optimization (HPO) and neural architecture search (NAS). To highlight the time efficiency of our proposed method, we include a runtime analysis. Additionally, a quantitative ablation study is presented to assess the impact of various components within our framework. Lastly, we evaluate our method on synthetic functions with multiplicative noise to study robustness towards data with heterogeneous scale and noise,

\paragraph{Baselines} We compare our method against multiple baselines across all problems. As methods without meta-learning, we pick random search \citep{bergstra2012random},
\lfbo \citep{song2022general} and Gaussian process (GP) \citep{snoek2012practical}) for our experiments. For meta-learning BO methods, we choose ABLR \citep{perrone2018scalable}, BaNNER \citep{berkenkamp2021probabilistic}, RGPE \citep{feurer2022practical}, GC3P \citep{salinas2020quantile}, FSBO \citep{wistuba2021fewshot}, MetaBO \citep{volpp2020metalearning}, PFN \cite{muller2023pfns4bo} and DRE \citep{khazi2023deep} as representative algorithms. Additionally, we consider a simple baseline for extending \lfbo with meta-learning, called \lfbobb, which combines \lfbo with bounding-box search space pruning \citep{perrone2019searchspace} as a meta-learning approach. For all \lfbo-based methods, including \ourmethod, we set the required threshold $\gamma = 1/3$ following \citet{song2022general}.

\paragraph{Evaluation metrics} In order to aggregate performances across tasks, we use \textit{normalized regret} as the quantitative performance measure for AutoML problems \citep{wistuba2018scalable}. This is defined as $\min_{\mathbf x \in \mathcal X_N} (f^t(\mathbf x) - f^t_{\text{min}}) / (f^t_{\text{max}} - f^t_{\text{min}})$, where $\mathcal X_N$ denotes the set of inputs that have been selected by an optimizer up to iteration $N$, $f^t_{\text{min}}$ and $f^t_{\text{max}}$ respectively represent the minimum and the maximum objective computed across all offline evaluations available for task $t$. We report the mean normalized regret across all tasks within a benchmark as the aggregated result. For all benchmarks, we report the results by mean and standard error across \textit{100 random runs}.

\paragraph{Effects of exploration and residual prediction}
\begin{figure}
    \centering
    \includegraphics[width=\linewidth]{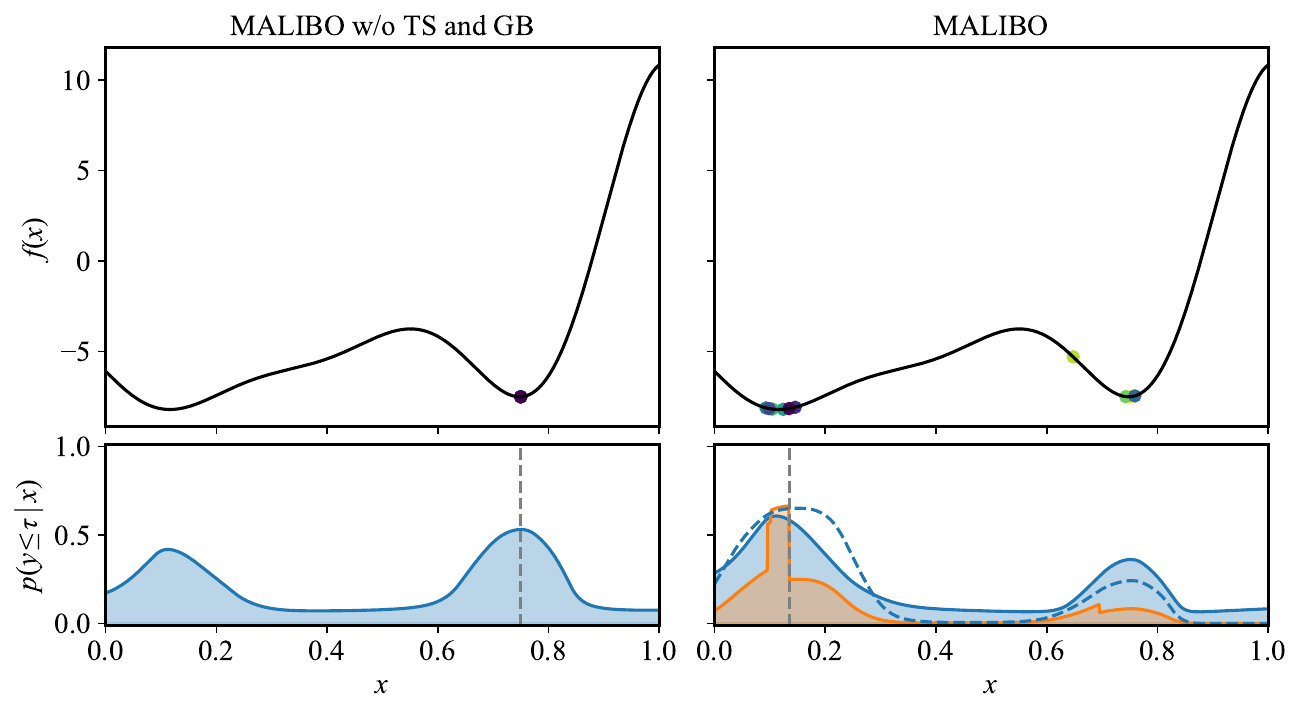}
    \caption{
        Effects of exploration and residual predictions. Color circles denote the optimization queries (from bright to dark), the dashed curve denotes a Thompson sample (TS) of the acquisition function and the orange curve shows the sample combined with gradient boosting (GB).
    }
    \label{fig:malibo_ablation}
\end{figure}
We first investigate the effect of Thompson sampling and the residual prediction model when optimizing a Forrester function \citep{sobester2008engineering} as a toy example. By using the meta-learned model as shown in \cref{fig:meta_forrester}, \ourmethod performs task adaptation on a new Forrester function for \num{10} iterations. We compare the results of \ourmethod against a variant without the proposed Thompson sampling and gradient boosting, which only uses the approximated posterior predictive distribution in \cref{eq:predictive_q} by probit approximation \citep{bishop2006pattern} for the acquisition function. As shown in \cref{fig:malibo_ablation}, \ourmethod without Thompson sampling fails to adapt the new task with little exploration and optimizes greedily around the local optimum. This greedy optimization occurs due to the strong dependence of \lfbo on a good initialization to not over-exploit.
In contrast, the proposed \ourmethod allows the queries to cover both possible optima by encouraging explorations. In addition, gradient boosting performs the refinement beyond the smooth meta-learned acquisition function, which can be seen in the discontinuity in the predictions. By suppressing the predicted utility in the less promising areas, gradient boosting refines the acquisition function to focus on the lower value region. We provide an extensive ablation study on the effects of different components in \ourmethod and refer to \cref{appendix:ablation} for more details.

\paragraph{Runtime analysis}
\begin{figure}[t]
    \centering
    \includegraphics[width=\linewidth]{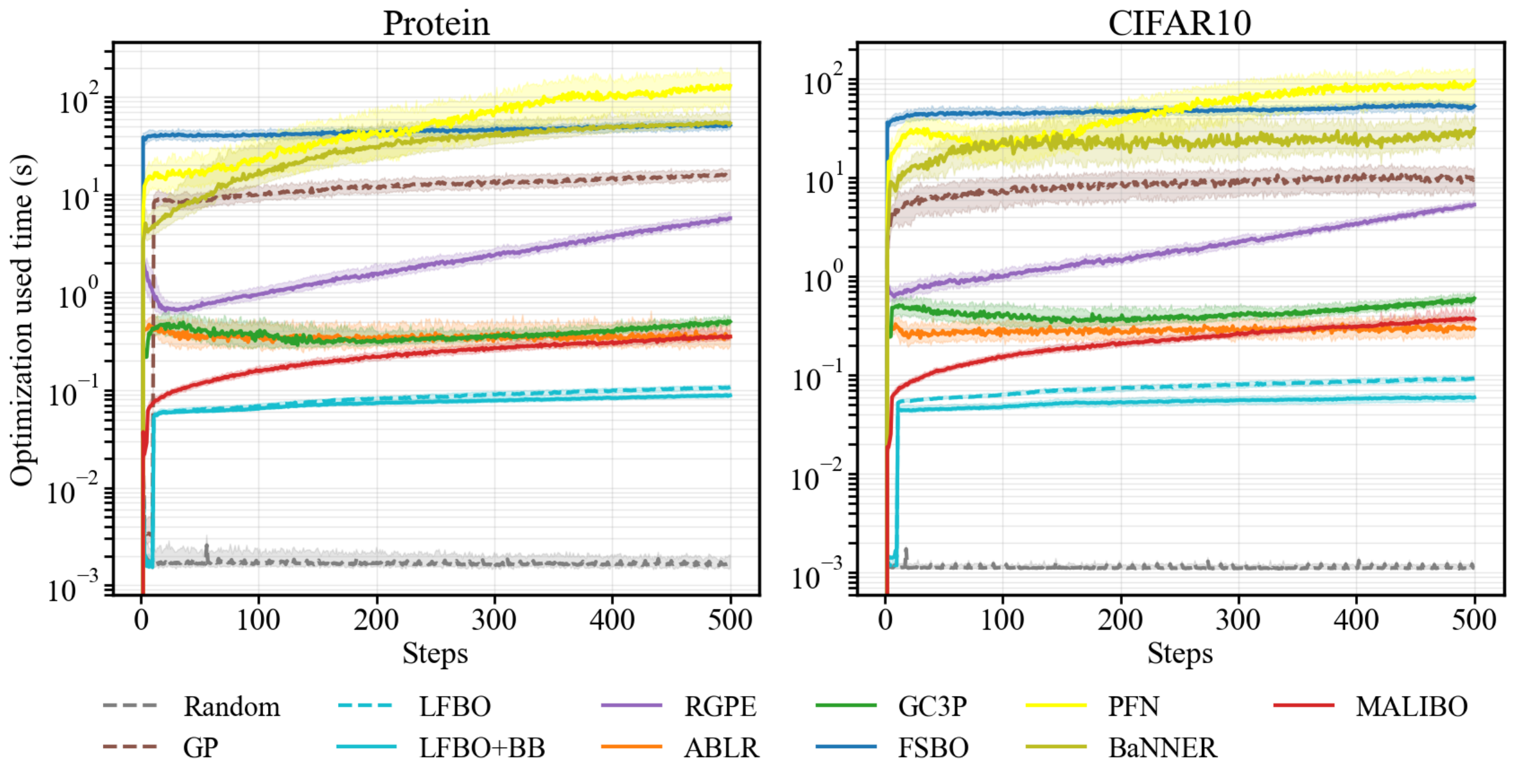}
	\caption{Runtime of different BO algorithms over optimization steps. We show the typical results for two benchmarks and plot the medial inter-quantiles to remove outliers.}
	\label{fig:opt_time}
\end{figure}
To confirm the scalability of \ourmethod, we compare the runtime between methods, specifically the time required for the algorithm to propose a new candidate. As shown in \cref{fig:opt_time}, the introduction of latent features and the Laplace approximation only adds negligible overhead compared to \lfbo, while \ourmethod's runtime increases slowly with the number of observations. In contrast, all other meta-learning methods, except for \lfbobb, are considerably slower than \ourmethod. We include more detailed experimental results in \cref{ssec:runtime_analysis}.

\paragraph{Real-world benchmarks}
\begin{figure*}[t]
    \centering
    \includegraphics[width=\linewidth]{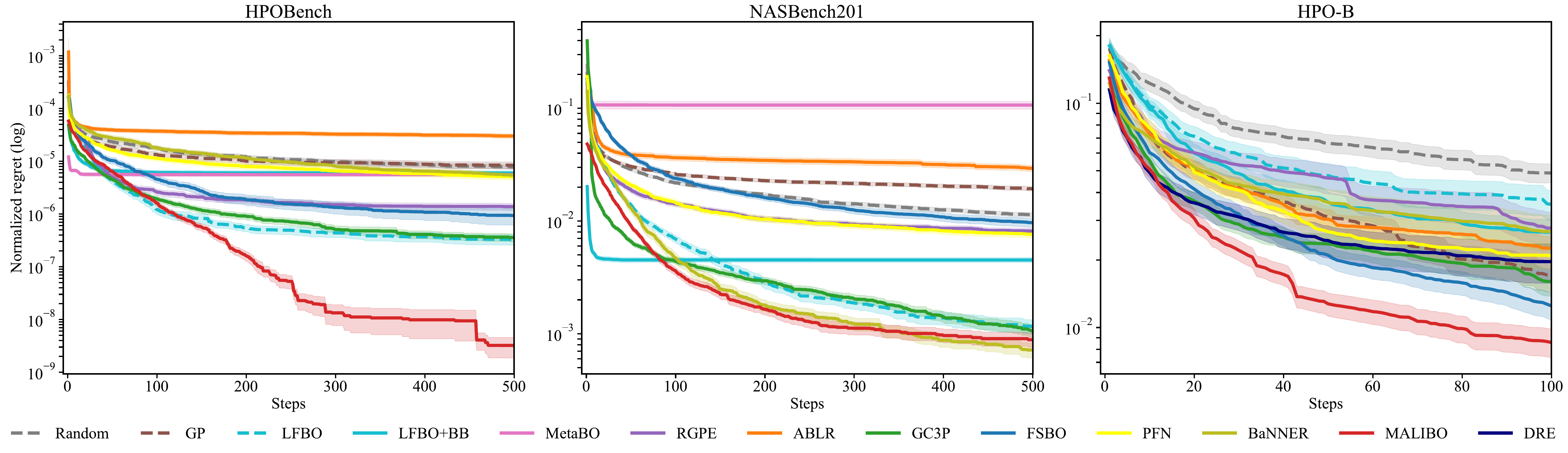}
	\caption{Aggregated normalized regrets for BO algorithms on real-world AutoML problems.}
	\label{fig:benchmark}
\end{figure*}
We empirically evaluate our method on various real-world optimization tasks, focusing on AutoML problems, including neural architecture search (NASBench201) \citep{dong2020nasbench}, hyperparameter optimization for neural networks (HPOBench) \citep{klein2019tabular} and machine learning algorithms (HPO-B) \citep{pineda2021hpob}. In NASBench201, we consider designing a neural cell with \num{6} discrete parameters, totaling $15,625$ unique architectures, evaluated on CIFAR-10, CIFAR-100 \citep{krizhevsky2009learning} and ImageNet-16 \citep{chrabaszcz2017downsampled}. The goal is to find the optimal architecture for a neural network that yields the highest validation accuracy. For HPOBench, the aim is to find the optimal hyperparameters for a two-layer feed-forward regression network on four popular UCI datasets \citep{dua2019uci}. The search space is \num{9}-dimensional and the optimization objective is the validation mean squared error after training with the corresponding network configuration. In HPO-B, the focus is on optimizing the hyperparameters for different machine learning models to maximize accuracy across various tasks. This benchmark comprises about \num{6} million evaluations of hyperparameters, across \num{16} search spaces that correspond to different machine learning models. Each search space varies in dimensionality ranging from \num{2} to \num{18} and includes several tasks, which are divided into training, validation and test tasks. Compared to the extensive evaluations in HPO-B, both HPOBench and NASBench201 have significantly fewer related tasks and serve as examples of performance with limited meta-data. We provide details for benchmarks in \cref{appendix:benchmark_details}.

To train and evaluate the meta-learning BO methods in HPOBench and NASBench201, we conduct our experiments in a leave-one-task-out way: all meta-learning methods use one task as the target task and all others as related tasks. In this way, every task in a benchmark is picked as the target task once. To construct the meta-datasets, we randomly select \num{512} configuration-objective pairs from the related tasks, considering the limitations of RGPE in handling large meta-datasets. All meta-learning methods, except MetaBO, are trained from scratch for each independent run, to account for variations due to the randomly sampled meta-data. Because of its long training time, MetaBO is trained once for each target problem on more meta-data than other methods to avoid limiting its performance with a bad subsample and we show its results only for HPOBench and NASBench201. As for HPO-B, we utilize the provided meta-train and meta-validation dataset to train the meta-learning methods and evaluate all methods on the meta-test data. While all methods optimize the target tasks from scratch in the other two benchmarks, the first five initial observations in HPO-B is fixed as random seed and therefore we only show the performances starting after the initialization. We refer to \cref{appendix:experimental_deatails} for more experimental details.

The aggregated results for all three benchmarks are summarized in \cref{fig:benchmark}. It is evident that \ourmethod consistently achieves strong anytime performance, surpassing other methods that either exhibit poor warm-starting or experience early saturation of performance. Notably, \ourmethod outperforms other methods by a large margin in HPOBench, possibly because we focus on minimizing the validation error of a regression model in this benchmark. This task poses a significant challenge for GP-based regression models, as the observation values undergo abrupt changes and have varying scales across tasks, thereby violating the smoothness and noise assumptions inherent in these models. In most benchmarks, PFN and BaNNER exhibit comparable performance to GP, except in NASBench201. GC3P performs competitively only after the Copula process is fitted and \lfbo matches its final performance. \lfbobb exhibits similar performance as \ourmethod in warm-starting and converges quickly, but the search space pruning technique forbids the method to explore regions beyond the promising areas in the meta-data, making its final performance even worse than its non-meta-learning counterpart. ABLR, RGPE and FSBO perform poorly on most of the benchmarks, except for HPO-B, because their meta-learning techniques require more meta-data for effective warm-starting, making them less data-efficient than \ourmethod. MetaBO shows strong warm-starting performance in HPOBench while it fails in NASBench201. This is possibly due to the higher diversity in NASBench201 compared to HPOBench, and MetaBO fails to transfer knowledge from tasks that are significantly different from the target task. The poor task adaptation ability of MetaBO is also found by other studies \citep{wistuba2021fewshot,wang2022pretraining}. For more experimental results, we refer to \cref{ssec:real_world_benchmarks}.

\begin{figure*}[t]
    \centering
    \includegraphics[width=.9\linewidth]{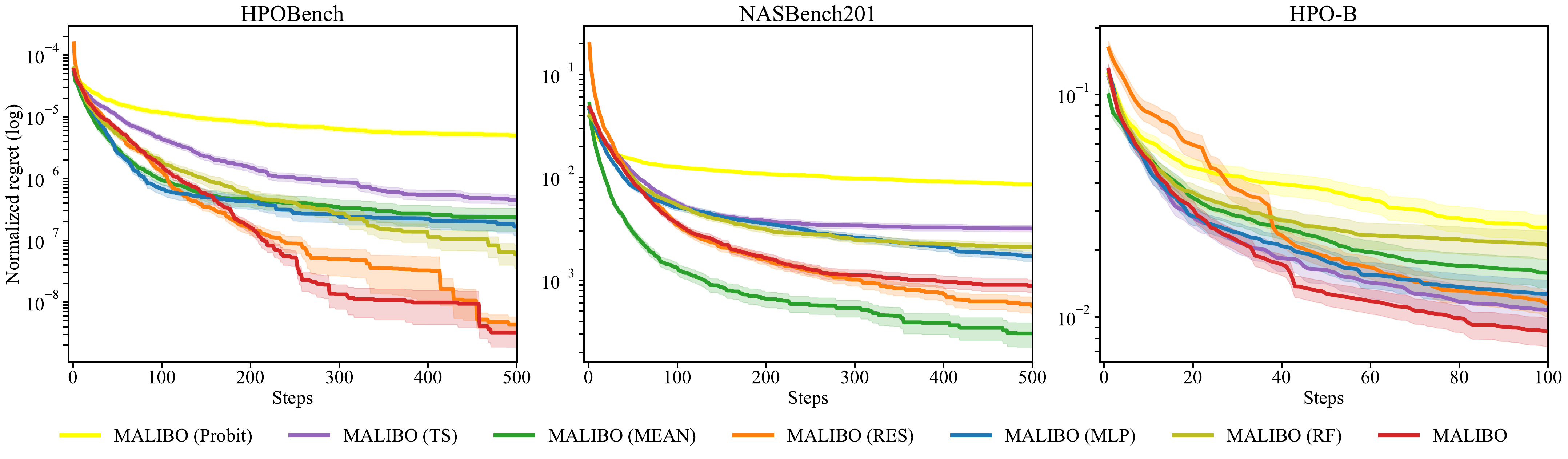}
	\caption{Aggregated normalized regrets of \ourmethod variants on real-world AutoML problems.}
	\label{fig:malibo_variant}
\end{figure*}

\paragraph{Ablation study}
To understand the impact of each component within MALIBO, we conduct a quantitative ablation study and introduce the following variants:
\begin{itemize}
    \item \ourmethod (Probit): Employs a probit approximation (detailed in \cref{sec:probit_approximation}) for the marginalized form of the acquisition function. This variant does not use gradient boosting.
    \item \ourmethod (TS): Utilizes only Thompson sampling, omitting the gradient boosting component.
    \item \ourmethod (RES): Excludes the mean prediction layer $m(\cdot)$.
    \item \ourmethod (MEAN): Removes the task prediction layer $h_{\mathbf z_t}(\cdot)$ and utilizes only the task-agnostic component $g_{\bm{\omega}}$.
    \item \ourmethod (RF): Substitutes gradient boosting with a random forest (RF) classifier.
    \item \ourmethod (MLP): Replaces gradient boosting with a multi-layer perceptron (MLP) classifier.
\end{itemize}
The results illustrated in \cref{fig:malibo_variant} reveal several key insights. Due to the lack of a mean prediction layer, \ourmethod (RES) exhibits the poorest warm-starting performance across all benchmarks, potentially leading to worse final performance. Although the mean prediction layer improves initial performance, relying solely on it may result in over-fitting to the meta-data due to insufficient exploration capabilities. As demonstrated by the performance of \ourmethod (MEAN), while it achieves the best results among all variants in NASBench201, its performance on the other two benchmarks is inferior to the proposed method. In contrast, variants that include an uncertainty-aware task prediction layer, such as \ourmethod (RES) and \ourmethod, perform task adaptation more reliably. Although \ourmethod (TS) also encourages exploration, the absence of a residual prediction model results in a significant performance decrease when the amount of meta-data is limited, as observed in HPOBench and NASBench201. The comparison of \ourmethod (MLP) and \ourmethod (RF) highlights the superiority of gradient boosting over other classifiers such as random forest and MLP for the residual prediction model. For more detailed experimental results, refer to \cref{ssec:quantitative_comparison}.

\paragraph{Robustness against heterogeneous noise}
\label{sssec:noise}
\begin{figure*}[t]
    \centering
    \includegraphics[width=.9\textwidth]{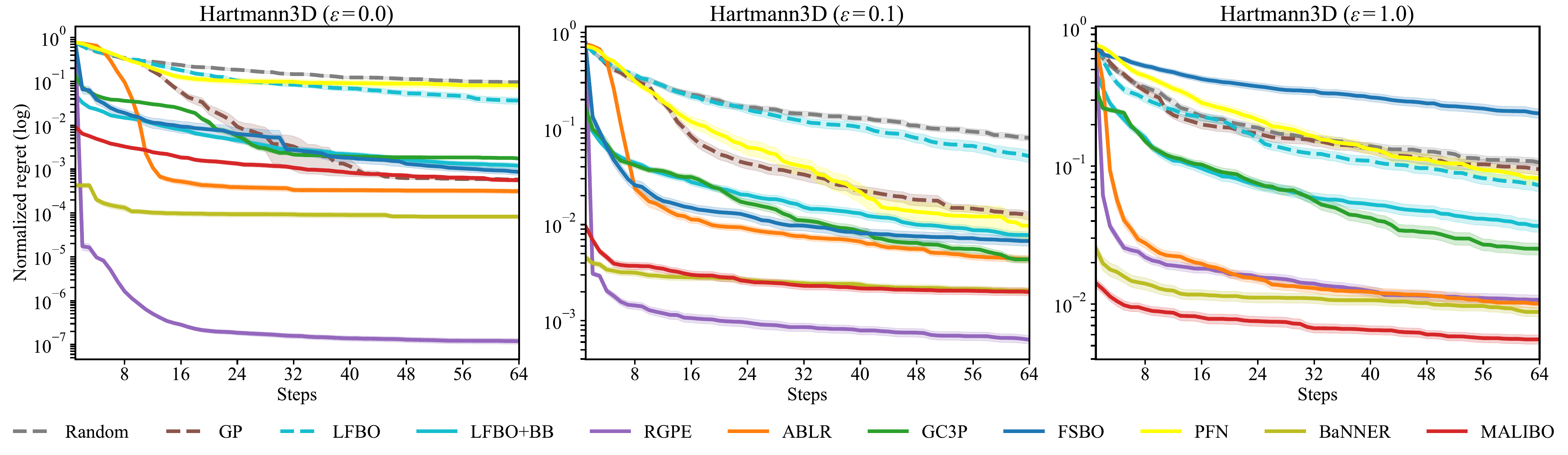}
    \caption{Normalized regret for BO algorithms on Hartmann3D with various levels of multiplicative noise.}
	\label{fig:synthetic_benchmark}
\end{figure*}
We use synthetic function ensembles \citep{berkenkamp2021probabilistic} to test the robustness against heterogeneous noise in the data. We focus on the Hartmann3D function ensemble \citep{dixon1978global}, which is a three-dimensional problem with four local minima. Their locations and the global minimum vary across different functions. See \cref{ssec:hartmann3} for more details.

To avoid biasing this experiment towards a single method, we use a heteroscedastic noise incompatible with any assumptions about the noise of any method. In particular, this violates the GP methods' and ABLR's assumption of homoscedastic, Gaussian noise. GC3P makes a similar assumption after the nonlinear transformation of the observation values, which does not translate to any well-known noise model. \lfbo, \lfbobb and \ourmethod make no explicit noise assumptions, but optimize for the best mean. We choose a multiplicative noise, \ie $y = f(\mathbf x) \cdot (1+\epsilon\cdot n)$, where $n\sim\mathcal{N}(\mathbf{0}, \mathbf{1})$. The noise corrupts observations with larger values more, while having a smaller effect on those with lower values. To see the robustness with different noise levels, we evaluate $\epsilon \in \{0, 0.1, 1.0 \}$. For meta-training, we randomly sample \num{512} noisy observations from \num{256} functions in the ensemble. We show our results in \cref{fig:synthetic_benchmark}, where we can see across all noise levels, our method learns a meaningful prior for the optimization. The GP-based methods, despite their strong performance in the noise-free case, especially RGPE, degrade significantly with increasing noise levels.

\section{Conclusion}
\label{sec:conclusion}
We introduced MetA-learning for LIkelihood-free BO (\ourmethod), a method that directly models the acquisition function from observations coupled with meta-learning. This method is computationally efficient and robust to heterogeneous scale and noise across tasks, which poses challenges for other methods. Furthermore, \ourmethod enhances data efficiency and incorporates a Bayesian classifier with Thompson sampling to account for task uncertainty, ensuring reliable task adaptation. For robust adaptation to tasks that are not captured by meta-learning, we integrate gradient boosting as a residual prediction model into our framework. Empirical results demonstrate the superior performance of the proposed method across various benchmarks.

Despite promising experimental results, some limitations of the method should be noted.
\begin{enumerate*}[label=(\roman*)]
	\item The exploitation and exploration parameter $\tau$ in likelihood-free BO algorithms could be treated more carefully, \eg via a probabilistic treatment \citep{tiao2021bore}.
    \item The regularization hyperparameter $\lambda$, while robust across our experiments, may lead to suboptimal outcomes in other scenarios.
    \item Using a uni-modal prior could be restrictive for more complex task distributions. Although a generalization to a Gaussian mixture model exists \citep{saseendran2021shape}, its efficacy within \ourmethod remains unverified.
\end{enumerate*}

\section*{Acknowledgments}
Robert Bosch GmbH is acknowledged for financial support. The authors acknowledges support from the European Research Council (ERC) under grant no. 952215 (TAILOR).

\section*{Impact Statement}
This paper presents work whose goal is to advance the field of Machine Learning. There are many potential societal consequences of our work, none which we feel must be specifically highlighted here.

\bibliography{references}
\bibliographystyle{icml2024}

\newpage
\appendix
\onecolumn



\section{Likelihood-free acquisition functions}
\label{appendix:likelihood_free}

For completeness, we provide the proofs and derivations for TPE \citep{bergstra2011algorithms}, BORE \citep{tiao2021bore}, and \lfbo \citep{song2022general}. Recall from \cref{eq:expected_utility} that the expected utility function is defined as the expectation of the improvement of the utility function $U(y; \tau)$ over the posterior predictive distribution. Given $N$ observations on the target task in a non-meta-learning setting, for the specific expected improvement (EI) acquisition function, where the utility function is $U(y; \tau) := \max(\tau - y, 0)$, the function reads:
\begin{equation}
\begin{split}
    \alpha^{U}(\mathbf{x}; \mathcal{D}_N, \tau) 
    &= \mathbb{E}_{p(y \mid \mathbf{x}, \mathcal{D}_N)} [U(y; \tau)] \\
    & = \int_{-\infty}^{\infty} U(y; \tau) p(y \mid \mathbf{x}, \mathcal{D}_N)\,\mathrm{d}y \\
    & = \int_{-\infty}^{\tau} (\tau - y) p(y \mid \mathbf{x}, \mathcal{D}_N)\,\mathrm{d}y \\
    & = \frac {\int_{-\infty}^{\tau} (\tau - y) p(\mathbf{x} \mid y, \mathcal{D}_N) p(y \mid \mathcal{D}_N)\,\mathrm{d}y} {p(\mathbf{x} \mid \mathcal{D}_N)}\,.
\end{split}
\end{equation}
We follow the prove from \citet{tiao2021bore} and consider $\ell(\mathbf{x}) = p(\mathbf{x} \mid y \leq \tau, \mathcal{D}_N)$ and $g(\mathbf{x}) = p(\mathbf{x} \mid y > \tau, \mathcal{D}_N)$. The denominator of the above equation can then be written as:
\begin{equation}
\begin{split}
\label{eq:ei_to_density_ratio}
    p(\mathbf{x} \mid \mathcal{D}_N) 
    & = \int_{-\infty}^{\infty} p(\mathbf{x} \mid y, \mathcal{D}_N) p(y \mid \mathcal{D}_N)\,\mathrm{d}y \\
    & = \ell(\mathbf{x}) \int_{-\infty}^{\tau} p(y \mid \mathcal{D}_N)\,\mathrm{d}y \\
    &\quad+ g(\mathbf{x}) \int_{\tau}^{\infty} p(y \mid \mathcal{D}_N)\,\mathrm{d}y \\
    & = \gamma \ell(\mathbf{x}) + (1 - \gamma) g(\mathbf{x})\,,
\end{split}
\end{equation}
where $\gamma = \Phi(\tau) \coloneqq p(y \leq \tau \mid \mathcal{D}_N)$. The numerator can be evaluated as:
\begin{align}
    \int^{\tau}_{-\infty} (\tau - y) p(\mathbf{x} \mid y, \mathcal{D}_N) p(y \mid \mathcal{D}_N)\,\mathrm{d}y
    & = \ell(\mathbf{x}) \int^{\tau}_{-\infty} (\tau - y) p(y \mid \mathcal{D}_N)\,\mathrm{d}y \label{eq:10} \\
    & = \ell(\mathbf{x}) \tau \int^{\tau}_{-\infty} p(y \mid \mathcal{D}_N)\,\mathrm{d}y  - \ell(\mathbf{x}) \int^{\tau}_{-\infty} y p(y \mid \mathcal{D}_N)\,\mathrm{d}y  \\
    & = \gamma \tau \ell(\mathbf{x}) - \ell(\mathbf{x}) \int^{\tau}_{-\infty} y p(y \mid \mathcal{D}_N)\,\mathrm{d} y \\
    & = K \cdot \ell(\mathbf{x})\,,
\end{align}
where $K = \gamma \tau - \int^{\tau}_{-\infty} y p(y \mid \mathcal{D}_N)\,\mathrm{d}y$. Therefore the EI acquisition function is equivalent to the $\gamma$-relative density ratio up to a constant $K$,
\begin{equation}
	\underbrace{\alpha(\mathbf{x}; \mathcal{D}_{N}, \tau)}_{\text{expected improvement}} \propto \underbrace{\frac{\ell(\mathbf{x})}{\gamma \ell(\mathbf{x}) + (1-\gamma)g(\mathbf{x})}}_{\gamma-\text{relative density ratio}}
\end{equation}
Intuitively, one can think of the configurations $\mathbf{x}$ with $y \leq \tau$ as \textit{good} configurations, and the those with $y > \tau$ as \textit{bad} configurations. Then the density ratio can be interpreted as the ratio between the model's prediction whether the configurations belong to the good or bad class.

The tree-structured Parzen estimator (TPE) \citep{bergstra2011algorithms} estimates this density ratio by explicitly modeling $\ell(\mathbf{x})$ and $g(\mathbf{x})$ using kernel density estimation for a fixed value of the hyperparameter $\gamma$. Within BORE \citep{tiao2021bore}, the density ratio is modeled by class probabilities, where $\ell(\mathbf{x}) = p(\mathbf{x} \mid y \leq \tau, \mathcal{D}_N)$ and $g = p(\mathbf{x} \mid y > 0, \mathcal{D}_N)$.

\citet{song2022general} proof that the density ratio acquisition functions are not always equivalent to EI. \citet{bergstra2011algorithms} and \citet{tiao2021bore} claim that \cref{eq:10} holds true by assuming $\ell(\mathbf{x})$ is independent of $y$ once $y \leq \tau$ and therefore can be treated as a constant inside the integral. In fact, $p(\mathbf{x}\mid y \leq \tau, \mathcal{D}_N)$ still depends on $y$ even if $y \leq \tau$, because it is a conditional probability conditioned on $y$ not $y \leq \tau$. Therefore, satisfying the condition $y \leq \tau$ does not imply independence of $y$. From the definition of conditional probability
\begin{equation}
	p(\mathbf{x}\mid y \leq \tau, \mathcal{D}_N) = \frac{\int_{-\infty}^{\tau} p(\mathbf{x}, y \mid\mathcal{D}_N)\,\mathrm{d}y}{\int_{-\infty}^{\tau} p(y \mid\mathcal{D}_N)\,\mathrm{d}y} \neq p(\mathbf{x}\mid y, \mathcal{D}_N)\,,
\end{equation}
we can see that they are not equivalent. Intuitively, the probability of the configuration $\mathbf{x}$ for a given $y$ value should still depend on $y$ even if $y < \tau$ holds. By making this independence assumption, the resulting density ratio acquisition function treats all $(\mathbf{x}, y)$ pairs below the threshold with equal probability (importance), when, in fact, EI weights the importance of $(\mathbf{x}, y)$ pairs by the utility $\max(\tau - y, 0)$

To tackle this issue, \citet{song2022general} propose to directly approximate EI inspired by the idea of variational f-divergence estimation \citep{nguyen2010divergence}. They provide a variational representation for the expected utility function at any point $\mathbf{x}$, provided samples from $p(y \mid \mathbf{x})$. Thereby, their approach replaces the potentially intractable integration with the variational objective function that can be solved based on samples:
\begin{equation}
    \mathbb{E}_{p(y \mid \mathbf{x})}[U(y; \tau)]= \argmax_{s \in [0, \infty)} \mathbb{E}_{p(y \mid \mathbf{x})}[U(y; \tau) f'(s)] - \\f^{\star} (f'(s))\,,
\end{equation}
where the utility function $U: \mathbb R \times \mathcal{T} \rightarrow [0, \infty)$ is non-negative, $\tau \in \mathcal{T}$, $f: [0, \infty) \rightarrow \mathbb{R}$ is a strictly convex function with third order derivatives, and $f^{\star}$ is the convex conjugate of $f$. The maximization is performed over $s \in [0, \infty)$ and it does not model distributions with probability but only samples from the observations $\mathcal{D}_N$.

They consider the approximated expected utility acquisition function as $\alpha^{\lfbo} = \hat{S}_{\mathcal{D}_{N, \tau}}(\mathbf{x})$, which can be written as:
\begin{equation}
\label{eq:lfbo_theorem}
    \hat{S}_{\mathcal{D}_{N, \tau}}(\mathbf{x}) = \argmax_{S: \mathcal{X} \rightarrow \mathbb R} \mathbb{E}_{\mathcal{D}_N}[U(y; \tau) f'(S(\mathbf{x})) - f^{\star} (f'(S(\mathbf{x}))].
\end{equation}
By optimizing a variational objective in the search space $\mathcal{X}$, the expected utility acquisition function over $\mathbf{x}$ can be recovered. For practical purpose, they choose a specific convex function $f$: $f(r) = r \log \frac{r}{r+1} + \log \frac{1}{r+1}$ for all $r > 0$, and a specific form of $S = C / (1 - C)$, where $C: \mathcal{X} \rightarrow (0, 1)$ and can be considered as a probabilistic classifier. By applying these into \cref{eq:lfbo_theorem}, the resulting acquisition function reads:
\begin{equation}
\label{eq:lfbo_af}
    \alpha^{\text{LFBO}}(\mathbf{x}; \mathcal{D}_{N}, \tau) = \hat{S}_{\mathcal D_N, \tau}(\mathbf x) = \hat{C}_{\mathcal{D}_N, \tau}(\mathbf{x}) / (1 - \hat{C}_{\mathcal{D}_N, \tau}(\mathbf{x})),
\end{equation}
where $\hat{C}_{\mathcal{D}_N, \tau}$ is the maximizer of an objective over $C$:
\begin{equation}
    \mathbb{E}_{(\mathbf{x}, y) \sim \mathcal{D}_{N}} [ U(y; \tau) \ln C(\mathbf{x}) + \ln (1 - C(\mathbf{x})) ]\,.
\end{equation}
This is can be reinterpreted as a classification loss with training examples weighted by the utility function.

\section{Probit approximation}
\label{sec:probit_approximation}
While one can sample the predictive posterior to make class prediction as in \cref{eq:predictive}, an alternative way is to approximate the integral in \cref{eq:predictive_q} via probit approximation. Let $a = m(\bm{\Phi}) + \mathbf{z}^\mathsf{T} \bm{\Phi}$ and $q(\mathbf{z}) = \mathcal{N}(\mathbf{z} \mid \mathbf{z}_{\text{MAP}}, \mathbf{\Sigma}_N)$ be the approximated posterior obtained through the Laplace approximation. The distribution of $a$ then follows the Gaussian $\mathcal{N}(a \mid \mu_a, \sigma_a^2)$ with the following parameters:
\begin{equation}
\label{eq:mean_laplace}
\begin{split}
    \mu_a = \mathbb{E}[a] &= \int p(a)a\, da \\
                          &= \int q(\mathbf{z})(m(\bm{\Phi}) + \mathbf{z}^\mathsf{T} \bm{\Phi})\, d\mathbf{z}\\
                          &= m(\bm{\Phi}) + \mathbf{z}_{\text{MAP}}^\mathsf{T} \bm{\Phi}\,,
\end{split}
\end{equation}
\begin{equation}
\label{eq:variance_laplace}
\begin{split}
	\sigma_a^2 
	&= \int p(a) [a^2 - \mathbb{E}[a]^2 ]\,\mathrm{d}a \\
	&= \int q(\mathbf{z}) \left( (m(\bm{\Phi}) + \mathbf{z}^\mathsf{T} \bm{\Phi})^2 - (m(\bm{\Phi}) + \mathbf{z}_{\text{MAP}}^\mathsf{T} \bm{\Phi})^2 \right) \,\mathrm{d}\mathbf{z} \\
	&= \bm{\Phi}^\mathsf{T} \mathbf{\Sigma}_N \bm{\Phi}\,.
\end{split}
\end{equation}
Thus our approximation to the predictive distribution in \cref{eq:predictive_q} becomes
\begin{equation}
\label{eq:predictive_a}
    C(\mathbf x) \simeq \int p(k=1 \mid \bm \omega, \mathbf z) q(\mathbf z)\,\mathrm{d} \mathbf z = \int \sigma(a) \mathcal{N}(a\mid\mu_a, \sigma_a^2)\,\mathrm{d}a\,.
\end{equation}
Since the integral in \cref{eq:predictive_a} cannot be evaluated analytically due to the sigmoid function, we need to approximate it to obtain the marginal class prediction. One can approximate the integral by exploiting the similarity between the logistic sigmoid function $\sigma(a)$ and the inverse probit function \citet{bishop2006pattern,murphy2012probabilistic}, which is given by the cumulative distribution of the standard Gaussian $\Phi(a)$. In order to obtain good approximation results, we need to rescale the horizontal axis so that $\sigma(a)$ has the same slope as $\Phi(\lambda a)$, where $\lambda^2 = \pi / 8$. By replacing $\sigma (a)$ with $\Phi ( \lambda a)$ in \cref{eq:predictive_a}, we obtain the approximated predictive distribution:
\begin{equation}
\label{eq:probit_approximation}
\begin{split}
	p(k=1 \mid \bm \omega, \mathcal{D}_N) 
    &\approx \int \Phi(\lambda a) \mathcal{N}(a \mid \mu_a, \sigma_a^2) \,\mathrm{d}a\\
    &= \Phi \left( \frac{\mu_a}{(\lambda^{-2}+\sigma_a^2)^{1 / 2}} \right)\\
    &= \sigma \left( (1 + \pi \sigma_a^2 / 8)^{-1/2} \mu_a \right)\,.
\end{split}
\end{equation}

\section{Regularizing the latent task space}
\label{sec:latent_reg}
To make sure our task distribution conforms to the prior distribution $p(\mathcal{Z})$, we followed the approach in \citet{saseendran2021shape}, where they regularize the learned latent representation towards a given prior distribution in a tractable way. Their approach builds on the non-parametric Kolmogorov-Smirnov (KS) test for one-dimension probability distributions and extend it to a multivariate setting, which allows for gradient-based optimization and can be easily applied to expressive multi-modal prior distributions. 

Directly extending the KS test to high-dimensional distributions is challenging, since it requires matching joint CDFs, which is especially infeasible in this case. Therefore, they propose to match the marginal CDFs of the prior, making the regularization tractable. Given $d$-dimensional task embedding $\mathbf{z}_1, \dots, \mathbf{z}_T$ for $T$ related tasks, the empirical CDF in dimension $j$ is defined as:
\begin{equation}
    F(z) = \frac{1}{T} \sum^T_{t=1} \mathbbm{1}([\mathbf{z}_t]_j \leq z)\,,
\end{equation}
where $\mathbbm{1}([\mathbf{z}_t]_j \leq z)$ is a indicator function if $j$-th component of $\mathbf{z}_t$ is smaller or equal than a certain value $z$. In addition to the marginals, they also regularize the empirical covariance matrix $\text{Cov}({\mathbf{z}_1, \dots, \mathbf{z}_T})$ to be close to the covariance of the prior. In our case, the prior task distribution is a isotropic Gaussian $p(\mathcal{Z}) = \mathcal{N}(\mathbf{0}, \mathbf{I})$, and the resulting regularizer can be written as:
\begin{equation}
    \mathcal{R}(\{ \mathbf{z}_t \}_{t=1}^T ; p(\mathcal{Z}))
    = \lambda_{\text{KS}}\underbrace{\sum^d_{j=1} (F([\mathbf{z}_t]_j) - \Phi([\mathbf{z}_t]_j))^2}_{\text{match marginal CDF of } p(\mathcal{Z)}}\\
    + \underbrace{
         \vphantom{ \sum^d_{j=1} }
         \lambda_{\text{Cov}} \| \mathbf{I} - \text{Cov}(\{ \mathbf{z}_t\}_{t=1}^T ) \|^2_\mathrm{F}}
         _{\text{match covariance of } p(\mathcal{Z})}\,,    
\end{equation}
where the marginal CDFs and correlations are compared through squared errors, the $\lambda_{\text{KS}}$ and $\lambda_{\text{CV}}$ are weighting factors controls the trade-off between matching the empirical marginal CDFs and the covariance.

\paragraph{Regularization coefficients estimation}
The two regularization coefficients $\lambda_{\text{KS}}$ and $\lambda_{\text{Cov}}$ are important hyperparameters and needed to be carefully treated. Notice that, the more tasks we have, the closer the match between the empirical CDFs and the prior marginal CDF based on the assumption that the related tasks are i.i.d.\ samples from the task distribution. To avoid the regularization being dominated by one term, we aim to scale them in a way that all terms converge to similar magnitudes with more tasks. Therefore, we choose the two factors such that
\begin{equation}
\begin{split}
    \lambda_{\text{KS}}^{-1} &= 2 \sum^d_{j=1} (F([\mathbf{z}_t]_j) - \Phi([\mathbf{z}_t]_j))^2\,,\quad \text{with}\ \mathbf{z}_t \sim p(\mathcal{Z})\,, \\
    \lambda_{\text{Cov}}^{-1} &= 2 \| \mathbf{I} - \text{Cov}(\{ \mathbf{z}\}_{t=1}^T) \|^2_\mathrm{F}\,,\quad \text{with}\  \mathbf{z}_t \sim p(\mathcal{Z})\,,
\end{split}
\end{equation}
where $\mathbf z_t$ is i.i.d.\ sample from the prior distribution for the coefficient estimation. This normalizes the regularizer to be approximately of order $1$ for samples follows the prior distribution.

\section{Ablation studies}
\label{appendix:ablation}
In this section, we conduct ablation studies to illustrate the impact of different components of \ourmethod on its performance. We introduce the following variants of \ourmethod as in \cref{sec:experiments}:
\begin{itemize}
    \item \ourmethod (Probit): Utilizes the marginalized form of the acquisition function (see \cref{sec:probit_approximation}) without gradient boosting.
    \item \ourmethod (TS): Employs only Thompson sampling without gradient boosting.
    \item \ourmethod (RES): Removes the mean prediction layer $m(\cdot)$ while keeping other components unchanged. 
    \item \ourmethod (MEAN): Excludes the task prediction layer $h_{\mathbf z_t}(\cdot)$ and uses only the task-agnostic meta-learning component $g_{\bm\omega}$ with gradient boosting. This variant focuses on meta-learning the initial design for optimization. Note that, due to the absence of task embedding, the model is trained without the task space regularization in \cref{eq:regularization}, making the Thompson sampling strategy inapplicable.
    \item \ourmethod (RF): Replaces gradient boosting with a random forest (RF) classifier, using the implementation from scikit-learn \citep{pedregosa2011scikitlearn}. Following \citet{song2022general}, the hyperparameters are set as: $\text{n\_estimator} = 1000$, $\text{min\_samples\_split} = 2$, $\text{max\_depth}=\text{None}$, $\text{min\_samples\_leaf} = 1$. Unlike in gradient boosting, this variant requires explicit balancing of the results from the meta-learning and residual models, which we achieve by averaging their predictions without applying a sophisticated weighting scheme.
    \item \ourmethod (MLP): Substitutes gradient boosting with a two-layer multi-layer perceptron (MLP) classifier, with 32 hidden units per layer and ReLU activation. The MLP is optimized using ADAM \citep{kingma2015adam} with learning rate $\text{lr}=10^{-3}$ and batch size $B=64$. Predictions from the meta-learning and residual models are averaged similarly to the RF variant.
\end{itemize}

\subsection{Latent feature analysis}
\label{ssec:features_analysis}
\begin{figure*}[tp]
    \begin{subfigure}{.5\textwidth}
        \centering
        \includegraphics[width=\linewidth]{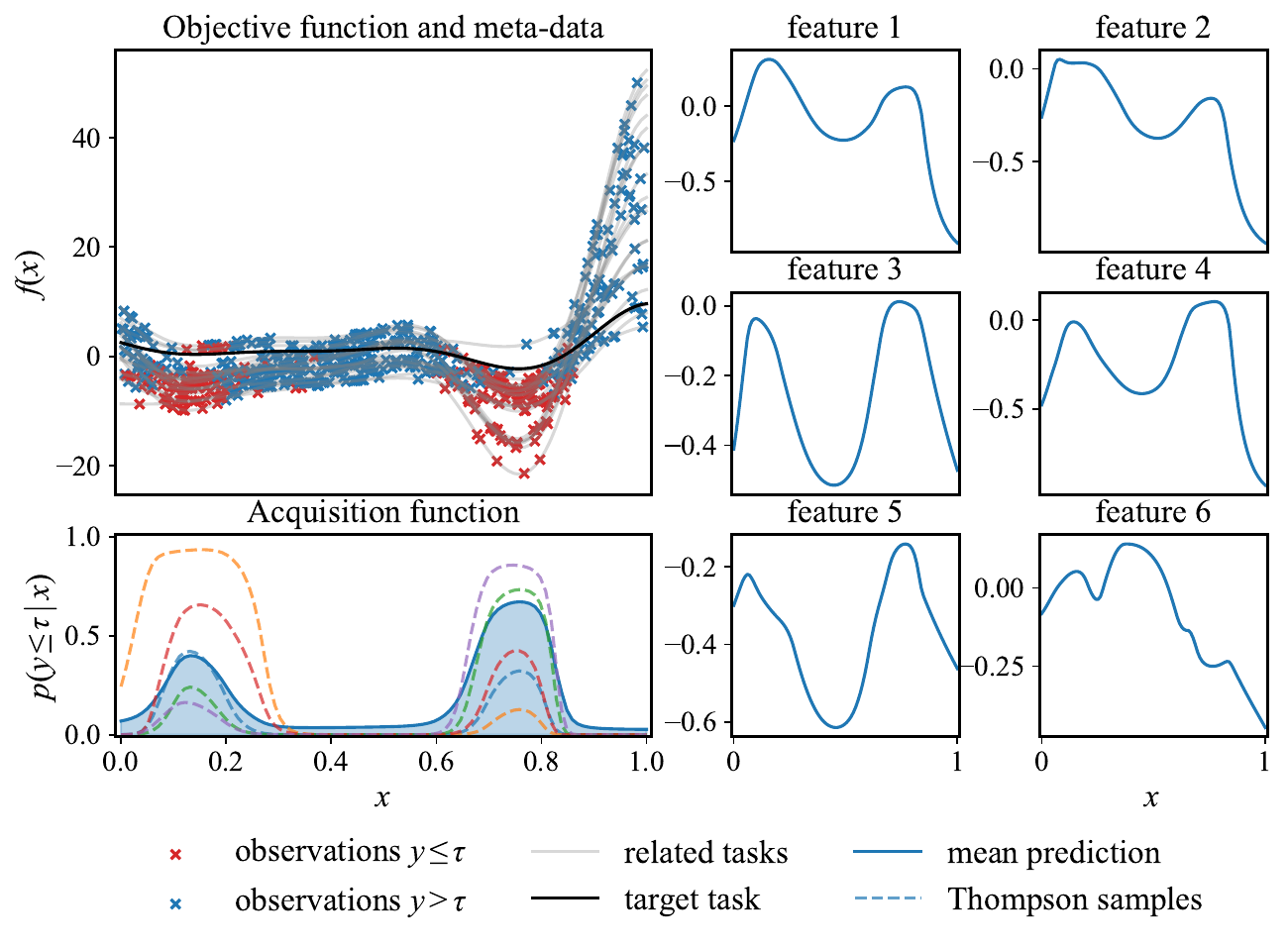}
        \caption{Forrester function features}
        \label{fig:forrester_features}
    \end{subfigure}
    \hfill
    \begin{subfigure}{.5\textwidth}
        \centering
        \includegraphics[width=\linewidth]{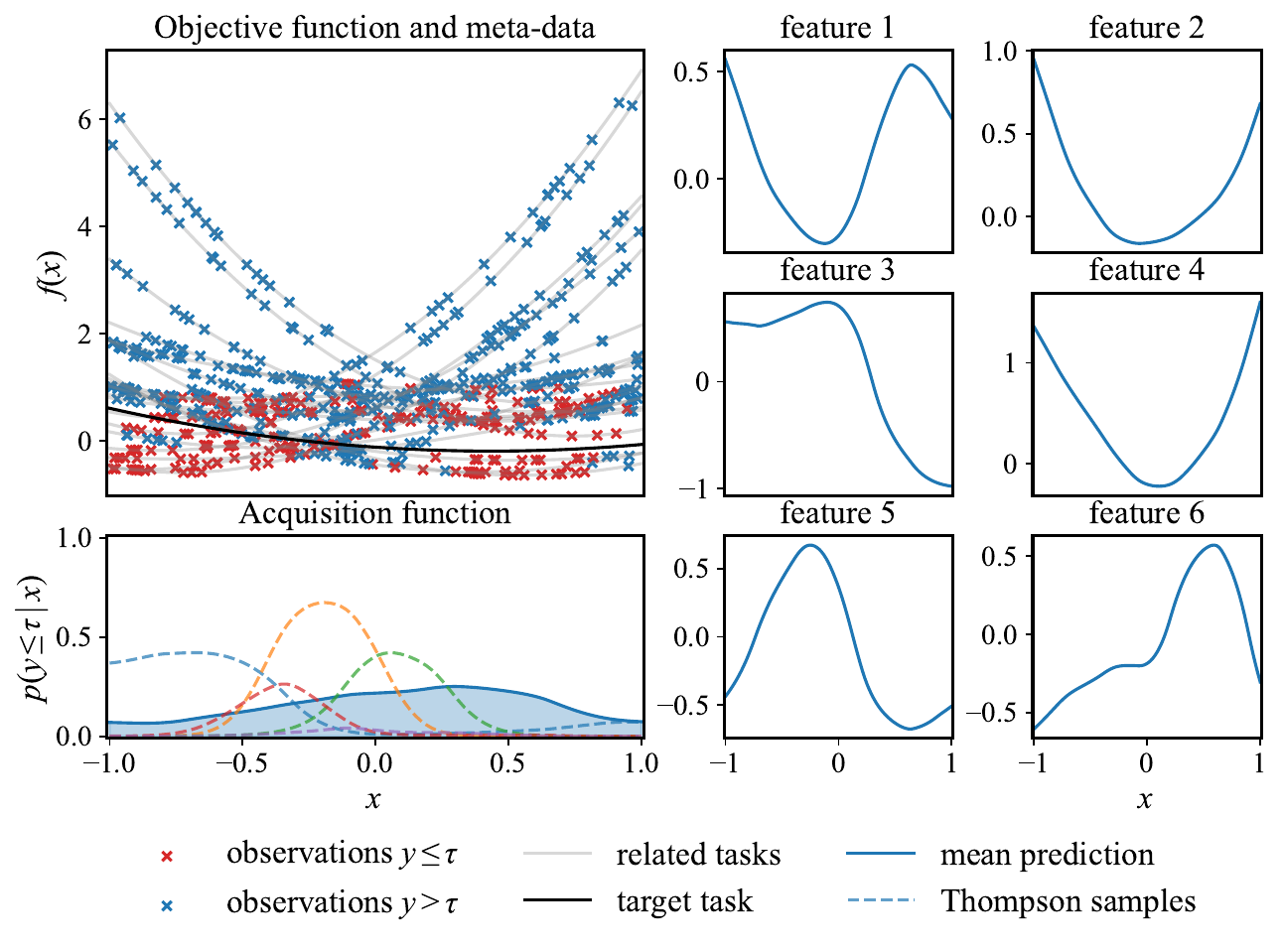}
        \caption{Quadratic function features}
        \label{fig:quadratic_features}
    \end{subfigure}
    \caption{Left: Forrester functions with two likely optima as target function and related tasks. The learned acquisition function is shown below. The meta-learned latent features show that the model successfully infers the location of two optima, resulting in a acquisition with two modes around the optima. Right: Quadratic functions with varying optima as target function and related tasks. The meta-learned latent features show that the model is able to capture the global function shape shared across all tasks, even though there is no clear location for optima.}
\end{figure*}
To provide an intuition of the meta-learning in our method, we visualize the feature representation extracted from the meta-data by our meta-learning model. The latent features $\mathbf \Phi$ represent basis functions for the Bayesian logistic regression and should represent the structure of the meta-data distribution. With successfully learned features $\mathbf \Phi$ and the mean layer, our model performs the task adaptation by reasoning about the latent task embedding vector $\mathbf{z}$. It produces predictions with similar structure to the meta-data that match the class labels on the target function. In order to learn a effective feature representation, one should capture both the local and global structure of the function. Therefore, we select two types of function to study the effectiveness of feature learning for \ourmethod:
\begin{enumerate*}[label=\roman*)]
  \item Forrester functions \citep{sobester2008engineering} with two very likely positions for the global optimum, which allows for effective warm-starting and requires local adaptation.
  \item quadratic functions, where the functions share a certain global shape, but the optima could be located anywhere in the search space.
\end{enumerate*}
For more details on the synthetic functions and the generation of meta-data, we refer to \cref{appendix:benchmark_details}.

The results for these two synthetic functions are shown in \cref{fig:forrester_features} and \cref{fig:quadratic_features} respectively. In \cref{fig:forrester_features}, we observe that the features learned by \ourmethod exhibit either a maximum or a minimum around the two likely optima, indicating that the model successfully infers the location of the most promising values from the meta-data. In \cref{fig:quadratic_features}, even without a clear location of optima, the features still follow the shape of quadratic functions with different minima.

\subsection{Effects of Thompson Sampling}
\label{ssec:ablation_thompson_sample}
\begin{figure*}[tp]
    \begin{subfigure}{.5\textwidth}
        \centering
        \includegraphics[width=\linewidth]{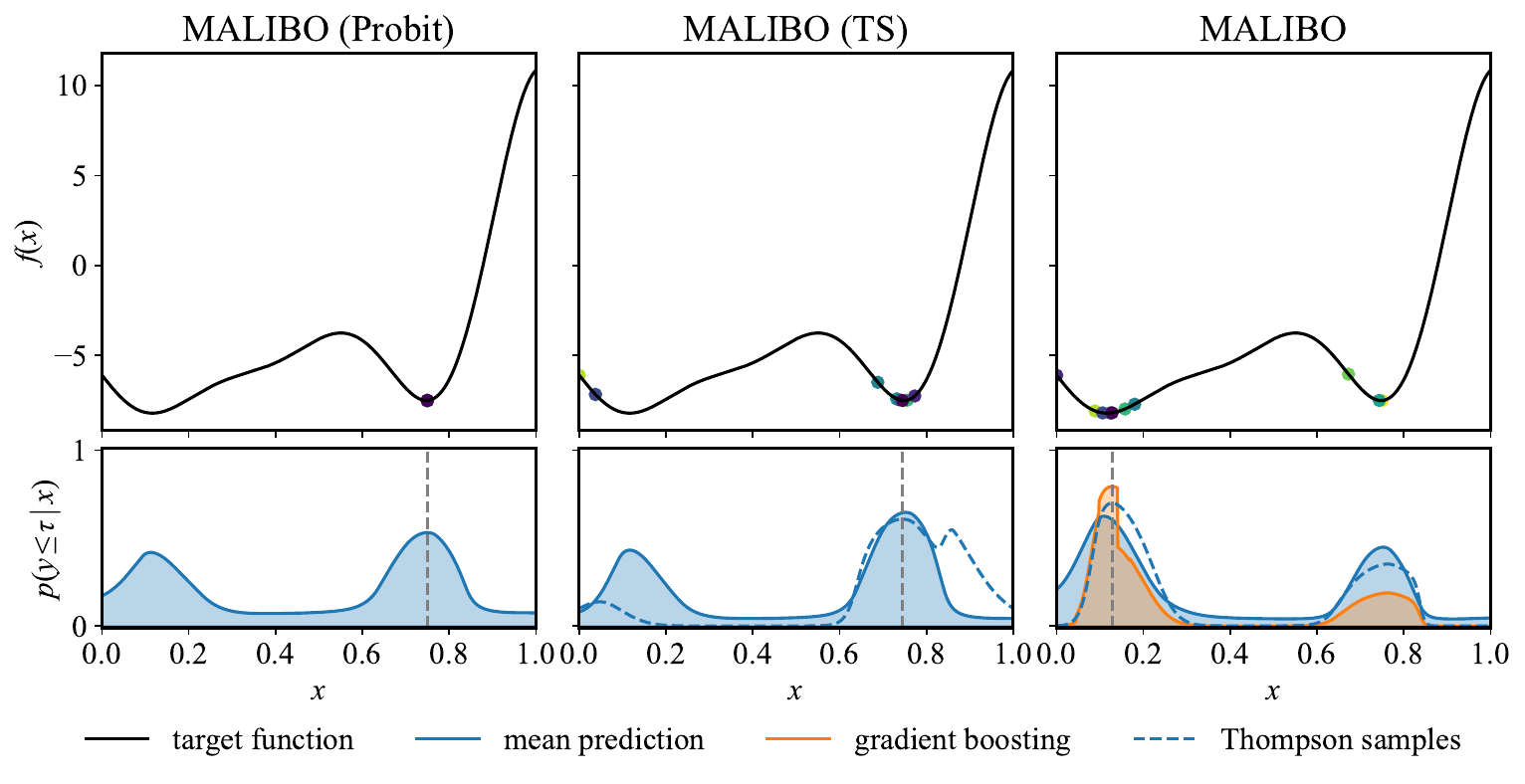}
        \caption{Forrester function features}
        \label{fig:comparison_forrester}
    \end{subfigure}
    \hfill
    \begin{subfigure}{.5\textwidth}
        \centering
        \includegraphics[width=\linewidth]{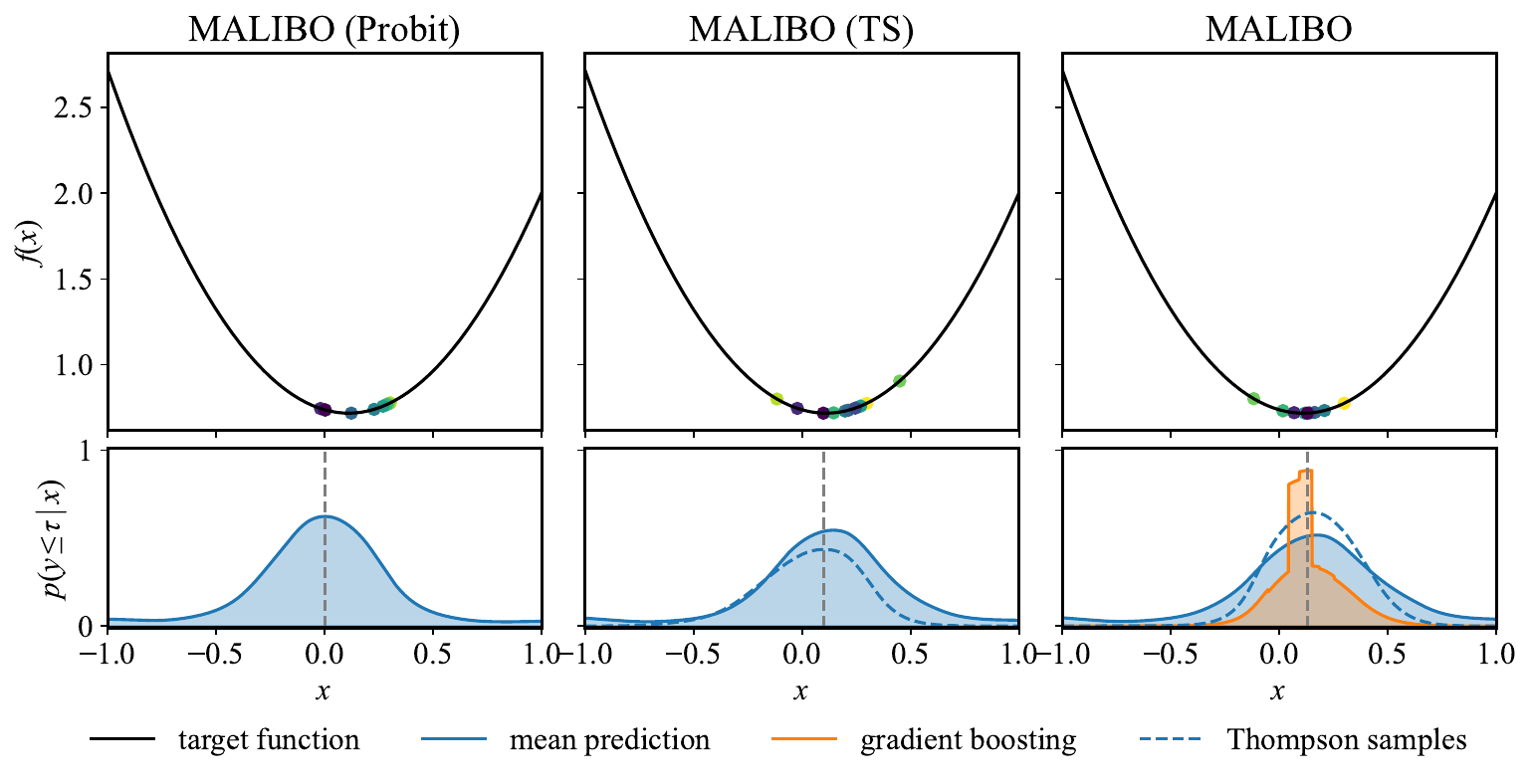}
        \caption{Quadratic function features}
        \label{fig:comparison_quadratic}
    \end{subfigure}
    \caption{Task adaptation of different \ourmethod variants on Forrester and quadratic functions after meta-learning. Each method optimizes for \num{10} iterations.}
    \label{fig:thompson_sampling_comparison}
\end{figure*}

To understand how the exploration help with the optimization, we compare the task adaptation performance among \ourmethod (Probit), \ourmethod (TS) and \ourmethod on synthetic benchmarks. As illustrated in \cref{fig:thompson_sampling_comparison}, \ourmethod (Probit) tends to exhibit conservative behavior in both the Forrester and quadratic function scenarios, leading to sub-optimal performance. This is primarily due to its limited exploration capabilities. In contrast, \ourmethod (TS), which incorporates Thompson sampling, demonstrates more robust exploration in both cases. Interestingly, even though \ourmethod is enhanced with gradient boosting, its exploration performance appears similar to the Thompson sampling-only variant. The influence of gradient boosting is evident in the change of acquisition function values. It suppresses the values in regions where the sampled function might predict high values, but existing observations suggest otherwise. Conversely, it amplifies the acquisition function values near the current best observations, thereby fostering stronger convergence. In the initial stages of optimization, both task adaptation and gradient boosting face a challenge due to the scarcity of observations, which limits confident prediction. Unlike other methods that rely on random search as an exploration strategy, Thompson sampling incorporates task uncertainty. It efficiently explores potential optima by leveraging meta-learned information from related tasks, making it a more effective approach in the context of exploration and optimization.

\subsection{Effects of gradient boosting}
\label{ssec:ablation_gradient_boosting}
\begin{figure*}[tp]
    \centering
    \includegraphics[width=.8\linewidth]{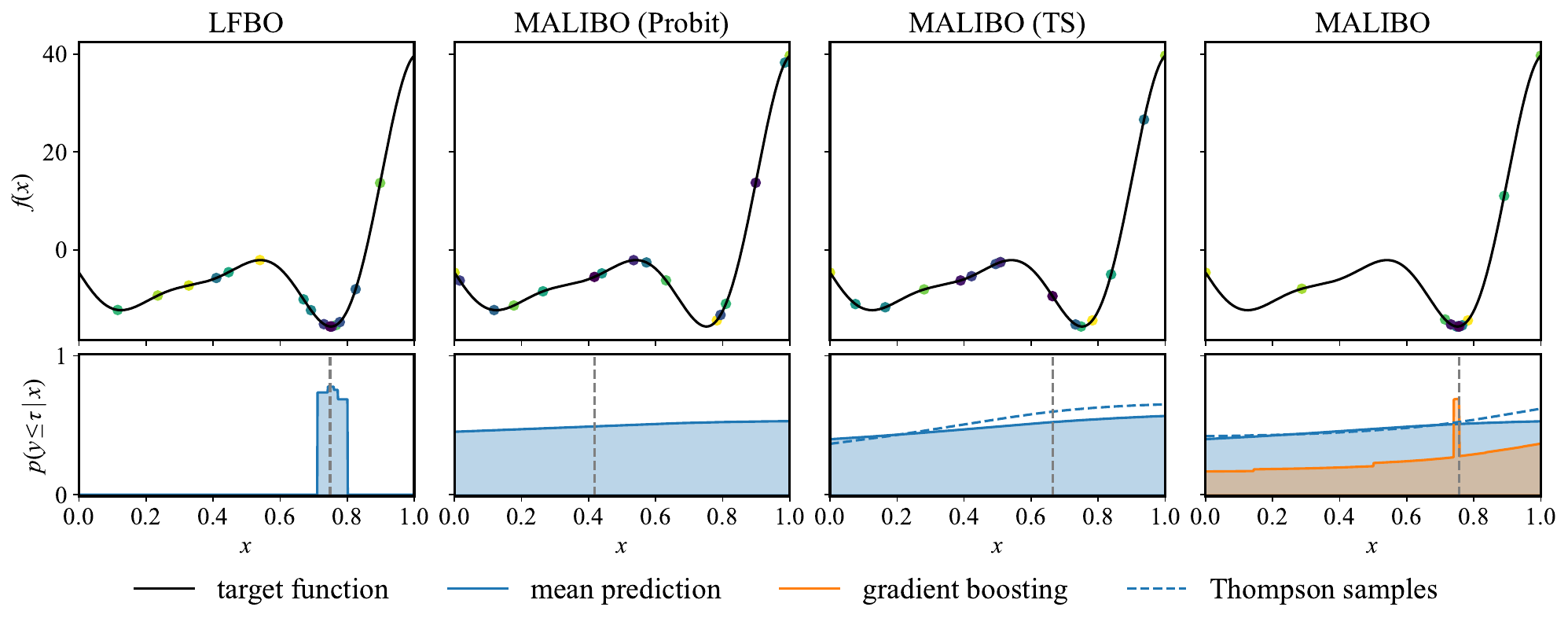}
    \caption{Task adaptation of \lfbo and different MALIBO variants on a Forrester function without meta-learning. Each method optimize for \num{16} iterations}
    \label{fig:ablation_forrester}
\end{figure*}
To illustrate the effectiveness of gradient boosting in \ourmethod, we conduct an experiment that focus on this aspect by excluding meta-learning. This approach mimics scenarios where meta-learning fails to aid task adaptation. As shown in \cref{fig:ablation_forrester}, our experiments compare various \ourmethod variants on a Forrester function, and contrast these results with \lfbo to assess their performances without meta-learning. The findings, depicted in Figure \cref{fig:ablation_forrester}, reveal that \ourmethod (Probit) and \ourmethod (TS) are inefficient in optimizing the function due to their exclusive reliance on meta-learned features for task adaptation. This reliance results in poor performance when the meta-learned priors are uninformative. In contrast, the proposed \ourmethod efficiently locates the optima and performs similarly to \lfbo. This efficiency is attributed to the ability of gradient boosting to counteract ineffective predictions from the meta-learning process. Specifically, gradient boosting enables subsequent learners to correct initial errors from the meta-learned model, thereby aligning the performance of \ourmethod with that of \lfbo.

\subsection{Effects of different inference methods}
\label{ssec:inference_method}
\begin{figure*}[tp]
    \centering
    \includegraphics[width=\linewidth]{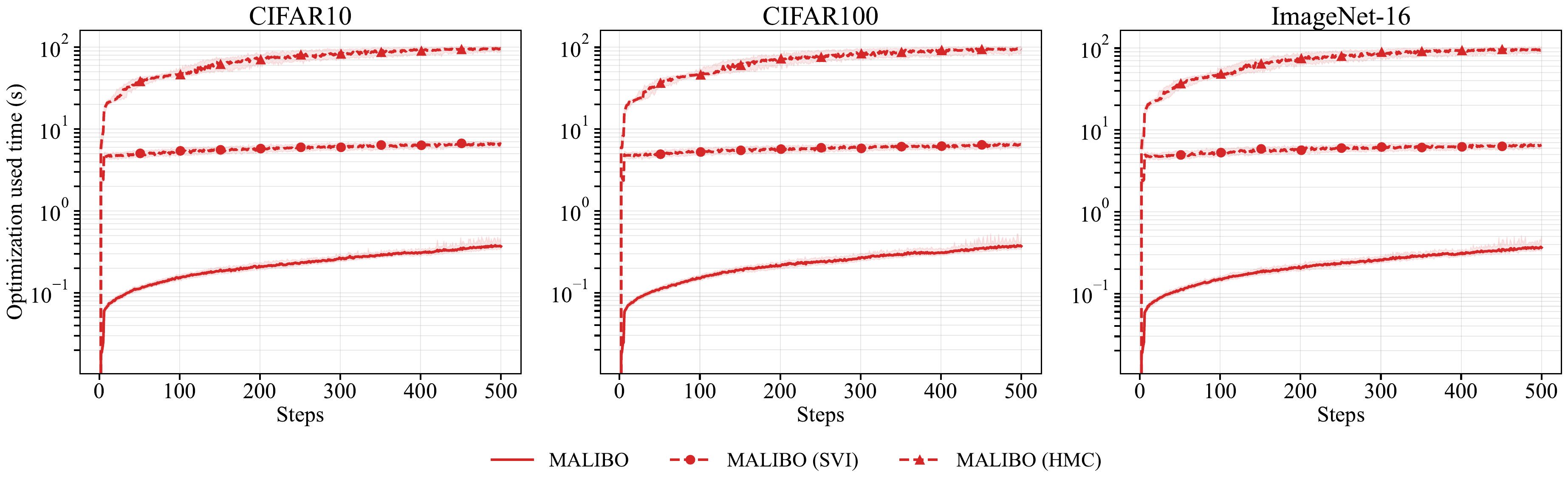}
	\caption{Runtime of \ourmethod using different inference methods over optimization steps on NASBench201. We plot the medial inter-quantiles to remove outliers.}
	\label{fig:nas_inference_opt_time}
\end{figure*}
\begin{figure*}[tp]
    \centering
    \includegraphics[width=\linewidth]{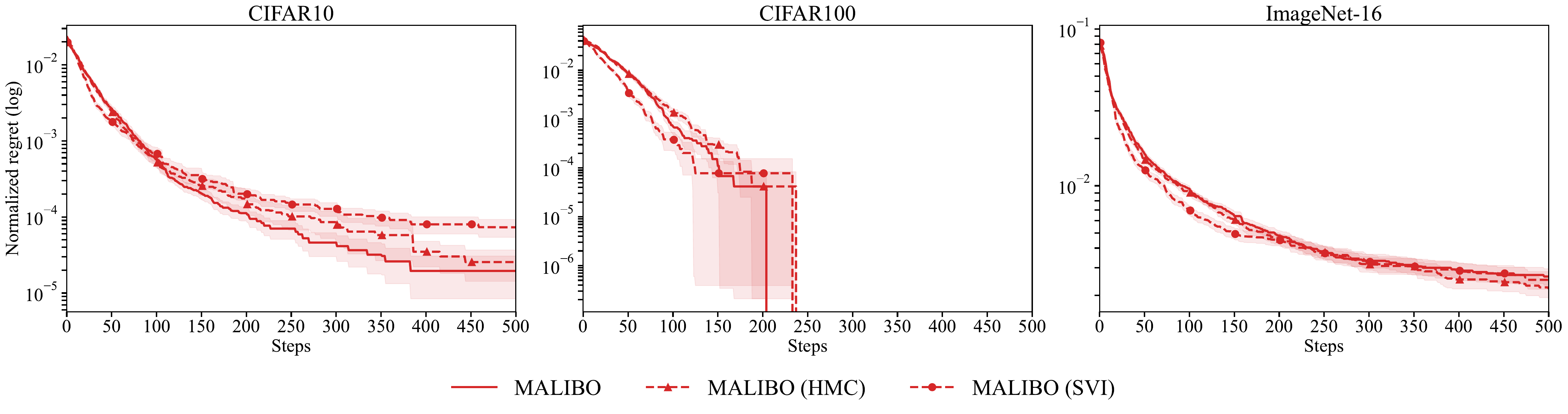}
	\caption{Normalized regrets of \ourmethod using different inference methods on NASBench201.}
	\label{fig:nas_inference}
\end{figure*}
In this section, we investigate the performance of different approximation methods for the posterior task embedding $p(\mathbf{z} \mid \mathcal{D}_N)$. Specifically, we consider three different inference methods, namely Hamiltonian Monte Carlo (HMC), stochastic variational inference (SVI) and the Laplace approximation. Compared to the Laplace approximation, SVI and HMC normally take longer time for the approximation, especially for HMC, as it needs multiple samples to estimate the expectation. To show their speed and scalability, we compare their runtime for optimization on NASBench201 in \cref{fig:nas_inference_opt_time} and show that the \ourmethod with Laplace approximation takes around \num{0.1} second for every iteration, while \ourmethod (SVI) takes around \num{10} seconds and \ourmethod (HMC) \num{100} seconds. Although the SVI and HMC variants take longer time for inference, we show that the performance among these methods are close in \cref{fig:nas_inference}. Similar behaviors are also observed in other benchmarks, including the HPOBench and HPO-B. Due to the fast inference time and competitive performance, we use Laplace approximation as our proposed inference method for \ourmethod.

\subsection{Effects of task embedding dimension}
\label{ssec:embedding_dimension}
\begin{figure*}[ht]
    \centering
    \includegraphics[width=.7\linewidth]{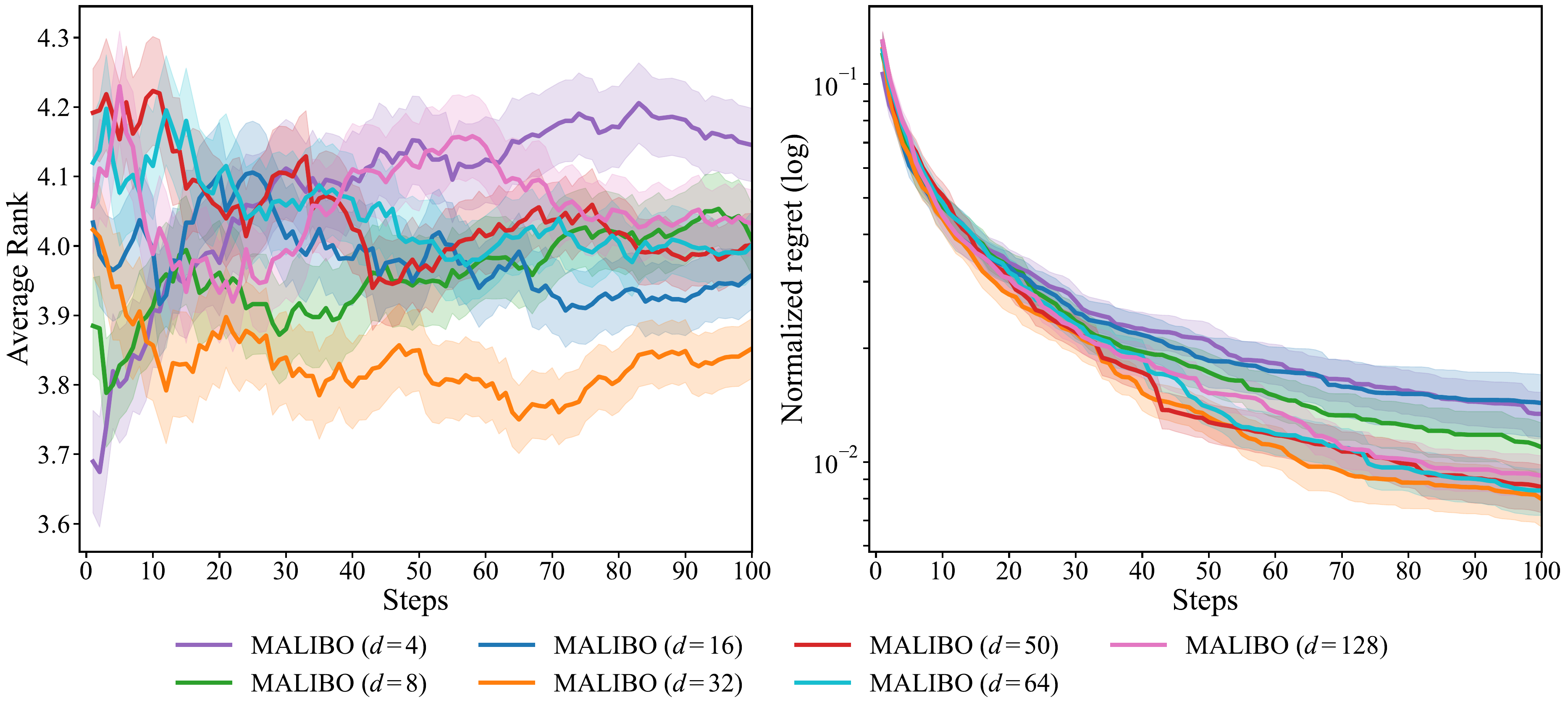}
	\caption{Aggregated comparisons across all search spaces for MALIBO with different task embedding dimensions on HPO-B.}
	\label{fig:latent_results}
\end{figure*}
Given that the task embedding dimension $d=50$ is fixed across all experiments, its impact on the performance of \ourmethod remains uncertain. To address this, we conduct an ablation study using the HPO-B benchmark, which encompasses tasks with input dimensions ranging from 2 to 18. We test various task embedding dimensions $d={4,8,16,32,50,64,128}$. As illustrated in \cref{fig:latent_results}, lower dimensional embeddings tend to yield better initial performance. However, this performance often plateaus, possibly due to the embeddings' limited expressiveness. Conversely, while increasing the dimensionality generally enhances performance, this improvement persists only up to a certain point, specifically $d=32$ in our study, beyond which the influence of task embedding dimensionality on performance diminishes.

\subsection{Quantitative comparison}
\label{ssec:quantitative_comparison}
\begin{figure*}[tp]
    \centering
    \includegraphics[width=.7\linewidth]{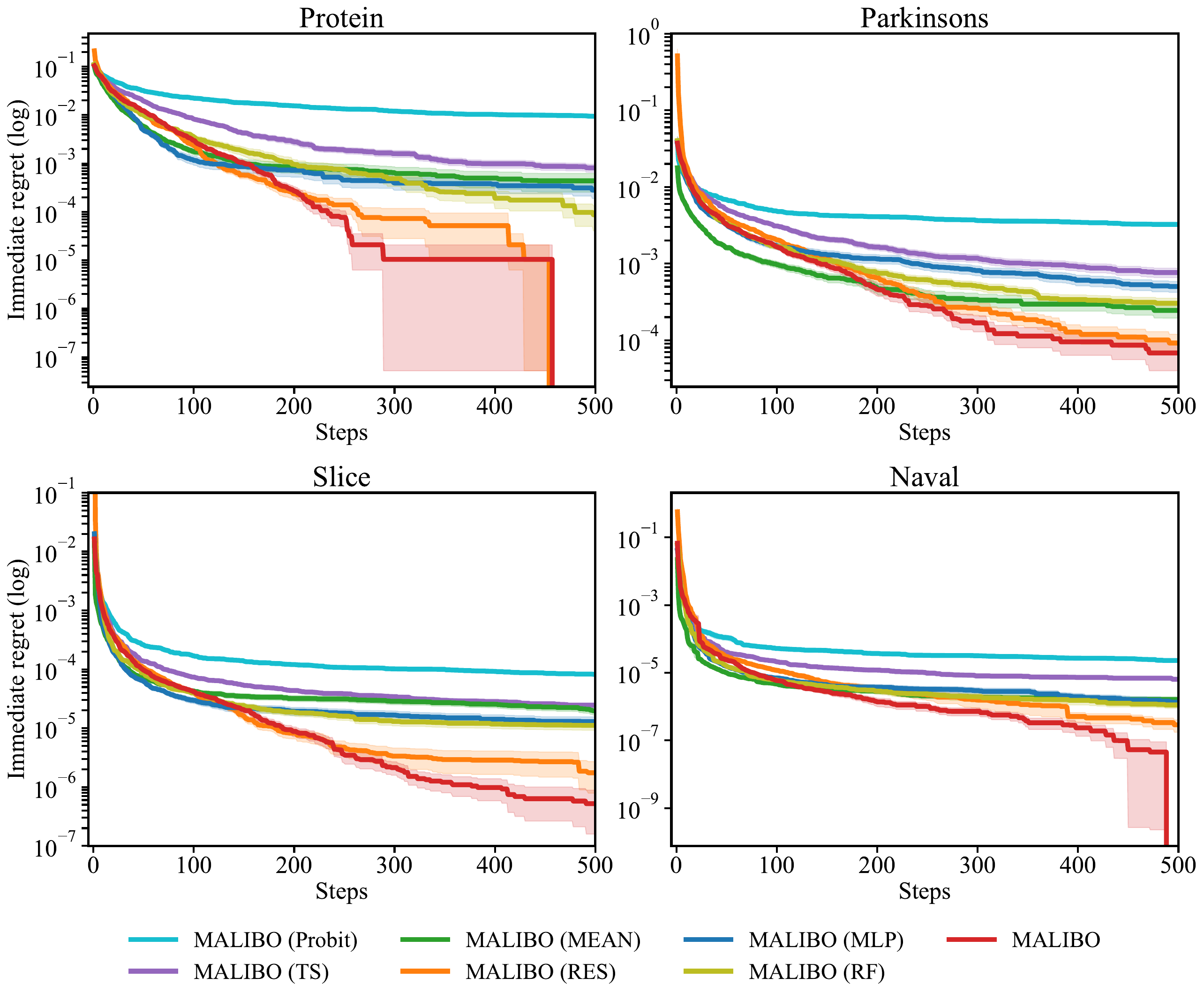}
    \caption{Immediate regrets of \ourmethod variants on all tasks in HPOBench.}
    \label{fig:comparison_hpo}
\end{figure*}

\begin{figure*}[tp]
    \centering
    \includegraphics[width=\linewidth]{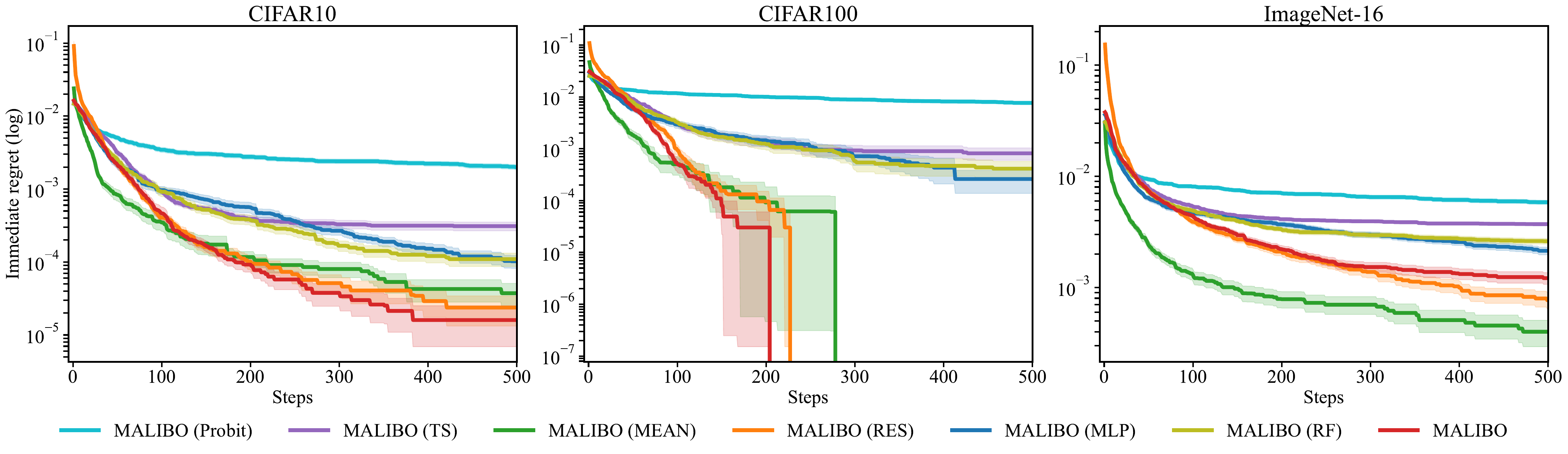}
    \caption{Immediate regrets of \ourmethod variants on all tasks in NASBench201.}
    \label{fig:comparison_nas}
\end{figure*}

\begin{figure*}[tp]
    \begin{subfigure}{.54\textwidth}
        \centering
        \includegraphics[width=.9\linewidth]{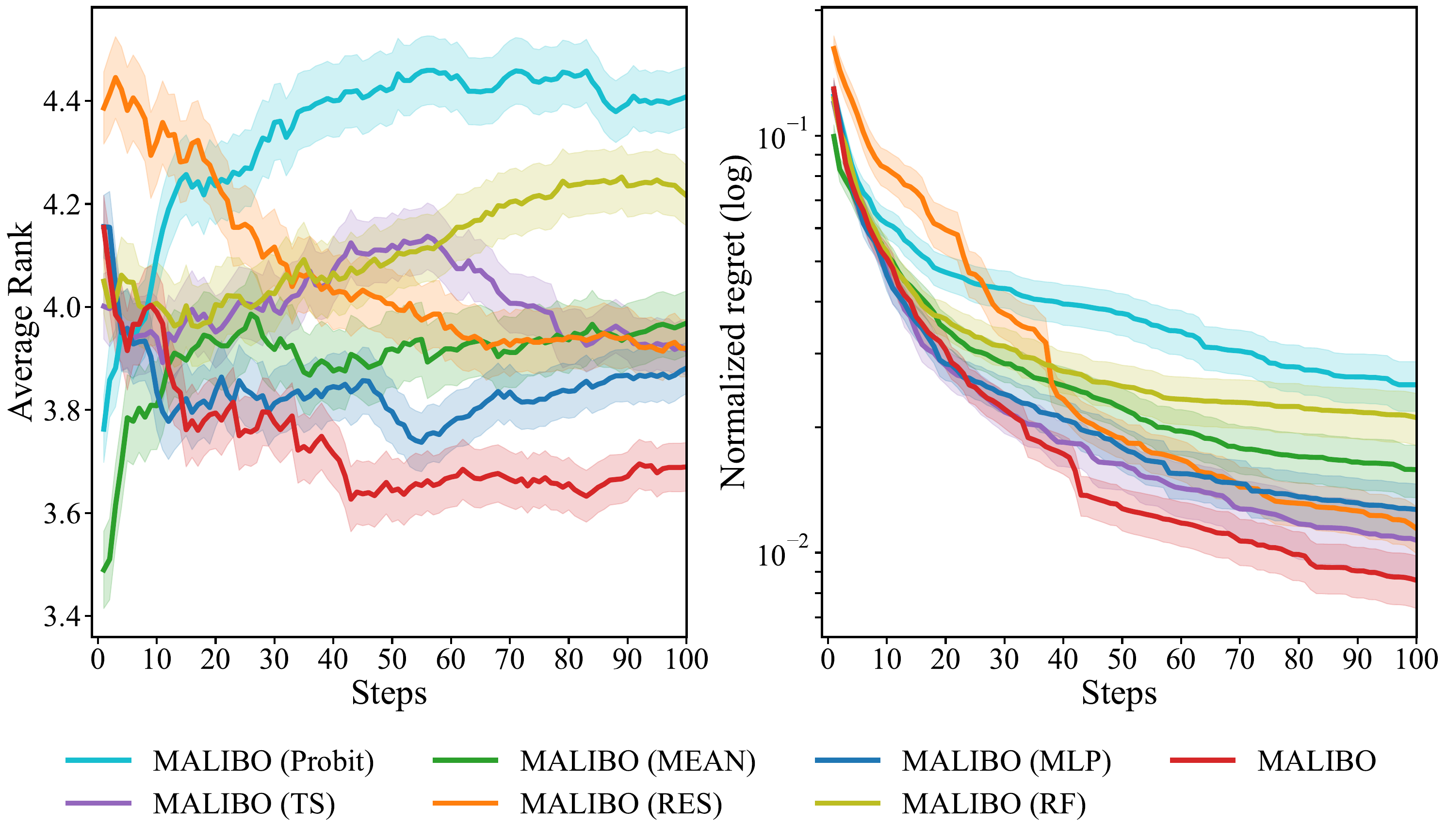}
    \end{subfigure}
    \hfill
    \begin{subfigure}{.45\textwidth}
        \includegraphics[width=.9\linewidth]{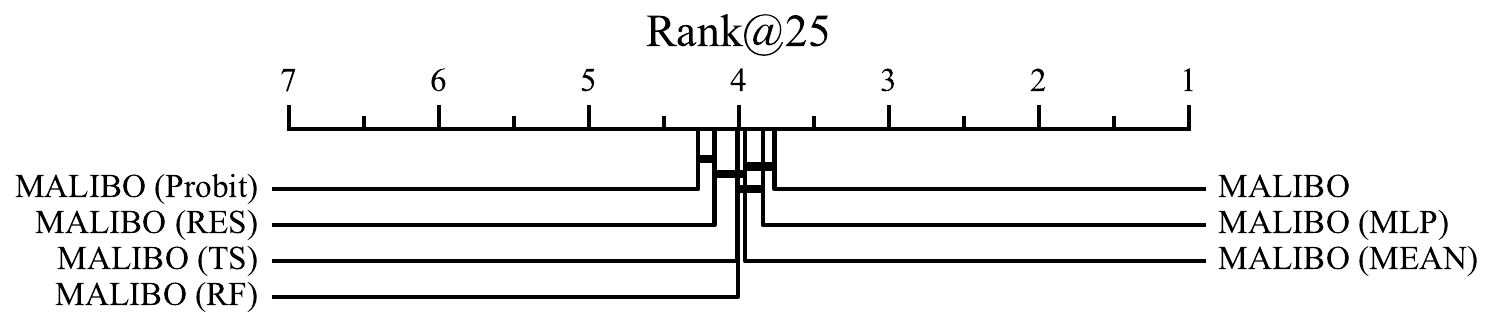}
        \includegraphics[width=.9\linewidth]{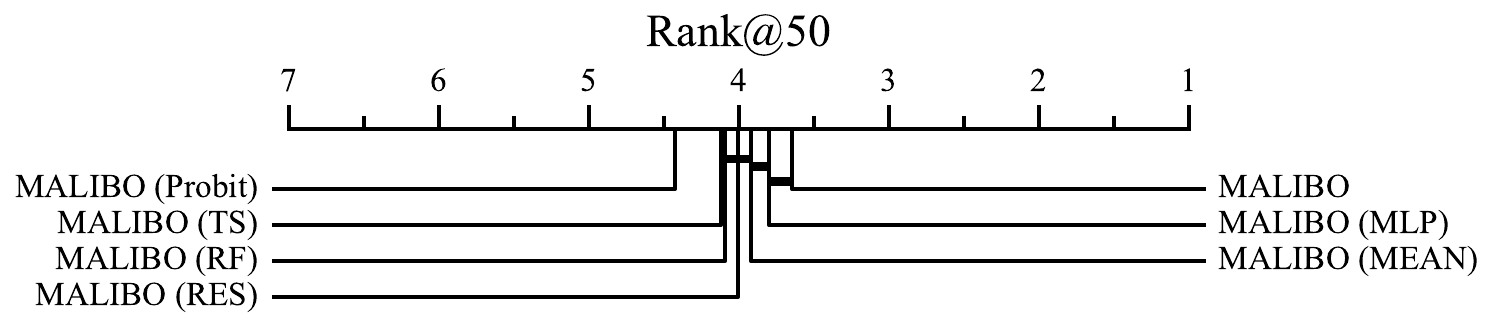}
        \includegraphics[width=.9\linewidth]{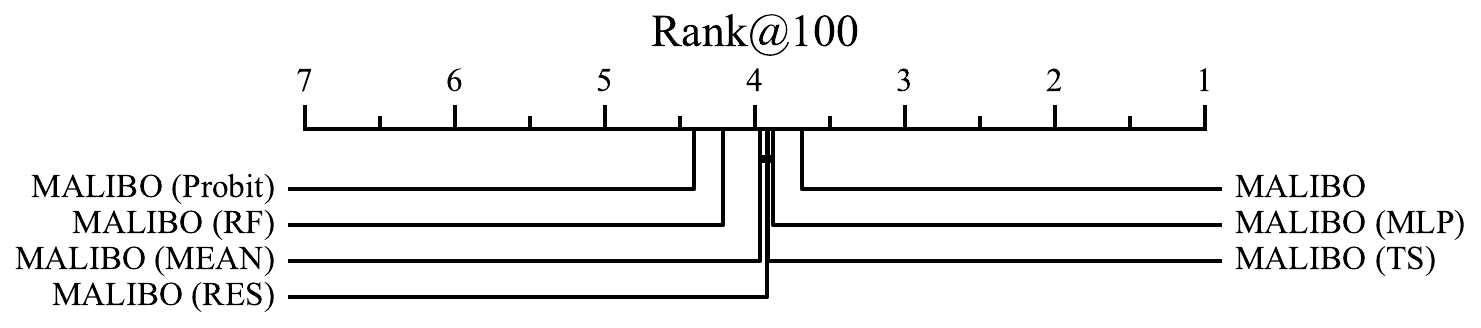}
    \end{subfigure}
    \caption{Aggregated comparisons of normalized regret and average ranks across all search spaces for \ourmethod variants on HPO-B.}
    \label{fig:comparison_hpob}
\end{figure*}

In this section, we demonstrate the detailed experimental results for the quantitative ablation study that is introduced in \cref{sec:experiments}. This study, illustrated in \cref{fig:comparison_hpo,fig:comparison_nas,fig:comparison_hpob}, compares seven variants of \ourmethod across all real-world benchmarks. These variants include \ourmethod (Probit), \ourmethod (TS), \ourmethod (MEAN), \ourmethod (RES), \ourmethod (RF), \ourmethod (MLP) and the proposed \ourmethod. We evaluated the performance using immediate regrets, which is the absolute error between the global minimum and the best evaluated results so far. This metric was applied to both HPOBench and NASBench201. For HPO-B, we followed the plotting protocol established by \citet{pineda2021hpob}., where the plots include the normalized regrets and average rank across benchmarks, alongside with the critical difference diagram \citep{demsar2006critical} for the ranks of all runs @25, @50, and @100 steps to assess the statistical significance of methods. 

\section{Additional results}
\label{appendix:additional_benchmarks}
In this section, we present results related to the benchmarks discussed in \cref{sec:experiments}. To further examine the robustness against heteroscedastic noise, we demonstrate additional experiments on two more synthetic functions. For the real-world benchmark, 
we include comprehensive results for all target tasks within the benchmarks, offering a more detailed analysis. Additionally, we provide a runtime analysis to demonstrate the the scalability of our proposed method.

\subsection{Noise experiment}
\begin{figure*}[tp]
    \centering
    \includegraphics[width=\linewidth]{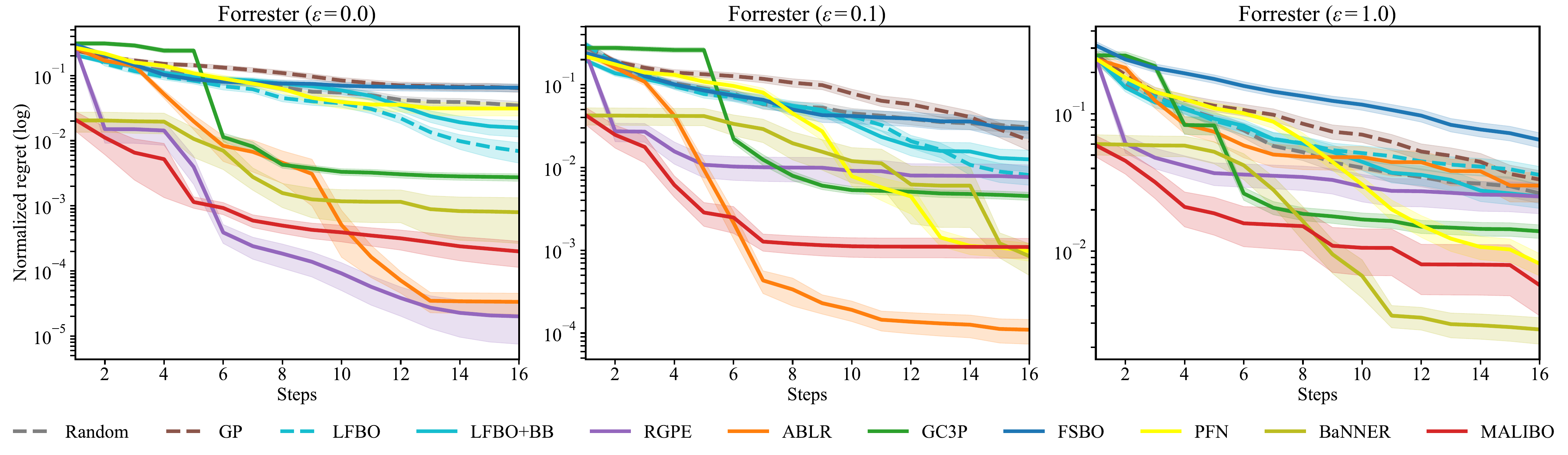}
	\caption{Normalized regret for BO algorithms on Forrester function ensembles ($D=1$) with different levels of multiplicative noise.}
	\label{fig:forrester}
\end{figure*}

\begin{figure*}[tp]
    \centering
    \includegraphics[width=\linewidth]{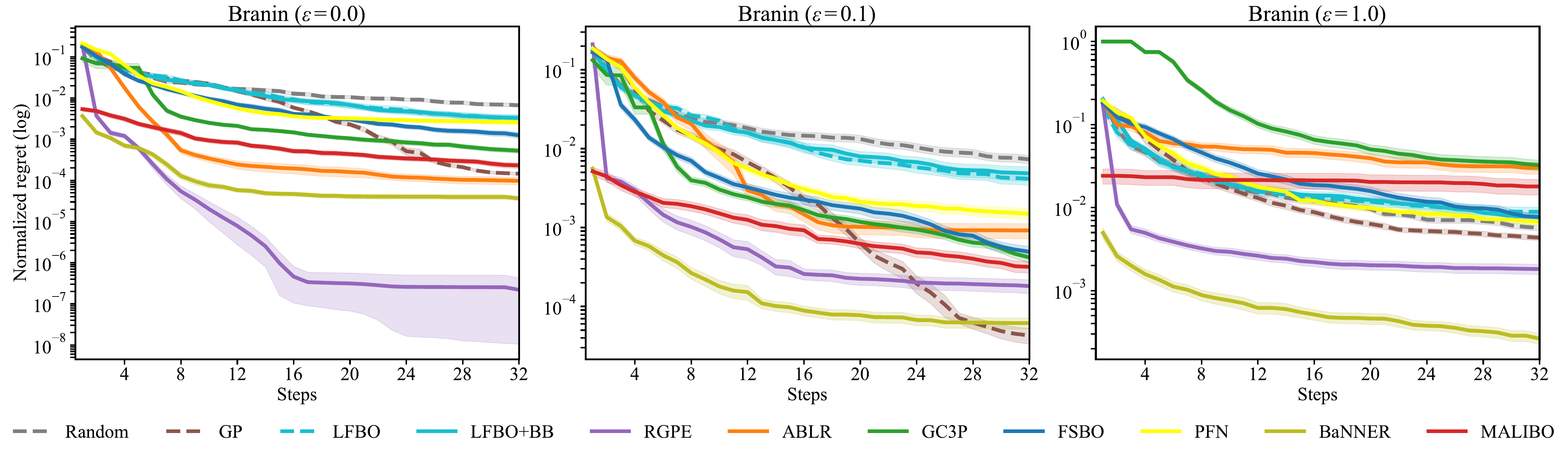}
	\caption{Normalized regret for BO algorithms on Branin function ensembles ($D=2$) with different levels of multiplicative noise.}
    \label{fig:branin}
\end{figure*}

In this experiments, we use the Forrester\citep{sobester2008engineering} and Branin \citep{dixon1978global} function ensembles as additional benchmarks, with detailed descriptions available in \cref{appendix:step-through-vis}. For meta-learning, we randomly sampled $N$ noisy observations in $T$ related tasks, setting $N=128, T=128$ for Forrester and $N=128, T=256$ for Branin. As illustrated in \cref{fig:forrester,fig:branin}, \ourmethod consistently demonstrate strong warm-starting performance and stay robust to noise compared to most of the baselines. Similarly, the performances of \lfbo and \lfbobb are relatively stable across different noise levels but are only comparable to random search. Notably, while RGPE and ABLR both outperform the other likelihood-free based methods in the noise-free setting, their performances degrade significantly with increased noise levels except for RGPE in Branin ($\epsilon=1.0$).

\subsection{Runtime analysis}
\label{ssec:runtime_analysis}
The runtime efficiency of \ourmethod is investigated in \cref{sec:experiments}, where we illustrate the runtime of the optimization algorithms for each step in \cref{fig:opt_time}. The runtime for \ourmethod is the second fastest among all the meta-learning methods, while only slightly slower than \lfbo and \lfbobb. Due to the increasing amount of observations, the runtime of almost all the methods grows over with the number of iterations, especially for RGPE and PFN. The most time-consuming methods are FSBO, BaNNER and PFN, which take almost 100 seconds for each steps. For PFN and BaNNER, this is mostly from optimizing the acquisition function, which requires multiple initializations to guarantee better convergence to the global optimum. In terms of FSBO, the time overhead also comes from the additional training in task-adaptation phase besides the acquisition function optimization. Although ABLR and GC3P are around one order of magnitude slower than \ourmethod at the beginning, but their runtime remain stable throughout the optimization. ABLR always retrain on all data, which constitutes the largest computational burden at each step, and the complexity of the Bayesian linear regression is more scalable than GPs. For GC3P, we attribute the almost constant runtime to aggressive settings for the GPs hyperparameter optimization, which is usually the most expensive step. The growth in runtime for \lfbo and \lfbobb can be attributed exclusively to the fitting of the gradient boosted trees. Similarly, \ourmethod uses gradient boosting as residual prediction model, which retrains on the dataset for every iteration, therefore the runtime grows with the number of iterations as well.

\begin{figure*}[tp]
    \centering
    \includegraphics[width=.9\linewidth]{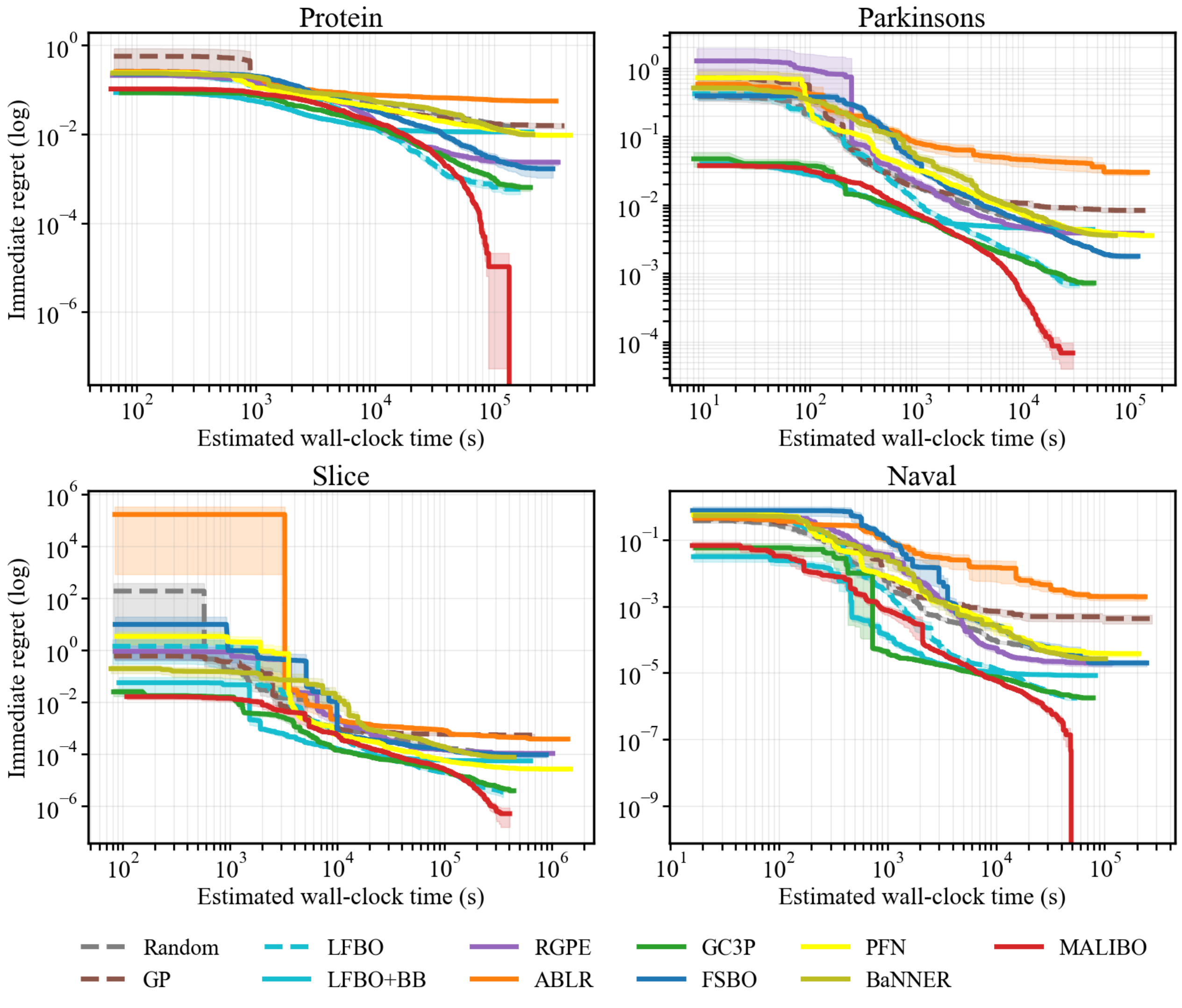}
	\caption{Immediate regrets of different BO algorithms on the HPOBench neural network tuning problem. Each algorithm runs for 500 iterations and we show the corresponding estimated wall-clock time on the $x$ axis in log scale.}
	\label{fig:hpo_time}
\end{figure*}

\begin{figure*}[tp]
    \centering
    \includegraphics[width=\linewidth]{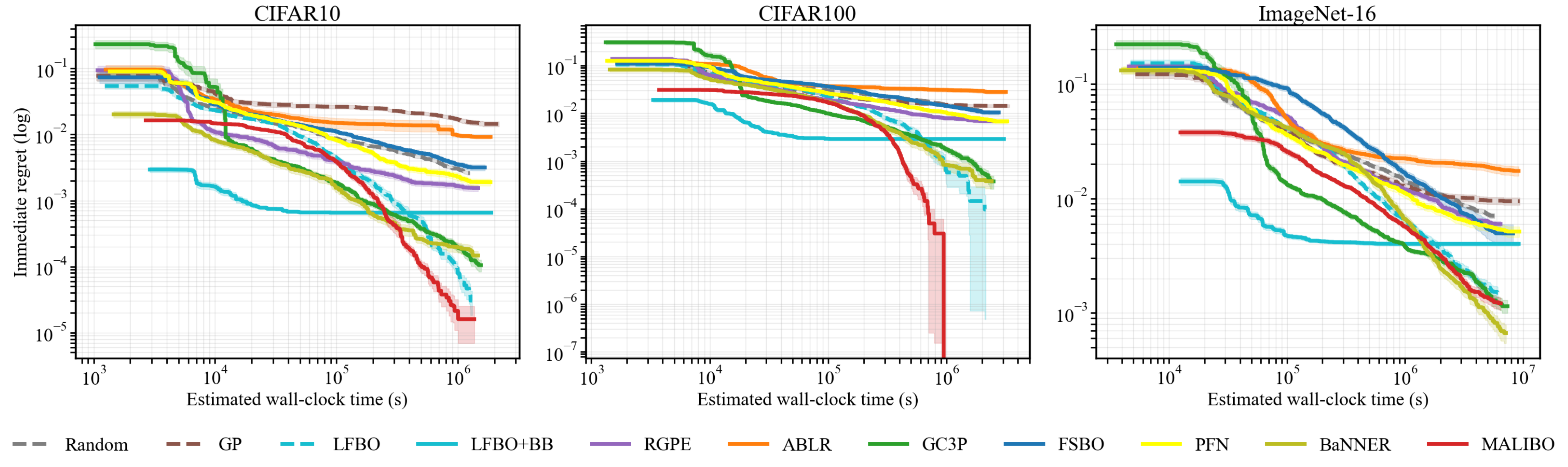}
	\caption{Immediate regrets of different BO algorithms on the NASBench201 neural network architecture search problem. Each algorithm runs for 500 iterations and we show the corresponding estimated wall-clock time on the $x$ axis in log scale.}
	\label{fig:nas_time}
\end{figure*}

In addition to the runtime, we also report results for HPOBench and NASBench201 focusing on immediate regrets as a function of the estimated wall-clock time\footnote{We have limited the runtime analysis to only HPOBench and NASBench201 because HPO-B lacks the necessary runtime information for such an evaluation.}. To obtain the realistic wall-clock time, we accumulate the time to optimize for corresponding BO methods and the recorded runtime for the configurations in the benchmarks. Notice that all the methods run for the same number of steps in an experiment. The results in \cref{fig:hpo_time,fig:nas_time} show that \ourmethod attains the best warm-starting performance across almost all benchmarks and constantly achieves one of the lowest final regrets in the same amount of time.

\subsection{Real-world benchmarks}
\label{ssec:real_world_benchmarks}
\begin{figure*}[tp]
    \centering
    \includegraphics[width=.65\linewidth]{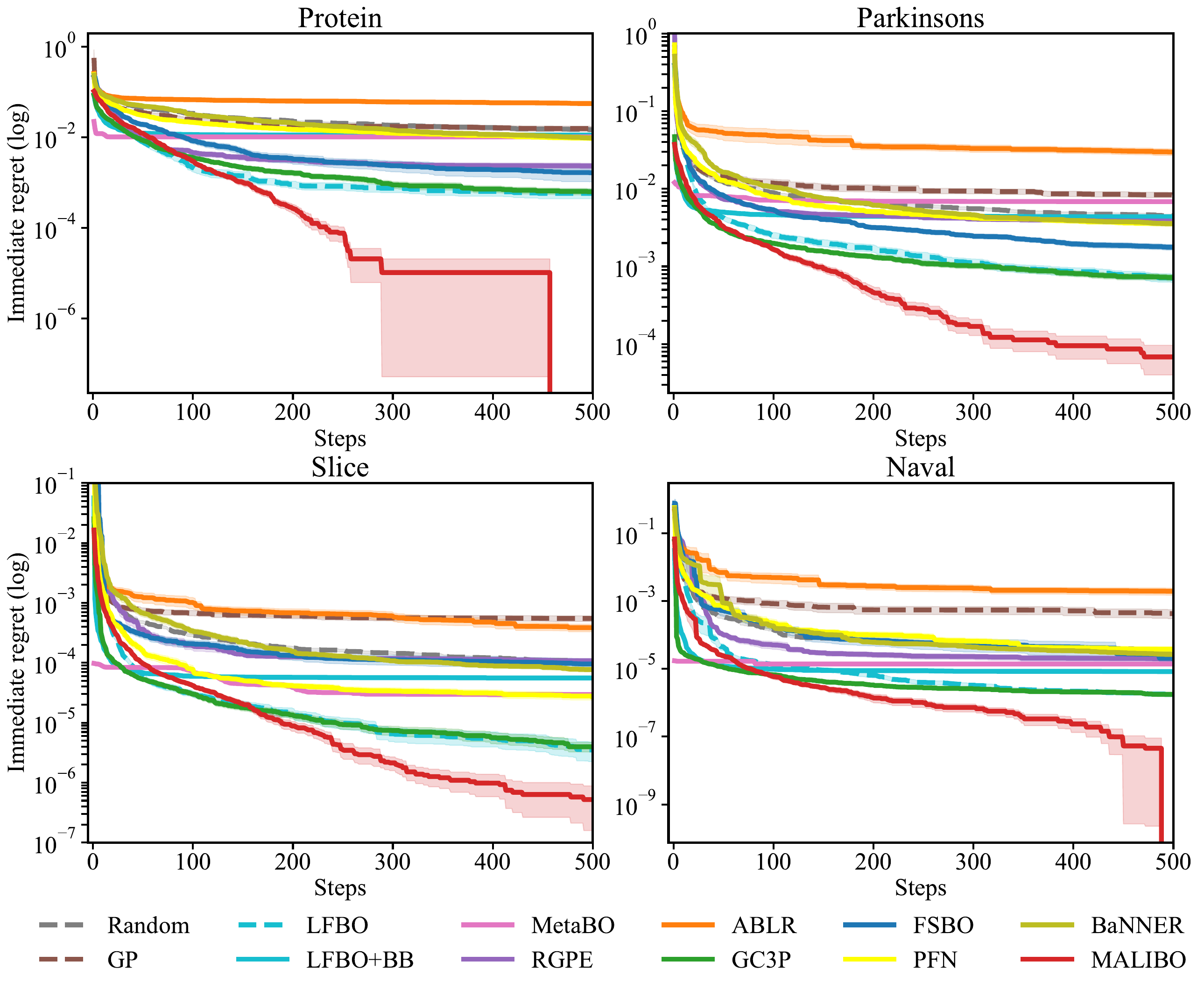}
	\caption{Immediate regrets for BO algorithms on HPOBench for \num{4} datasets.}
	\label{fig:hpo_benchmark}
\end{figure*}
\begin{figure*}[tp]
    \centering
    \includegraphics[width=.9\linewidth]{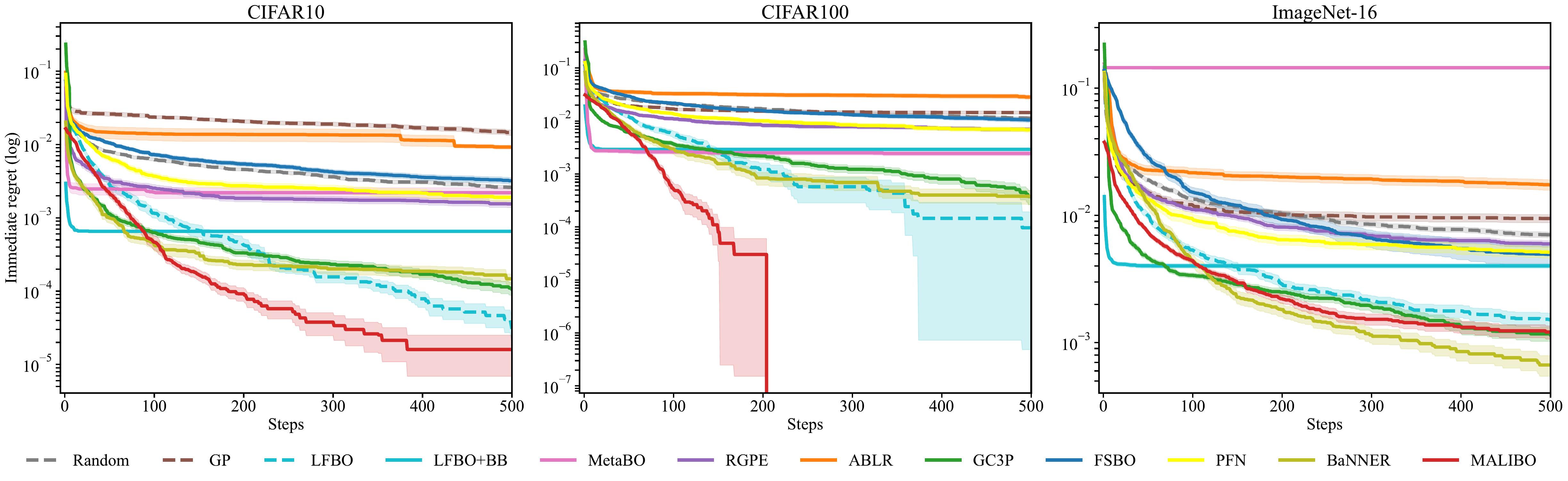}
	\caption{Immediate regrets for different BO algorithms on NASBench201 for \num{3} datasets.}
	\label{fig:nas_benchmark}
\end{figure*}
\begin{figure*}[tp]
    \begin{subfigure}{.53\textwidth}
        \centering
        \includegraphics[width=\linewidth]{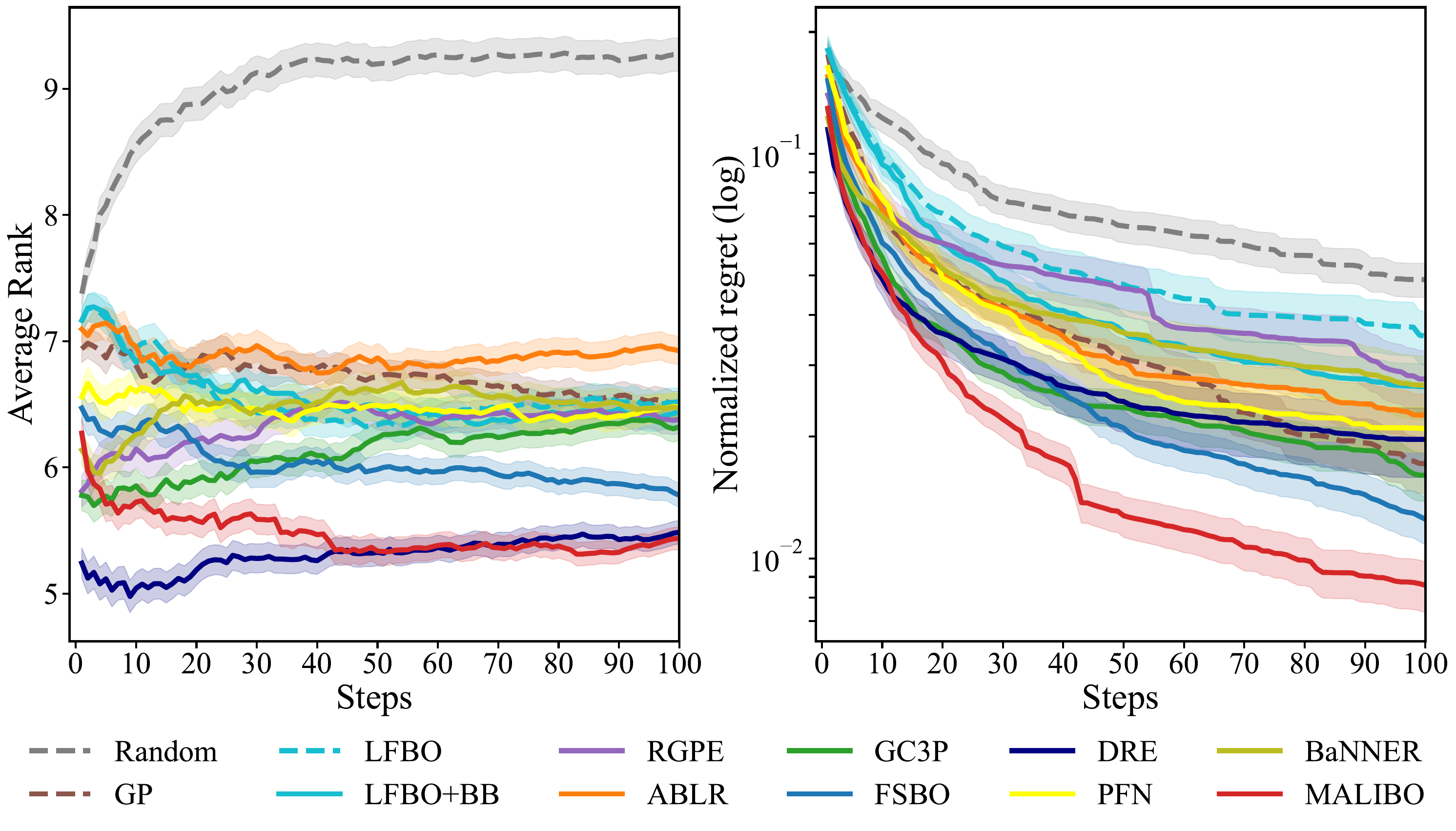}
    \end{subfigure}
    \hfill
    \begin{subfigure}{.45\textwidth}
        \includegraphics[width=.9\linewidth]{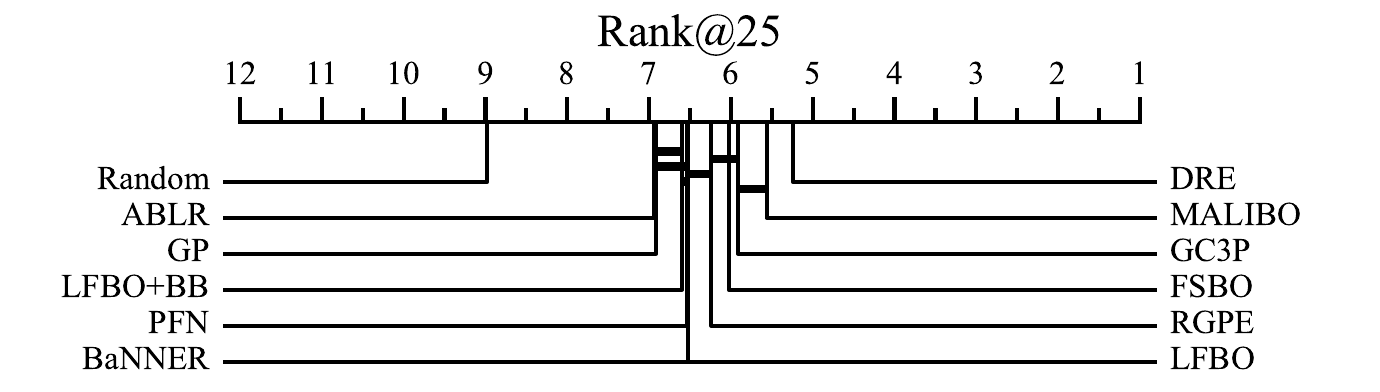}
        \includegraphics[width=.9\linewidth]{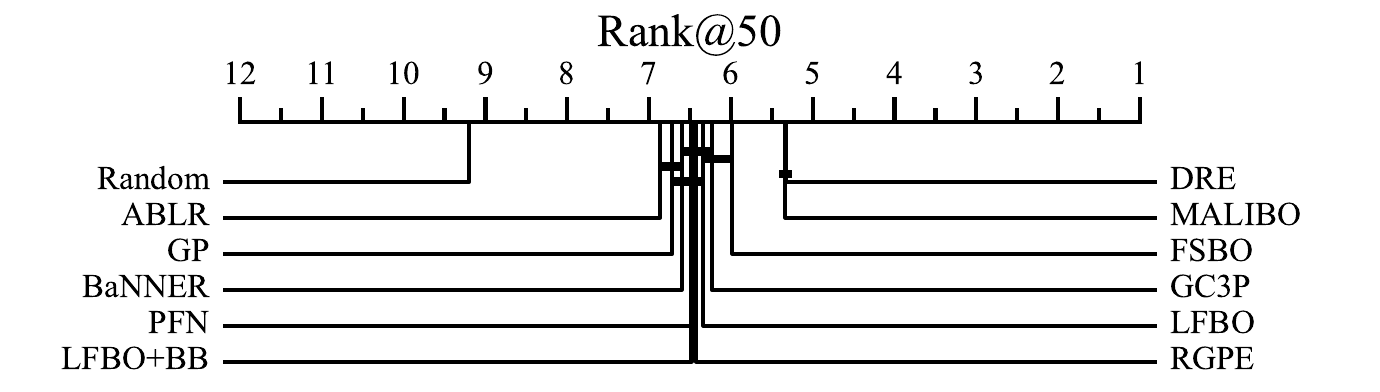}
        \includegraphics[width=.9\linewidth]{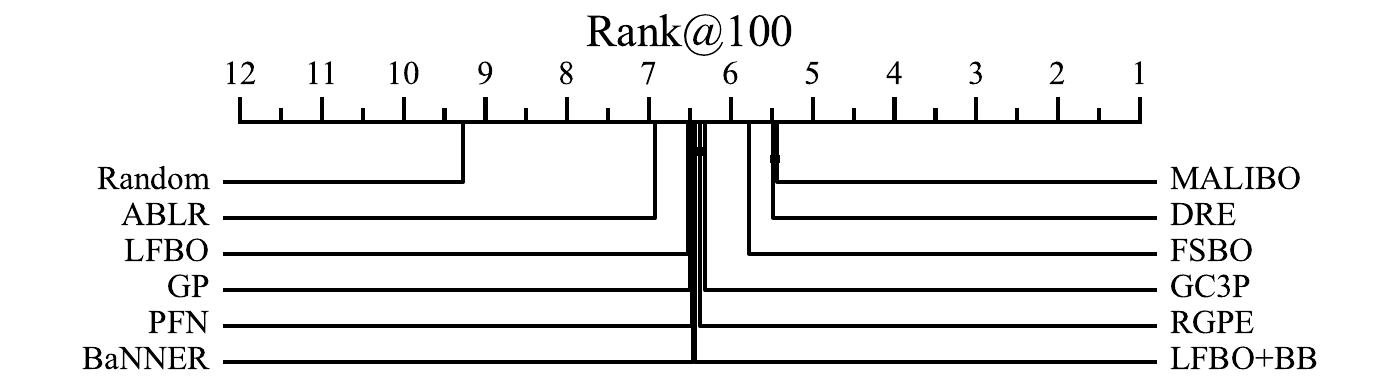}
    \end{subfigure}
    \caption{Aggregated comparisons of normalized regret and average ranks across all search spaces for BO methods on HPO-B.}
    \label{fig:hpob_benchmark}
\end{figure*}
\begin{figure*}[tp]
    \centering
    \includegraphics[width=\linewidth]{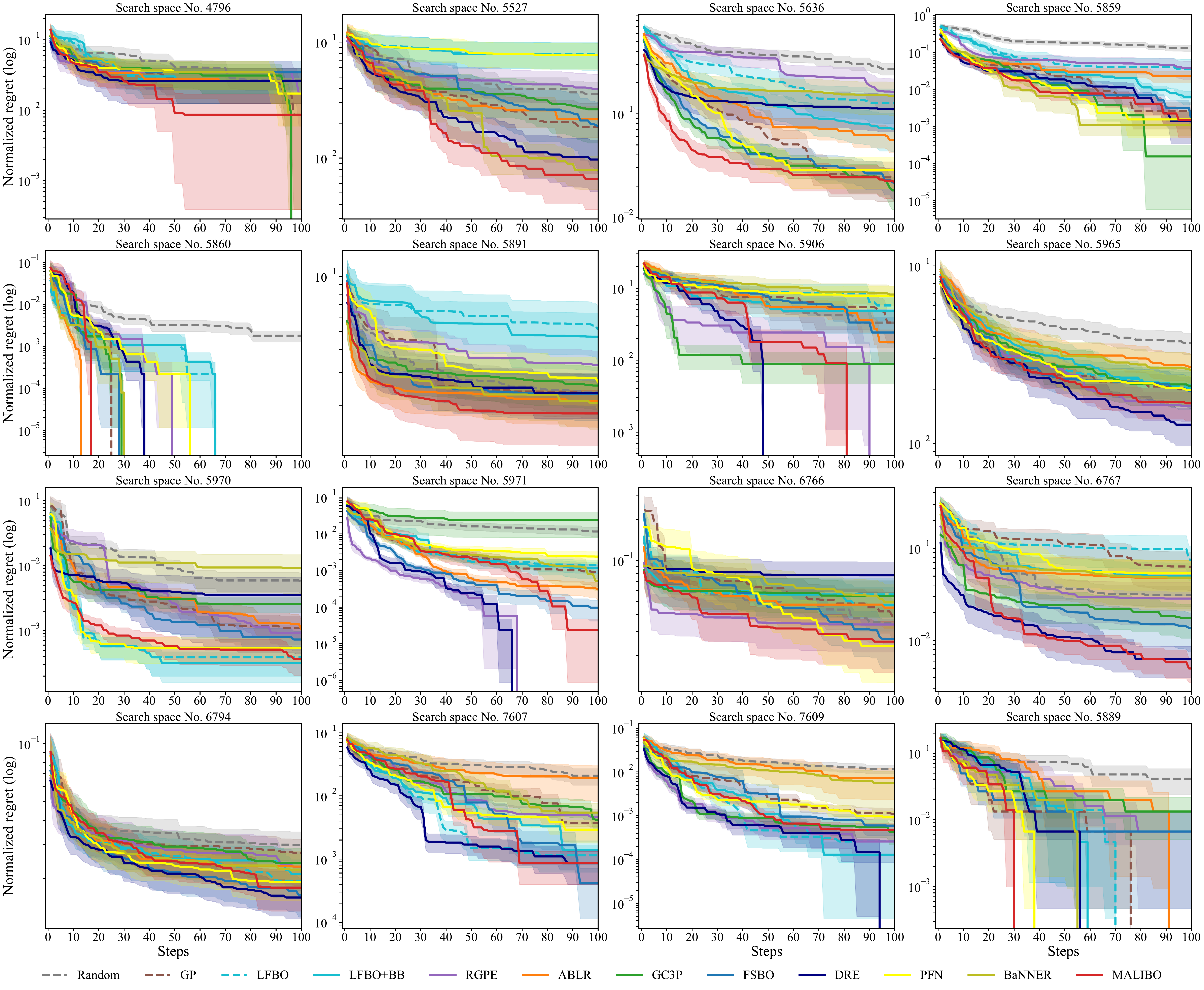}
	\caption{Normalized regret comparison of BO methods on HPO-B.}
	\label{fig:hpob_regret}
\end{figure*}
\begin{figure*}[tp]
    \centering
    \includegraphics[width= \linewidth]{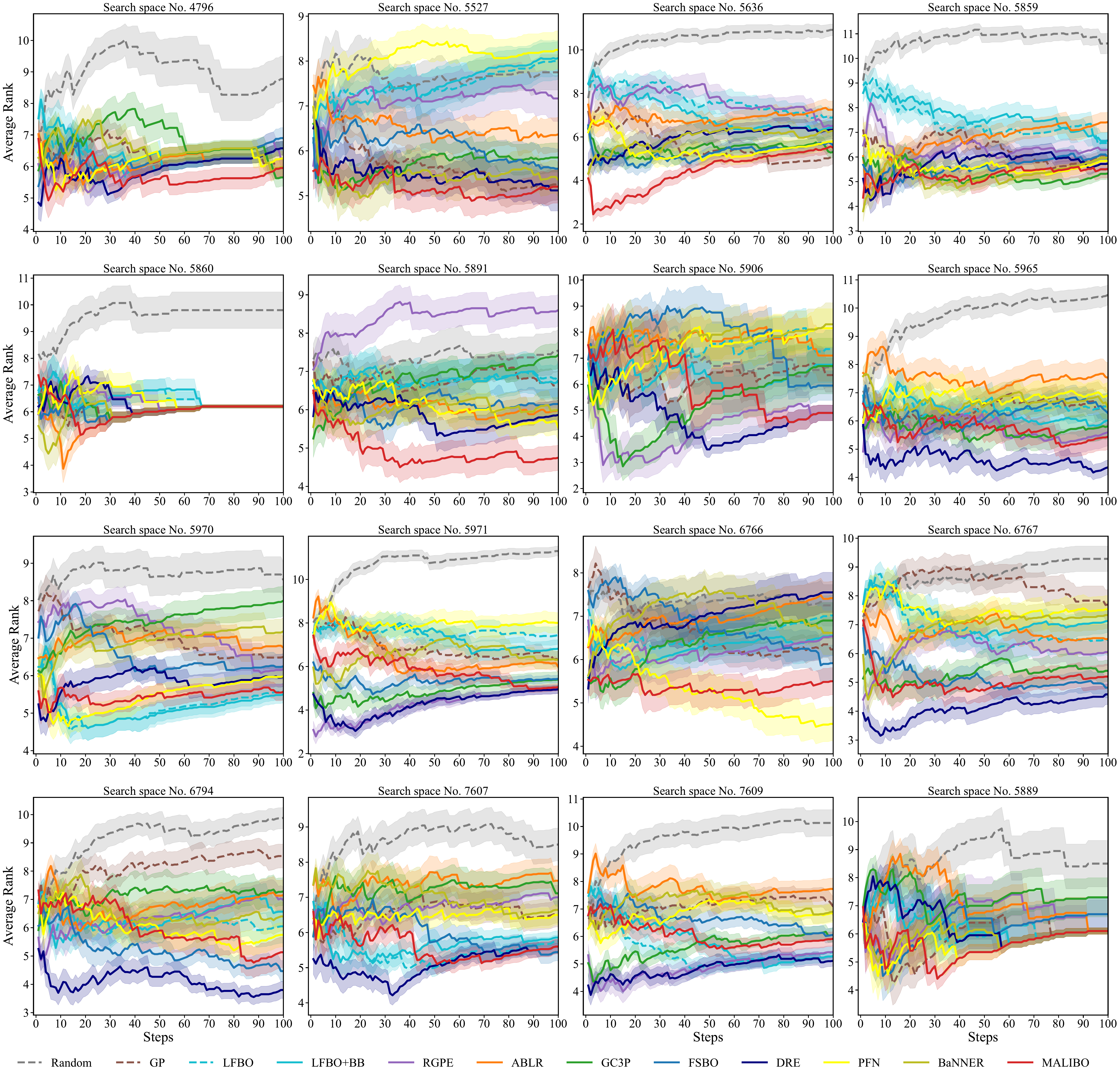}
	\caption{Average rank comparison of BO methods on HPO-B.}
	\label{fig:hpob_rank}
\end{figure*}
\begin{figure*}[tp]
    \begin{subfigure}{.53\textwidth}
        \centering
        \includegraphics[width=\linewidth]{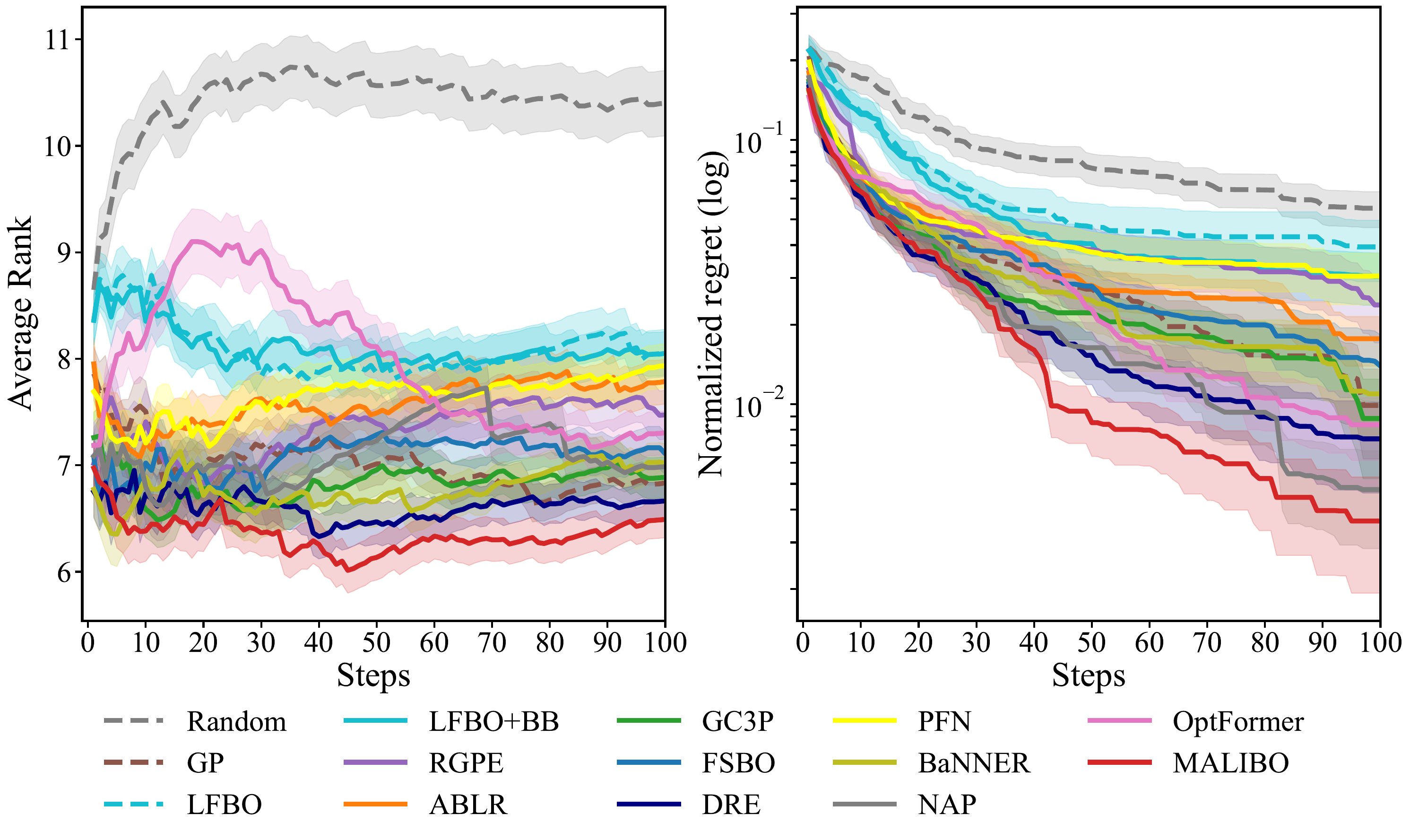}
    \end{subfigure}
    \hfill
    \begin{subfigure}{.45\textwidth}
        \centering
        \includegraphics[width=.8\linewidth]{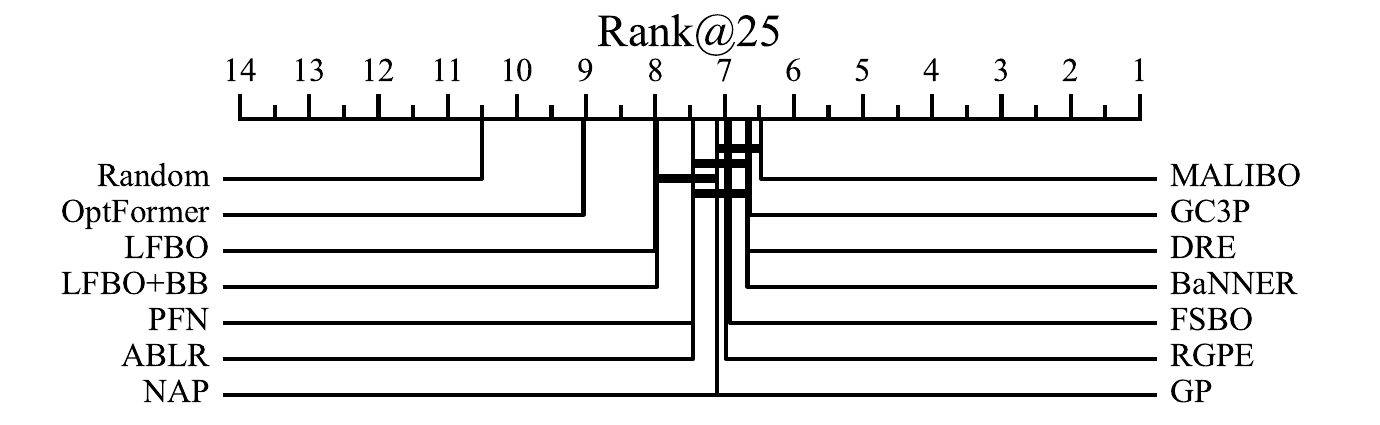}
        \includegraphics[width=.8\linewidth]{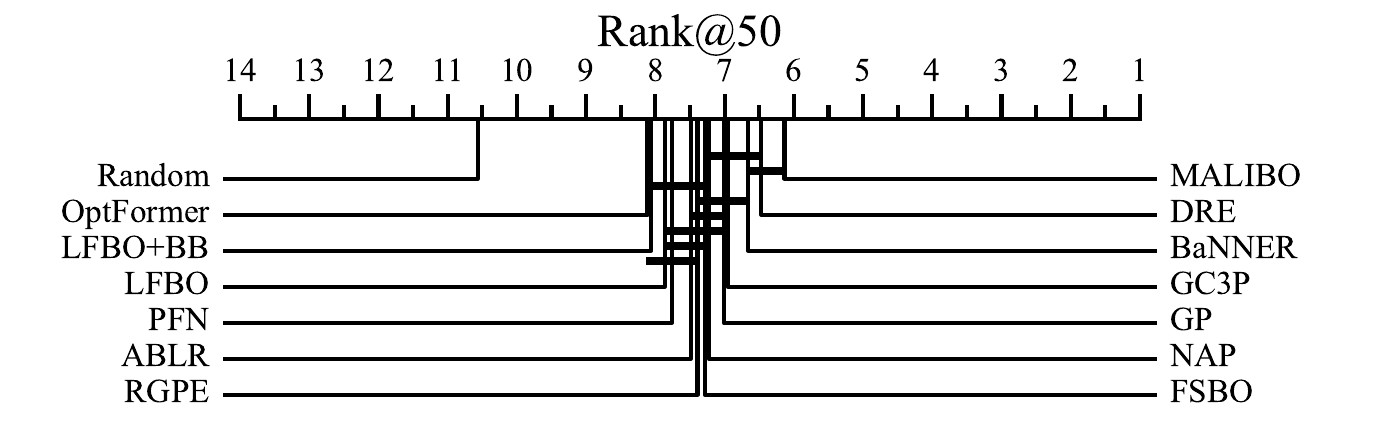}
        \includegraphics[width=.8\linewidth]{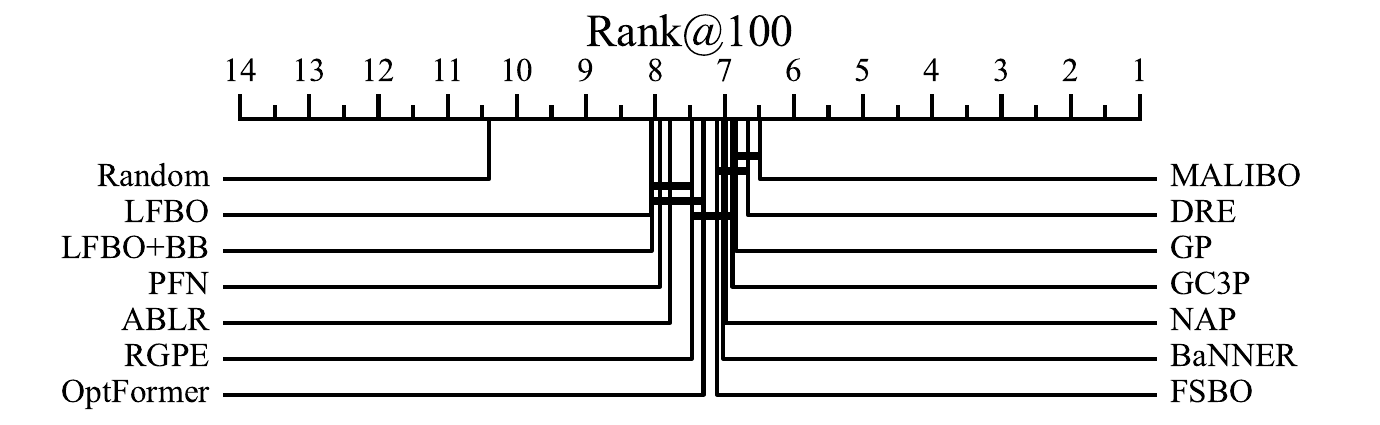}
    \end{subfigure}
    \caption{Aggregated comparisons of normalized regret and average ranks across 6 representative search spaces for BO methods on HPO-B.}
    \label{fig:nap_benchmark}
\end{figure*}
\begin{figure*}[tp]
    \centering
    \includegraphics[width=.6\linewidth]{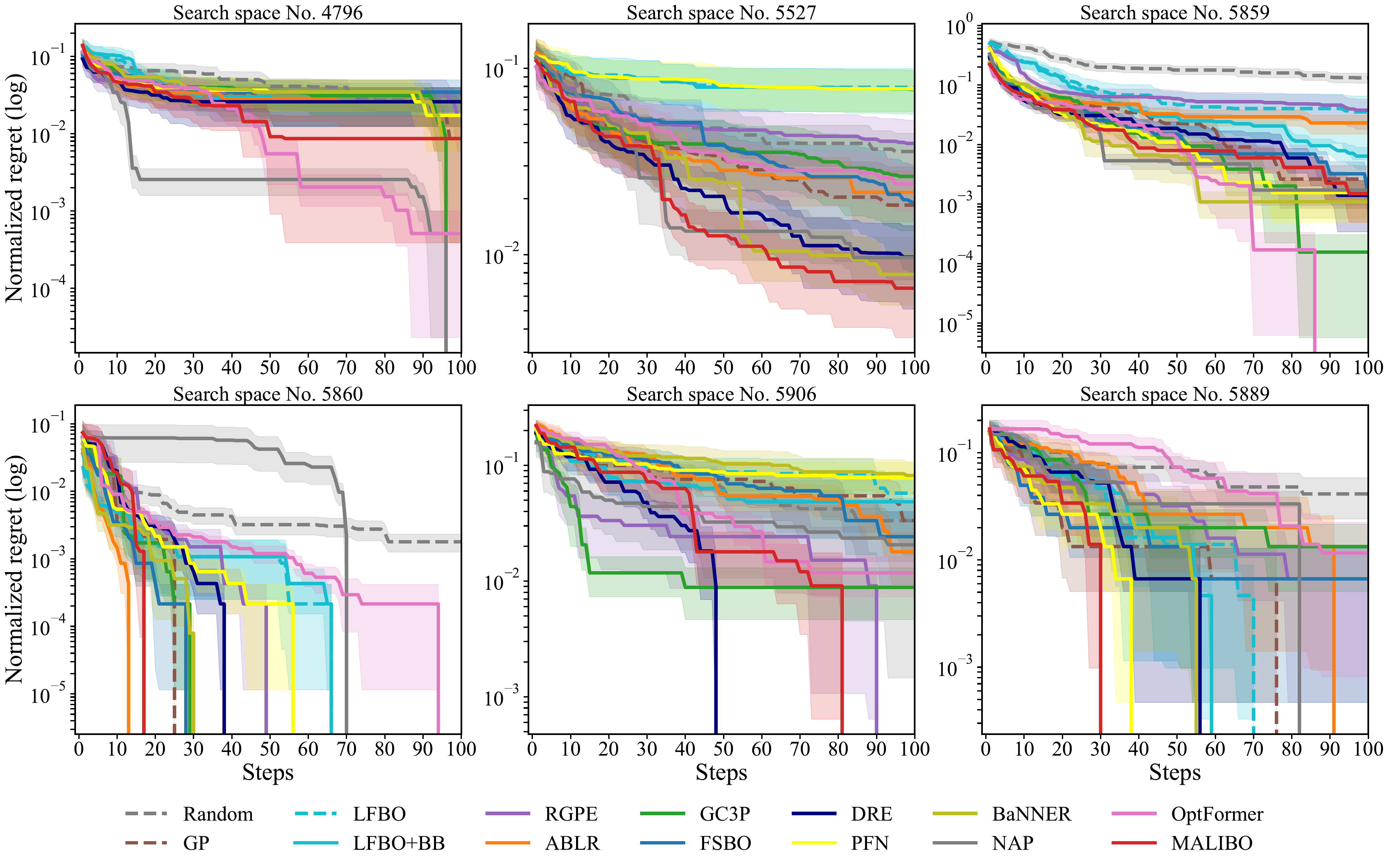}
	\caption{Normalized regret comparison of BO methods in 6 representative search space on HPO-B.}
	\label{fig:nap_regret}
\end{figure*}
\begin{figure*}[tp]
    \centering
    \includegraphics[width=.6\linewidth]{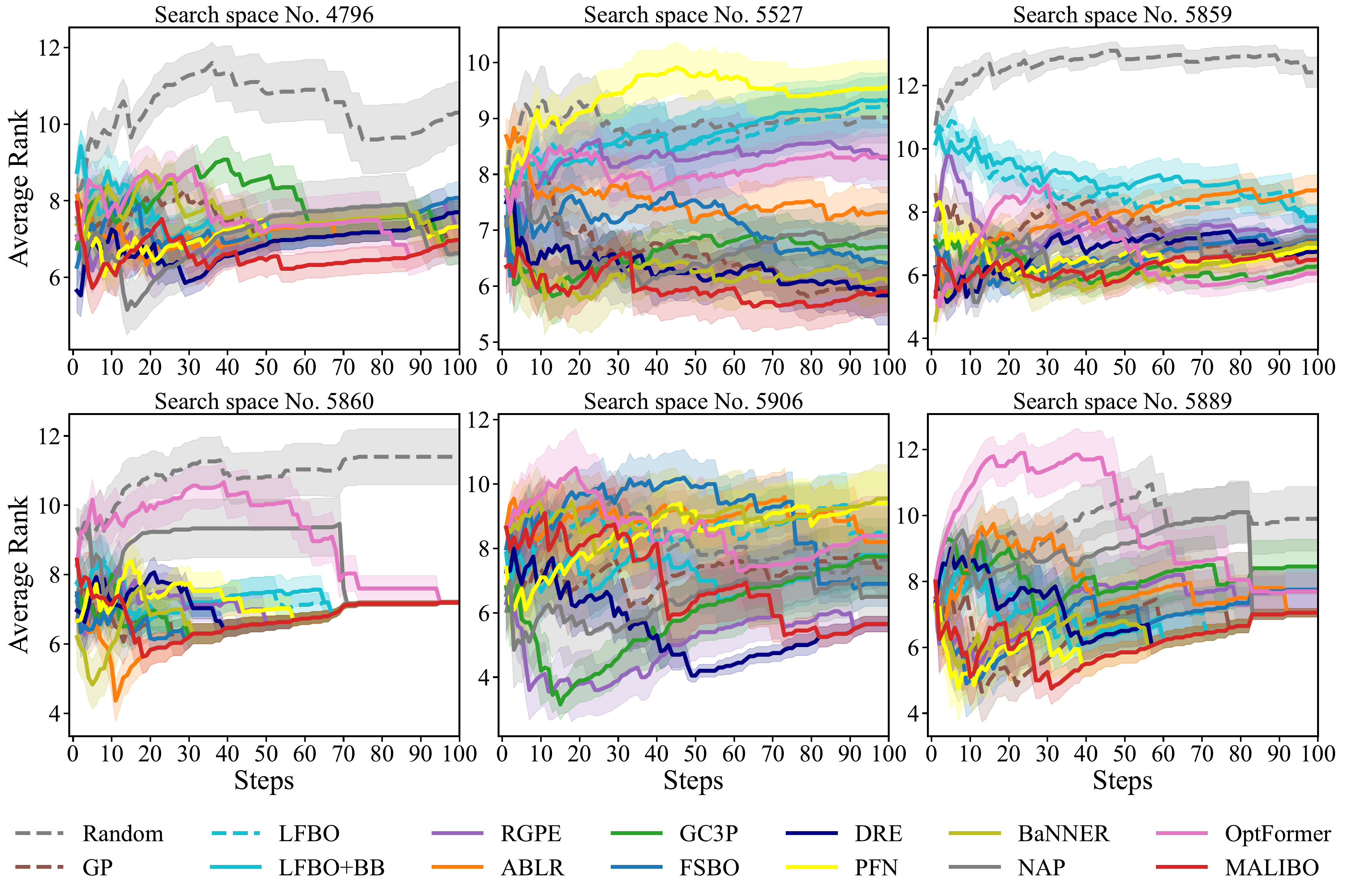}
	\caption{Average rank comparison of BO methods in 6 representative search space on HPO-B.}
	\label{fig:nap_rank}
\end{figure*}
In \cref{fig:hpo_benchmark,fig:nas_benchmark,fig:hpob_benchmark}, we demonstrate the complementary results for the real-world benchmarks that is introduced in \cref{sec:experiments}. Additionally, we report the results for two more recent baselines, namely OptFormer \cite{chen2022optformer} and NAP \cite{maraval2023nap} in \cref{fig:nap_benchmark,fig:nap_regret,fig:nap_rank}. However, due to the excessive training time of these two methods, we only compare our baselines in 6 representative search spaces in HPO-B benchmark as in \cite{maraval2023nap}.

\section{Step-through visualization}
\label{appendix:step-through-vis}
For illustration purposes, we provide step-through visualizations on a Forrester function. For details of the synthetic functions, we refer to \cref{ssec:forrester}. We use the same meta-trained model for the visualizations as the one used in \cref{ssec:features_analysis} for the corresponding problem.

\begin{figure*}[ht]
    \centering
    \includegraphics[width=.9\linewidth]{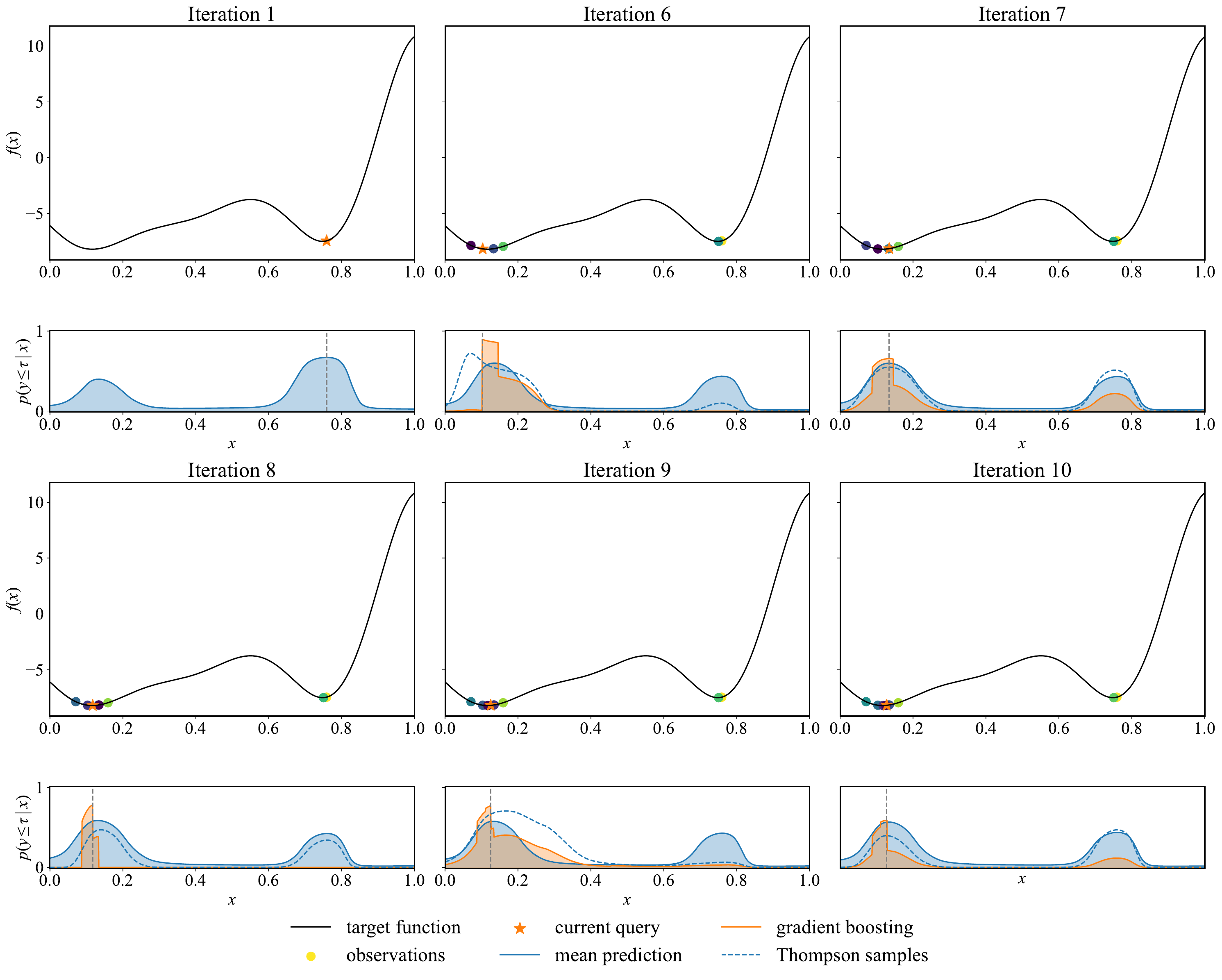}
    \caption{\ourmethod optimizing a Forrester function. We show the mean prediction, the Thompson samples of the acquisition function and the gradient boosting prediction in the lower part of each sub-figure. At the first iteration, \ourmethod picks the point with highest mean prediction of the acquisition function, which is often already close to the global optimum. Thereafter, we collect 4 more observations via the maximum prediction of a Thompson sample, in order to have sufficient data to train and apply early stopping for the gradient boosting model. Observations picked by Thompson samples show that \ourmethod explores another location of interest on the left-hand side and also area close to the true optimum. With gradient boosting, the model is still able to explore the function and the predictions on non promising area are suppressed in later iterations.
    }
    \label{fig:forrester_updates}
\end{figure*}

\paragraph{Sequential BO}
For the step-through visualization, the initial design is provided by the highest utility value of the mean predictions. After the first proposed query, we collect our the observations for the following 4 iterations using only the Thompson samples of the acquisition function. This is because we need to provide enough data to train and apply early stopping for our gradient boosting classifier. We provide the step-through visualization in \cref{fig:forrester_updates}.

\paragraph{Parallel BO with Thompson sampling}
After showing the step-through visualization for \ourmethod, we try to showcase a preliminary experiments about extending \ourmethod to parallel BO. We show a toy examples of synchronous parallel BO \citep{kandasamy2018parallelised} using \ourmethod (TS) on the same function. To be specific, we use three Thompson samples as acquisition functions in each iteration, and evaluates the three proposed points for the next optimization step. We demonstrate that, \ourmethod can be easily extended to parallel BO with the help of Thompson sampling.
\begin{figure*}[t]
    \centering
    \includegraphics[width=.9\linewidth]{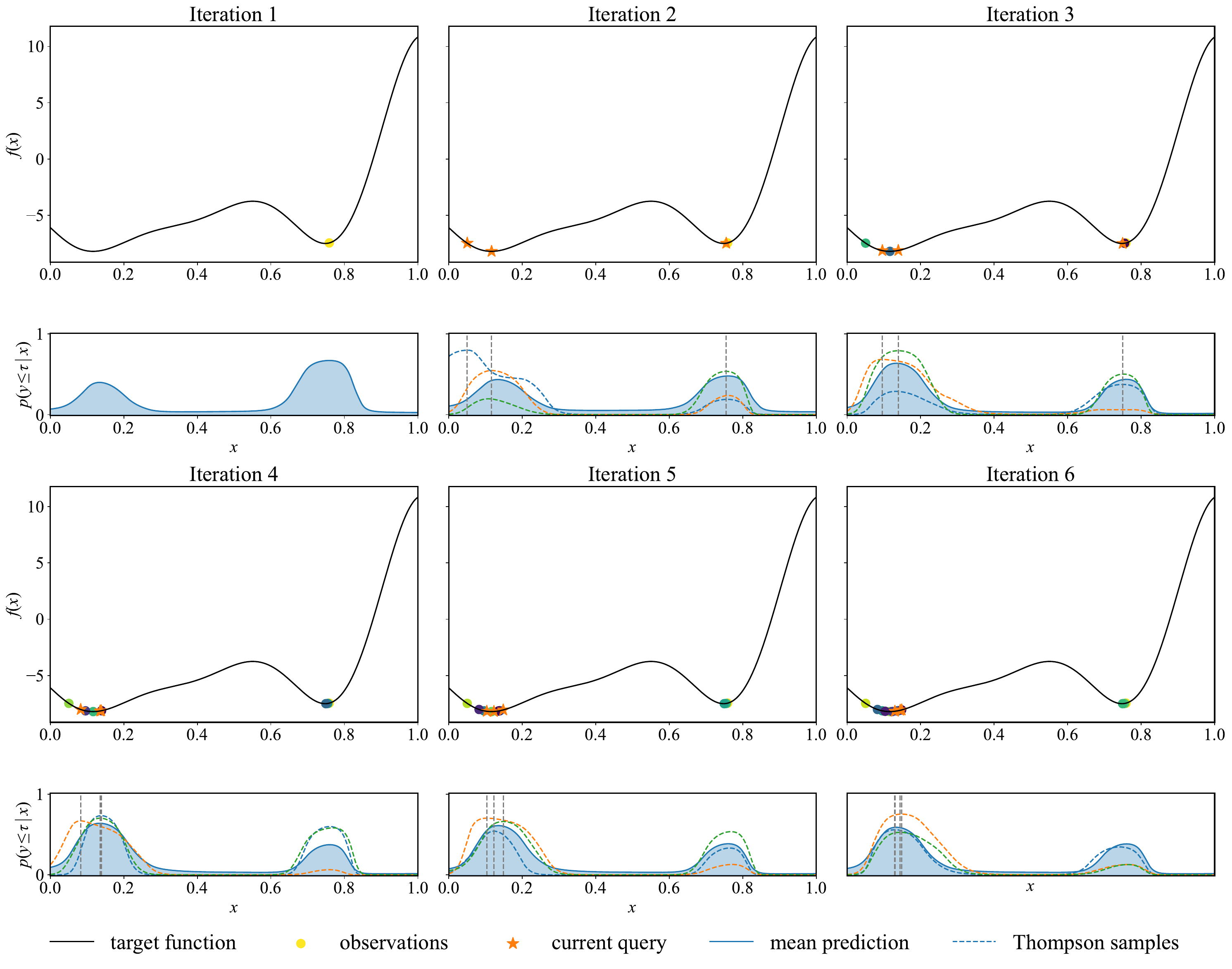}
    \caption{Synchronous parallel Thompson sampling using \ourmethod (TS) to optimize a Forrester function. At every iteration, three samples are drawn as acquisition functions and utilized to the determine the next query points. In the first iteration, \ourmethod (TS) already acquires three observations which cover both likely positions for the optimum. Subsequently, \ourmethod (TS) exploits more often around the area where the true optimum is located. At the last iteration, all of the three Thompson samples have already been skew toward the left-hand side, which shows \ourmethod (TS) converges to the correct region.}
    \label{fig:parallel_forrester}
\end{figure*}

\section{Experimental details}
\label{appendix:experimental_deatails}
In this section, we explain the setups for all baselines we used in the experiments. We ran all baselines on 4 CPUs (Intel(R) Xeon(R) Gold 6150 CPU @ 2.70GHz) except for MetaBO, which requires more computation and we explain the details down below. Since most baselines are tested on similar AutoML tasks, we keep their hyperparameters as in their official implementations (if available) with minor modifications. The hyperparameters of \ourmethod is selected by validating on the synthetic functions. We then fixed the selected hyperparameters for all experiments to ensure it is not overfit on a specific task. We will elaborate the experimental details in the following.

\paragraph{GP}
We use the SingleTaskGP implementation from BoTorch\footnote{\url{https://github.com/pytorch/botorch/}} with Mat\'ern $5/2$ kernel and Expected Improvement (EI) as acquisition function.

\paragraph{LFBO}
Our implementation of LFBO is based on the official repository\footnote{\url{https://github.com/lfbo-ml/lfbo}} from \citet{song2022general} and we use gradient boosting from scikit-learn \citep{pedregosa2011scikitlearn} as the classifier with the following settings: $\text{n\_estimator} = 100$, $\text{learning rate} = 0.1$, $\text{min\_samples\_split}=2$, $\text{min\_samples\_leaf}=1$. For each problem, LFBO first randomly samples \num10 observations to gather information and thereafter perform optimization using the classifier. For the threshold $\gamma$, which trade-off the exploration and exploitation, we set $\gamma=1/3$ following \citet{song2022general} for all experiments. To maximize the resulting acquisition function, we use random search with $5,120$ samples following \citet{tiao2021bore}, where they show that, the acquisition function is usually non-smooth and discontinuous for decision trees based method and using random search is on par or even outperforms the more expensive alternative evolutionary algorithm.

\paragraph{\lfbobb}
We extend LFBO to a meta-learning method with bounding box search space pruning \citep{perrone2019searchspace}, which reduces the search space based on the promising configurations in the related tasks. Our implementation of the search space pruning technique is based on the open-source implementation in Syne Tune\footnote{\url{https://github.com/awslabs/syne-tune/}}. To construct the bounding box, we select the top-$1$ performing configurations from each related task, and truncate the search space according to the these configurations. We then apply LFBO to optimize the target task in the pruned search space.

\paragraph{RGPE} 
Our implementation of Ranking-weighted GP Ensemble (RGPE) is based on \citet{feurer2022practical}. The key idea behind the algorithm is that, for optimization, the important information predicted by the surrogate model is not so much the function value $f(\mathbf x)$ at a given input $\mathbf x$, but rather if the value $f(\mathbf x)$ is larger or smaller relative to the function evaluated at other inputs. In other words, whether $f(\mathbf x)>f(\mathbf x')$ or vice versa. The algorithms propose to fit separate GPs on different tasks and use a ranking strategy to combine the model that fit the most on the target task with prior knowledge. Our implementation uses radidal basis function (RBF) kernel for each GP and Upper Confidence Bound (UCB) \citep{srinivas2010ucb} with $\beta=9.0$ as the acquisition function.

\paragraph{ABLR}
BO with multi-task adaptive Bayesian linear regression. Our implementation of ABLR is equivalent to a GP with 0 mean and a dot-product kernel with learned basis functions. We use a neural net (NN) with two hidden layers, each having 50 units, and Tanh activation as the basis functions. We train ABLR by optimizing the negative log likelihood (NLL) over NN weights and covariance matrix that define the dot-product kernel. In each iteration, ABLR is trained on all data including the meta-data as well as the observations from the target task using L-BFGS \citep{byrd1995lbfgs}.

\paragraph{GC3P}
We use the open-source implementation\footnote{\url{https://github.com/geoalgo/A-Quantile-based-Approach-for-Hyperparameter-Transfer-Learning}} for GC3P from \citet{salinas2020quantile}. For each target function, GC3P first samples five candidates from a meta-learned NN model before building a task-specific Copula process.

\paragraph{BaNNER}
Similar to \ourmethod, BaNNER uses a task-agnostic component to learn a feature transformation across tasks for the mean prediction and a task-specific component to predict the residual. The difference is that it uses Bayesian linear regression in the final layer, akin to the approach described in ABLR. We implemented the BaNNER-BLR-GP variant as detailed by \citet{berkenkamp2021probabilistic}, which integrates an additive non-parametric GP model to account for residual errors. In our implementation, BaNNER utilizes a  a three-layer ResFNN, each layer comprising 32 units, to learn the feature mapping function. We set the task embedding dimension to 16 and employed EI as the acquisition function.

\paragraph{FSBO}
We use the open-source implementation\footnote{\url{https://github.com/releaunifreiburg/FSBO}} for FSBO from \citet{wistuba2021fewshot}. Its idea is similar to ABLR, which aims to learn a task-independent deep kernel surrogate and allow a task-dependent head for adaptation. The difference is that it frames meta-learning BO as a few-shot learning problem, which means during training, each batch from stochastic gradient ascent contains only data from one task. For the experiments, we use the default setting from the official implementation.

\paragraph{DRE}
We use the open-source implementation\footnote{\url{https://github.com/releaunifreiburg/DeepRankingEnsembles}} for DRE from \citet{khazi2023deep}. As this implementation only provides pipelines and hyperparameter settings for the HPO-B benchmark, we confined our reporting to this benchmark to ensure a fair comparison.

\paragraph{PFN}
We use the open-source implementation\footnote{\url{https://github.com/automl/PFNs4BO}} for PFN from \citet{muller2023pfns4bo}.

\paragraph{NAP and OptFormer}
Due to the excessive training time, we use their results of HPO-B from the official repository\footnote{\url{https://github.com/huawei-noah/HEBO/tree/master/NAP}}.

\paragraph{MetaBO}
The training and evaluation of MetaBO was done based on the official implementation of \citet{volpp2020metalearning}. We followed the recommended hyperparameters and model architecture from the implementation with the following changes:
\begin{enumerate}
    \item We extended the training horizon to 60 steps, which is longer than the ones used by \citet{volpp2020metalearning}. We chose the longer episode length to adapt to the higher dimensional space and the evaluation horizon we chose for the benchmarks. The value 60 was chosen as a compromise between the full evaluation length of 500 iterations and the resulting increase in training time due to the scaling of the GP used in the method.
    \item We did not include the current time-step and total budget as features to the neural network policy due to poor performance on our benchmarks when including them.
    \item Due to the small number of data sets, we estimated the GP hyperparameters with independent sub-samples of the meta-data sets, but otherwise following the procedure of \citet{volpp2020metalearning}. This effectively gives MetaBO access to more meta-data, but is consistent with the evaluation scheme described below.
\end{enumerate}
Besides these changes to the method, we also employed a different evaluation scheme for MetaBO due to its high training cost. In contrast to the other meta-learning models that train in minutes on the meta-data using a single CPU, MetaBO required almost 2 hours, using one NVIDIA Titan X GPU and 10 Intel(R) Xeon(R) CPU E5-2697 v3 CPUs. This made the independent meta-training across the individual runs infeasible. To still include MetaBO into some of our benchmarks, we decided to train MetaBO once and reuse this model throughout the individual runs during the evaluation.

During the meta-training, MetaBO received the same number of samples per meta-task as the other methods, but the subsample was resampled for each training episode. While this gave MetaBO access to more meta-data compared to the other methods, we eliminated the risk of evaluating the method on a bad subsample of the data by chance. The advantage of effectively seeing more points of each meta-tasks should be considered when evaluating the early performance of MetaBO compared to the other methods. The evidently weak adaptation of MetaBO to new tasks dissimilar to the meta-data.

Based on the high meta-training cost of MetaBO and the relatively poor performance on NASBench201 and the HPOBench benchmarks, we decided to not include the method for the other evaluations, as scheme of leave-one-task-out validation would be too expensive and any other comparison would either benefit MetaBO or put it at a disadvantage rendering the results difficult to interpret.

\paragraph{\ourmethod}
We use a Residual Feed Forward Network (ResFFN) \citep{he2016resnet} for learning the latent feature representation, with \num{4} hidden layers, each with \num{64} units. For the mean prediction layer $m(\cdot)$ and task-specific layer $h_{\mathbf{z}_t}(\cdot)$, we use a fully connected layer with \num{50} units for each. The resulting meta-leaning model has $22,359$ learnable parameters. We use ELU \citep{clevert2016elu} as the activation function in the network following \citet{tiao2021bore}. Similar to LFBO, we set the threshold $\gamma=1/3$ and maximize the acquisition function using random search with $5,120$ samples..

During meta-training, we optimize the parameters in the network with the ADAM optimizer \citep{kingma2015adam}, with learning rate $\text{lr} = 10^{-3}$ and batch size of $B=256$. In addition, we apply exponential decay to the learning rate in each epoch with factor of $0.999$. The model is trained for $2,048$ epochs with early stopping. For the regularization loss, we set the regularization factor $\lambda = 0.1$ in \cref{eq:negative_log_likelihood} and follow the approach in \cref{sec:latent_reg} to estimate the coefficients $\lambda_{\text{KS}}$ and $\lambda_{\text{Cov}}$. The resulting meta-training is fast and efficient and we show the training time as well as the amount of meta-data for each benchmark in \cref{tab:meta_data_and_time}.

In task adaptation, we optimize the task embedding for the target task using L-BFGS \citep{byrd1995lbfgs}, with learning rate $\text{lr} = 1 $, maximal number of iterations per optimization step $\text{max\_iter}=20$, termination tolerance on first order optimality  $\text{tolerance\_grad}=10^{-7}$,  termination tolerance on function value/parameter changes $\text{tolerance\_change}=10^{-9}$, $\text{history\_size}=100$ and using strong-wolfe as line search method. After obtaining the model adapted on target task, we combine it with a gradient boosting classifier, which serves as a residual prediction model. We use the same setting for the gradient boosting as in \lfbo, except that we use the meta-learned \ourmethod classifier as the initial estimator. However, the gradient boosting classifier is trained only on the observations generated by the optimization process, which might lead to overfitting on limited amount of data during early iterations. Therefore, we apply early stopping to avoid such behavior. Specifically, we first estimate the number of trees that we need for training without overfitting. This is done by fitting a gradient boosting classifier with randomly chosen training data and validation data, which account for $70\%$ and $30\%$ of the whole data respectively. The resulting classifier estimates the number of trees that are needed to fit the partially observed data while offering good generalization ability. We then use it as our hyperparameter for the gradient boosting and train it on all observations.

\section{Details of benchmarks}
\label{appendix:benchmark_details}

\begin{table}[tp]
	\caption{Meta-data and training time}
	\label{tab:meta_data_and_time}
	\centering

	\vspace{1mm}
	\begin{tabular}{ p{2cm} p{2cm} p{2.5cm}}
		\toprule
		Benchmark & \# meta-data  & Training time (approximate)\\
		\midrule
		HPOBench      & $1,536$     &  15 seconds\\
		NASBench201   & $1,024$     &  15 seconds\\
		Branin        & $32,768$    &  480 seconds\\
		Hartmann3D     & $131,072$   &  1,800 seconds\\
		\bottomrule
	\end{tabular}
\end{table}
\subsection*{HPOBench}
\label{ssec:hpobench}
The hyperparameters for HPOBench and their ranges are demonstrated in \cref{tab:hpobench}. All hyperparameters are discrete and there are in total \num{66208} possible combinations. More details can be found in \citet{klein2019tabular}.
\begin{table*}[ht]
	\caption{Configuration spaces for HPOBench}
	\label{tab:hpobench}
	\centering

	\vspace{1mm}
	\begin{tabular}{ p{3cm} p{10cm} }
		\toprule
		Hyperparameter & Range\\
		\midrule
		Initial LR           & \{ $5 \times 10^{-4}$, $1 \times 10^{-3}$, $5 \times 10^{-3}$, $1 \times 10^{-2}$, $5 \times 10^{-2}$, $1 \times 10^{-1}$ \}\\
		LR Schedule           & \{ cosine, fixed \}\\
		Batch size            & \{ $2^3$, $2^4$, $2^5$, $2^6$ \}\\
		Layer 1  Width        & \{ $2^4$, $2^5$, $2^6$, $2^7$, $2^8$, $2^9$ \}\\
		\qquad \quad \, Activation   & \{ relu, tanh \}\\
		\qquad \quad \,  Dropout rate & \{ $0.0$, $0.3$, $0.6$ \}\\
		Layer 2  Width        & \{ $2^4$, $2^5$, $2^6$, $2^7$, $2^8$, $2^9$ \}\\
		\qquad \quad \,  Activation   & \{ relu, tanh \}\\
		\qquad \quad \,  Dropout rate & \{ $0.0$, $0.3$, $0.6$ \}\\
		\bottomrule
	\end{tabular}
\end{table*}

\subsection*{NASBench201}
\label{ssec:nasbench}
The hyperparameters for NASBench201 and their ranges are summarized in \cref{tab:nasbench}. All hyperparameters are discrete and there are in total \num{15625} possible combinations. More details can be found in \citet{dong2020nasbench}.
\begin{table*}[ht]
	\caption{Configuration spaces for NASBench201}
	\label{tab:nasbench}
	\centering
	\vspace{1mm}
	\begin{tabular}{ p{3cm} p{10cm} }
		\toprule
		Hyperparameter & Range\\
		\midrule
		ARC 0 & \{ none, skip-connect, conv-$1 \times 1$, conv-$3 \times 3$, avg-pool-$3 \times 3$ \}\\
		ARC 1 & \{ none, skip-connect, conv-$1 \times 1$, conv-$3 \times 3$, avg-pool-$3 \times 3$ \}\\
		ARC 2 & \{ none, skip-connect, conv-$1 \times 1$, conv-$3 \times 3$, avg-pool-$3 \times 3$ \}\\
		ARC 3 & \{ none, skip-connect, conv-$1 \times 1$, conv-$3 \times 3$, avg-pool-$3 \times 3$ \}\\
		ARC 4 & \{ none, skip-connect, conv-$1 \times 1$, conv-$3 \times 3$, avg-pool-$3 \times 3$ \}\\
		ARC 5 & \{ none, skip-connect, conv-$1 \times 1$, conv-$3 \times 3$, avg-pool-$3 \times 3$ \}\\
		\bottomrule
	\end{tabular}
\end{table*}

\subsection*{HPO-B}
\label{ssec:mlbench}
We use the HPO-B-v3 in our experiments and provide its description of search spaces in \cref{tab:hpo-b}. More details can be found in \citet{pineda2021hpob}.
\begin{table*}[ht]
	\caption{Description of the search spaces in HPO-B-v3. \#HPs stands for the number of hyperparameters, \#Evals. for the number of evaluations in a search space, while \#DS for the number of datasets across which the evaluations are collected. The search spaces are named with the respective OpenML version number (in parenthesis).}
	\label{tab:hpo-b}
	\centering

	\vspace{1mm}
     \begin{tabular}{lrrrcrcrc}
        \toprule
        \multirow{2}{*}{Search Space} & \multirow{2}{*}{ID} & \multirow{2}{*}{\#HPs} & \multicolumn{2}{c}{Meta-Train} & \multicolumn{2}{c}{Meta-Validation} & \multicolumn{2}{c}{Meta-Test} \\\cline{4-9}
                                      &                     &                        & \#Evals. & \#DS & \#Evals. & \#DS & \#Evals. & \#DS \\ \hline
        rpart.preproc (16)            & 4796                & 3                      & 10694    & 36   & 1198     & 4    & 1200     & 4 \\
        svm (6) & 5527 & 8 & 385115 & 51 & 196213 & 6 & 354316 & 6\\
        rpart (29) & 5636 & 6 & 503439 & 54 & 184204 & 7 & 339301 & 6\\
        rpart (31) & 5859 & 6 & 58809 & 56 & 17248 & 7 & 21060 & 6 \\
        glmnet (4) & 5860 & 2 & 3100 & 27 & 598 & 3 & 857 & 3\\
        svm (7) & 5891 & 8 & 44091 & 51 & 13008 & 6 & 17293 & 6\\
        xgboost (4) & 5906 & 16&  2289 & 24 & 584 & 3 & 513 & 2\\
        ranger (9) & 5965 & 10&  414678 & 60 & 73006 & 7 & 83597 & 7\\
        ranger (5) & 5970 & 2 & 68300 & 55 & 18511 & 7 & 19023 & 6\\
        xgboost (6) & 5971 & 16&  44401 & 52 & 11492 & 6 & 19637 & 6\\
        glmnet (11) & 6766 & 2 & 599056 & 51 & 210298 & 6 & 310114 & 6\\
        xgboost (9) & 6767 & 18&  491497 & 52 & 211498 & 7 & 299709 & 6\\
        ranger (13) & 6794 & 10&  591831 & 52 & 230100 & 6 & 406145 & 6\\
        ranger (15) & 7607 & 9 & 18686 & 58 & 4203 & 7 & 5028 & 7\\
        ranger (16) & 7609 & 9 & 41631 & 59 & 8215 & 7 & 9689 & 7\\
        ranger (7) & 5889 & 6 & 1433 & 20 & 410 & 2 & 598 & 2\\
        \bottomrule
    \end{tabular}
\end{table*}

\subsection*{The Quadratic Ensemble}
\label{ssec:quadratic}
The function for the quadratic ensemble is defined as:
\begin{equation}
    f(x, a, b, c) = (a \cdot (x-b))^2 - c \qquad x \in [0, 1]
\end{equation}
To form the ensemble, we choose the distribution for the parameters as:
\begin{equation}
    a \sim \mathcal{U}(0.5, 1.5) \quad b \sim \mathcal{U}(-0.9, 0.9) \quad c \sim \mathcal{U}(-1, 1)
\end{equation}
This distribution of parameters ensures that the search space contains the minimum of the quadratic function at $x^{*} = b$ with $f(x^{*}) = c$. The location of the optimum has a broad distribution over the function space, which is intended to highlight algorithms that learn the global structure of the ensemble rather than restricting on some small regions of interest.

\subsection*{The Forrester Ensemble}
\label{ssec:forrester}
The original Forrester function \citep{sobester2008engineering} is defined following:
\begin{multline}
f(x, a, b, c) = a \cdot (6x-2)^2 \dot \sin(12x-4) + b(x-0.5)-c\,,\\ \qquad x \in [0, 1]    
\end{multline}

The function has one local and one global minimum, and a zero-gradient inflection point in the domain $x \in [0,1]$. 
To form the ensemble, we choose the distribution for the parameters as:
\begin{equation}
    a \sim \mathcal{U}(0.2, 3) \quad b \sim \mathcal{U}(-5, 15) \quad c \sim \mathcal{U}(-5, 5)
\end{equation}
Let $\tau = \{ a, b, c \}$ and $p(\tau)$ is a three dimensional uniform distribution. The ranges are chosen around the usually used fixed values for the parameters, namely $a=0.5$, $b=10$, $c=-5$.

\subsection*{The Branin Ensemble}
\label{ssec:branin}
The function for the Branin ensemble is the following:
\begin{multline}
    f(x, a, b, c) = a (x_2 - b x_1^2 + c x_1 - r) + s(1 - t) \cos(x_1) + s\,, \\ \qquad x_1 \in [-5, 10], x_2 \in [0, 15]
\end{multline}
The distribution for the parameters are chosen as:
\begin{equation}
\begin{split}
    &a \sim \mathcal{U}(0.5, 1.5) \quad b \sim \mathcal{U}(0.1, 0.15) \quad c \sim \mathcal{U}(1.0, 2.0) \\
    &r \sim \mathcal{U}(5.0, 7,0) \quad s \sim \mathcal{U}(8.0, 12.0) \quad t \sim \mathcal{U}(0.03, 0.05)
\end{split}
\end{equation}
Let $\tau = \{ a, b, c, r, s, t\}$ and $p(\tau)$ is a six dimensional uniform distribution. The ranges are chosen around the usually used fixed values for the parameters, namely $a=1$, $b=5.1 / (4\pi^2)$, $c=5 / \pi$, $r=6$, $s=10$ and $t=1/(8\pi)$.

\subsection*{The Hartmann3D Ensemble}
\label{ssec:hartmann3}
The function for Hartmann3D \citep{dixon1978global} ensemble reads:
\begin{equation}
\begin{split}
    &f(x, \alpha_1, \alpha_2, \alpha_3, \alpha_4) =\\
    &\quad - \sum^{4}_{i=1} \alpha_{i} \exp \left( -\sum^3_{j=1} A_{i, j} (x_j - P_{i, j})^2 \right) \qquad x \in [0, 1] \\
    &\bm{A} = 
    \begin{bmatrix}
        3.0 & 10 & 30\\
        0.1 & 10 & 35\\
        3.0 & 10 & 30\\
        0.1 & 10 & 35
    \end{bmatrix}
    \quad
    \bm{P} = 10^{-4} \cdot
        \begin{bmatrix}
        3689 & 1170 & 2673\\
        4699 & 4387 & 7470\\
        1091 & 8732 & 5547\\
        381 & 5743 & 8828
    \end{bmatrix}
\end{split}
\end{equation}
To form the ensemble, we choose the distribution for the parameters as:
\begin{equation}
\begin{split}
    &\alpha_1 \sim \mathcal{U}(0.0, 2.0) \quad \alpha_2 \sim \mathcal{U}(0.0, 2.0)\\
    &\alpha_3 \sim \mathcal{U}(2.0, 4.0) \quad \alpha_4 \sim \mathcal{U}(2.0, 4.0)    
\end{split}
\end{equation}
Let $\tau = \{ \alpha_1, \alpha_2, \alpha_3, \alpha_4 \}$ and $p(\tau)$ is a four dimensional uniform distribution.


\end{document}